\let\mypdfximage\pdfximage
\def\pdfximage{\immediate\mypdfximage}
\documentclass[10pt,twocolumn,letterpaper]{article}

\usepackage[pagenumbers]{iccv} %

\usepackage{amsmath,amsfonts,bm}

\def\eqref#1{equation~\ref{#1}}

\def\1{\bm{1}}

\DeclareMathAlphabet{\mathsfit}{\encodingdefault}{\sfdefault}{m}{sl}
\SetMathAlphabet{\mathsfit}{bold}{\encodingdefault}{\sfdefault}{bx}{n}

\usepackage{url}

\usepackage{amsmath}
\usepackage{amssymb}

\usepackage{multirow}
\usepackage[export]{adjustbox}

\usepackage{xspace}

\usepackage{graphicx}
\usepackage{booktabs}

\usepackage[accsupp]{axessibility}  %
\usepackage{caption}
\usepackage{bbm}
\usepackage{arydshln}

\usepackage{listings}
\usepackage{placeins}
\usepackage{nicefrac}
\usepackage{xurl}
\usepackage{enumitem}
\usepackage{tabularx}
\usepackage{nicefrac}
\usepackage{makecell}
\usepackage{xspace}
\usepackage{graphbox}
\usepackage{pifont}
\usepackage{enumitem}
\usepackage{colortbl}
\usepackage{diagbox}
\usepackage{array}
\usepackage{wrapfig}
\usepackage{subcaption}
\usepackage{footmisc}
\usepackage{color}
\usepackage{slashbox}
\usepackage{diagbox}
\usepackage{float}
\usepackage{mathrsfs}
\usepackage{apptools}

\usepackage{caption}
\usepackage{rotating}   %

\usepackage{dblfloatfix}

\newcommand{\llava}{LLaVA\xspace}
\newcommand{\llavanext}{LLaVA-NeXT\xspace}

\newcommand{\qwenvl}{Qwen2-VL\xspace}

\newcommand{\pali}{PaliGemma\xspace}

\newcommand{\lnvic}{LN Vic\xspace}
\newcommand{\lnmis}{LN Mis\xspace}
\newcommand{\lnllama}{LN Llama\xspace}
\newcommand{\lnvicuna}{LN Vicuna\xspace}
\newcommand{\lnmistral}{LN Mistral\xspace}

\newcommand{\llama}{Llama\xspace}
\newcommand{\vicuna}{Vicuna\xspace}
\newcommand{\mistral}{Mistral\xspace}
\newcommand{\gemma}{Gemma\xspace}
\newcommand{\qwen}{Qwen2\xspace}
\newcommand{\prismatic}{Prismatic\xspace}

\newcommand{\siglip}{SigLIP\xspace}
\newcommand{\dino}{DinoV2\xspace}
\newcommand{\clip}{CLIP\xspace}

\newcommand{\owl}{OWLv2\xspace}

\newcommand{\obj}{OBJ\xspace}
\newcommand{\question}[1]{\text{qstn#1}}

\newcommand{\ours}{DASH\xspace}
\newcommand{\oursllm}{DASH-LLM\xspace}
\newcommand{\oursopt}{DASH-OPT\xspace}
\newcommand{\oursb}{DASH-B\xspace}

\newcommand{\doublemidrule}{
  \midrule
  \addlinespace[0.05em] %
  \midrule
}

\newcommand{\evalonly}{Validation\xspace}

\newcommand{\hallu}{FP-hallucination}
\newcommand{\yes}{Yes}
\newcommand{\no}{No}

\newcommand{\yn}[1]{{\color{black}#1}}

\definecolor{iccvblue}{rgb}{0.21,0.49,0.74}
\usepackage[pagebackref,breaklinks,colorlinks,allcolors=iccvblue]{hyperref}

\def\paperID{8489} %
\def\confName{ICCV}
\def\confYear{2025}

\title{\ours: Detection and Assessment of Systematic Hallucinations of VLMs}

\author{
  Maximilian Augustin\footnotemark \addtocounter{footnote}{-1}\\
\and
  Yannic Neuhaus\footnotemark \addtocounter{footnote}{-1}\\
  T\"ubingen AI Center – University of T\"ubingen\\
\and
  Matthias Hein \\
}

\begin{document}
\maketitle

\renewcommand{\thefootnote}{*}
\footnotetext{Equal contribution.}
\renewcommand{\thefootnote}{\arabic{footnote}}

\begin{abstract}
Vision-language models (VLMs) are prone to object hallucinations, where they erroneously indicate the presence of certain objects in an image. Existing benchmarks quantify hallucinations using relatively small, labeled datasets. However, this approach is i) insufficient to assess hallucinations that arise in open-world settings, where VLMs are widely used, and ii) inadequate for detecting systematic errors in VLMs. We propose
\textbf{\ours} (Detection and Assessment of Systematic Hallucinations), an automatic, large-scale pipeline designed to identify systematic hallucinations of VLMs on real-world images in an open-world setting.
A key component is \oursopt for image-based retrieval, where we optimize over the ``natural image manifold'' to generate images that mislead the VLM. The output of \ours consists of clusters of real and semantically similar images for which the VLM hallucinates an object. We apply \ours to \pali and two \llavanext models across 380 object classes and, in total, find more than $19k$ clusters with $950k$ images. 
We study the transfer of the identified systematic hallucinations to other VLMs and show that fine-tuning \pali with the model-specific images obtained with \ours mitigates object hallucinations. Code and data are available at \url{https://YanNeu.github.io/DASH}.
\end{abstract}
    
\section{Introduction}\label{sec:intro}
While vision-language models (VLMs) demonstrate remarkable comprehension of multimodal text and image data, they suffer from object hallucination errors %
such as incorrectly identifying objects that are not present or describing incorrect relationships or positions of objects within image descriptions~\cite{liu2024surveyvlmhallucination}. %
\begin{figure}[t]
    \centering
    \includegraphics[width=\linewidth]{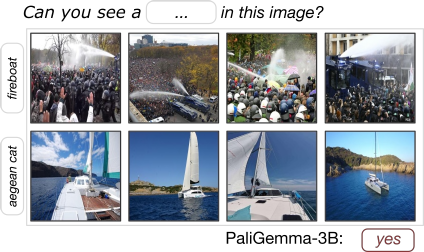}
    \caption{\textbf{\ours:} %
    Systematic Hallucinations of PaliGemma-3B.}
    \label{fig:teaser}
\end{figure}

Benchmarks like POPE \cite{li2023evaluating} and AMBER \cite{wang2023llm} assess the erroneous indication of the presence or absence of an object by a VLM, but are limited by %
the curated datasets they use, such as MSCOCO \cite{lin2014coco}. While these benchmarks provide a first assessment of these errors, they
have two main issues:
\begin{enumerate}
    \item  There is no systematic assessment of the types of images for which VLMs hallucinate%
, making it difficult to determine whether these errors are random or indicative of a systematic issue.

\item The reliance on small, curated datasets like MSCOCO  with a small number of object classes fails to accurately reflect the open-world application of VLMs in real-world scenarios. This could potentially lead to overlooking significant problems due to the limited and biased nature of these datasets. 
\end{enumerate}
\noindent In this paper, we focus on object hallucinations %
 of VLMs where the models respond ``\yes'' to the question ``Can you see an \textit{object} in this image?''\footnote{We apply the suffix ``Please answer only with yes or no.'' to ensure valid VLM responses.} although the \textit{object} is not actually present. For brevity, we refer to this type of object hallucination as \textbf{false-positive-hallucination (\hallu)} throughout the rest of the paper. The opposite case, where the VLM incorrectly answers ``\no'' despite the object's presence, is also of interest but can be %
 checked with existing detection benchmarks. As we demonstrate in this paper, object hallucinations may occur in entirely unexpected contexts (see Fig.~\ref{fig:teaser}). Thus, addressing them requires an open-world approach not limited to small benchmark datasets.
Instead we argue that assessing and fixing object hallucinations of a VLM requires (i) exploring web-scale datasets like ReLAION-5B \cite{relaion5b}, and (ii) finding model-specific hallucinations for the given VLM. %
We make the following contributions in this paper:
\begin{itemize}
\item We propose \ours (\underline{D}etection and \underline{A}ssessment of \underline{S}ystematic \underline{H}allucinations), a large-scale, fully automated pipeline requiring no human labeling for identifying systematic \hallu s in VLMs by detecting clusters of semantically similar images causing them. %
It consists of \oursllm, text-based retrieval using queries generated by a LLM, and \oursopt, image-based retrieval with generated images in ReLAION-5B.
\item In \oursopt, we propose a method for optimizing the generative process of a latent diffusion model to produce images where the VLM hallucinates an object, while an open-world object detector has low confidence that the object is present.
\item We apply \oursopt and \oursllm to \pali \cite{beyer2024paligemma} and \llavanext (Mistral and Vicuna) \cite{liu2024llavanext}. In total, we find more than $19k$ clusters of %
object hallucinations comprising 
more than $950k$ images.
\item We show that the hallucinations for the found images successfully transfer to seven other VLMs, including one of the top open-weights models, QwenV2-72B \cite{Qwen2VL}, on the HuggingFace Open VLM leaderboard~\cite{duan2024vlmevalkit}. %
\item We propose a new benchmark, \oursb, to %
enable a more reliable evaluation of this issue in current VLMs.
\item We show that fine-tuning \pali on our
large dataset of object hallucinations can help mitigate the problem.
\end{itemize}

\section{Related Work}

\textbf{Hallucinations of VLMs: Benchmarks and Mitigation.}
An early work to benchmark hallucinations in short image captions of VLMs was CHAIR \cite{rohrbach2018CHAIR}. A recent study \cite{kaul2024throne} differentiates between type I hallucinations in free-form answers and type II hallucinations in response to factual questions about the image (\eg ``Is there a car in this image? Answer yes or no''). POPE \cite{li2023evaluating} benchmarks %
type II hallucinations; however, being limited to 80 objects in MSCOCO it does not capture the variability in VLM usage. AMBER \cite{wang2023llm} addresses different kinds of hallucinations including ``existential'' (object hallucinations). While object hallucinations saturate in their benchmark for VLMs, we demonstrate that a substantial number of systematic object hallucinations persist even in current models such as \qwenvl-72B~\cite{Qwen2VL} and Llama 3.2-VL-11B~\cite{dubey2024llama3herdmodels}. 
Other recent benchmarks such as HALO-QUEST \cite{wang2024haloquest} (type II) or MMVP~\cite{tong2024eyeswideshutexploring} (type I and II) cover multiple modalities but are too small to provide sufficient statistics on specific VLM errors.

The literature on mitigation of hallucinations is limited. Visual contrastive decoding \cite{leng2023mitigatingobjecthallucinationslarge} is a training-free method contrasting original and noisy images, robust instruction tuning \cite{liu2024mitigatinghallucinationlargemultimodal}, LURE \cite{zhou2024analyzingmitigatingobjecthallucination} is a post-hoc method aimed at reducing hallucinations, and M-HalDetect  \cite{gunjal2024detectingpreventinghallucinationslarge} provides a detection (type I) and mitigation technique.

\textbf{Spurious Correlations/Hallucinations in Image Classification:} Spurious correlations in image-based data \cite{CHOI2012853}, such as a cow on the beach not being recognized due to the absence of the spurious feature (e.g., grass), are known issues in image classification. In \cite{singla2021salient,neuhaus2023spurious}, spurious features are identified on ImageNet, and \cite{neuhaus2023spurious} proposes ``Spurious ImageNet'', a benchmark for hallucinations in ImageNet classifiers (classifier predicts presence of class despite not being present in the image) for 100 ImageNet classes. 

\textbf{Debugging of ML Models using Guided Image Generation.}
\citet{zhang2024imagenetd} utilize stable diffusion to generate images with variations in background, material, and texture to deceive models like CLIP in zero-shot classification, showing partial transfer to VQA-type tasks in miniGPT4 and \llava-1.5. Similarly, \citet{metzen2023identification} identify errors in image classifiers on rare subgroups by evaluating generated images with variations in color, size, background, and weather. DiG-IN \cite{augustin2024dig} is a debugging technique that uses optimization-guided image generation, for instance, to find images that maximize prediction differences for a certain class between two classifiers. \cite{augustin2024dig} employ this approach to systematically identify subgroups that CLIP misclassifies.

\begin{figure*}
    \centering
    \includegraphics[width=0.9\linewidth]{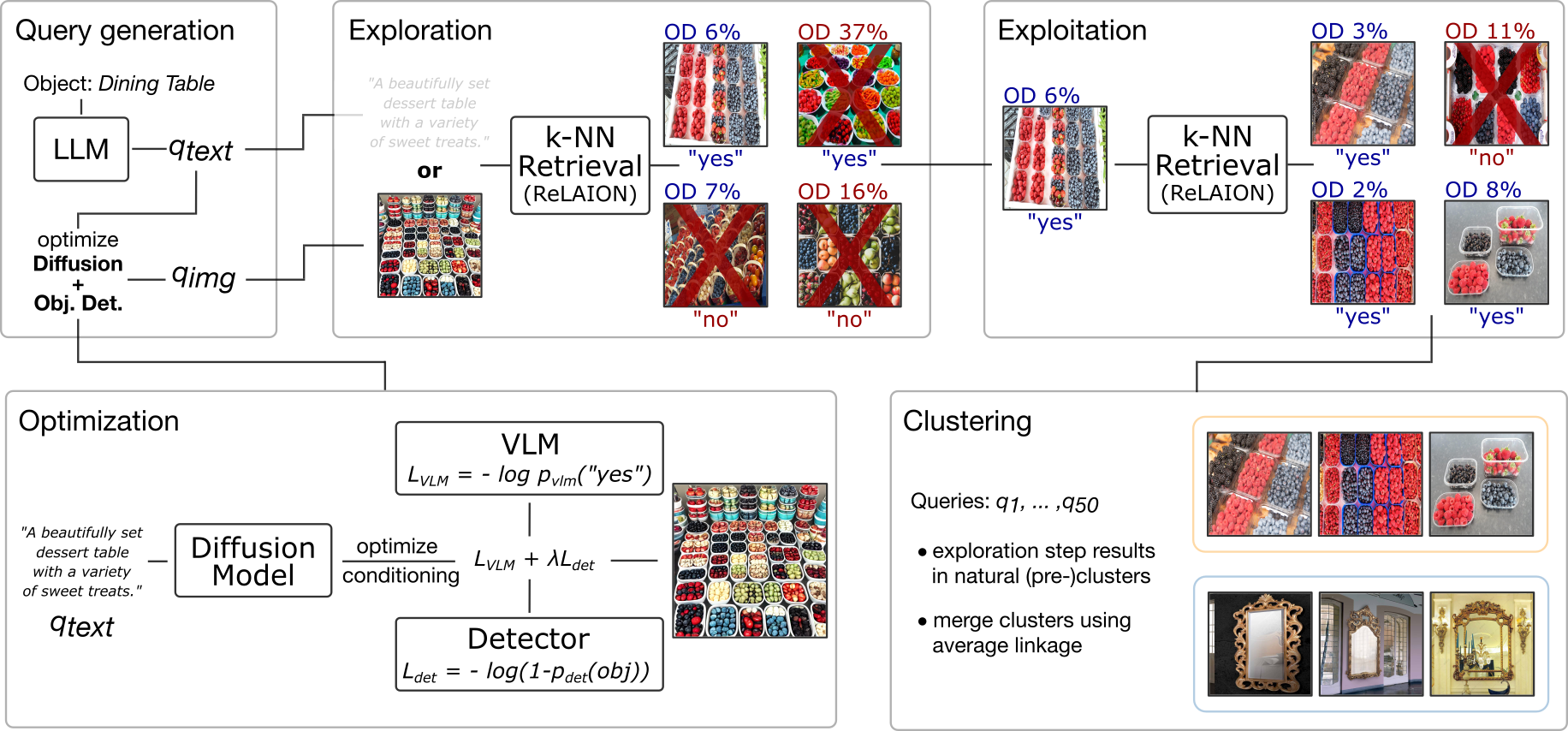}
    \vspace{-1mm}
    \caption{\textbf{\ours:} Given an object class, e.g. dining table, we generate text-based queries with \textbf{\oursllm} or image-based queries with \textbf{\oursopt}. \textbf{Optimization:} we optimize the latent variables of a diffusion process to generate an image which yields ``yes'' for the VLM (``Can you see a dining table in this image?'') and at the same time the object detector states that no ``dining table'' is present in the image. \textbf{Exploration:} the text and image queries are used for kNN-retrieval using CLIP similarity on ReLaion-5B. \textbf{Exploitation:} for successful images (VLM ``yes'', object detector ``no'') of the exploration phase we retrieve novel images via kNN-retrieval to check if the hallucination transfers to semantically similar images. \textbf{Clustering:} Finally, we cluster successful images of the exploitation step into semantically similar clusters of hallucinations of the VLM.
    }
    \vspace{-2mm}
    \label{fig:pipeline-overview}
\end{figure*}

\section{\ours: Detection and Assessment of 
Systematic Hallucinations}\label{sec:method}

The goal of \ours is to identify systematic \hallu s by detecting clusters of semantically similar images that trigger them (see Fig.~\ref{fig:teaser} or \ref{fig:retrieval} for examples found by \ours). This type of object hallucination is increasingly relevant as AI agents begin to automatically process image data. 
We tackle this challenge by searching over ReLAION-5B, avoiding the limitations of smaller datasets that cover only few object classes compared to the unrestricted application of VLMs. Importantly, an object category's presence in a dataset does not imply that only objects from this dataset could lead the VLM to hallucinate. Thus, benchmarks such as POPE \cite{li2023evaluating} or AMBER \cite{wang2023llm} underestimate the problem of \hallu s, while exhaustive testing of each image-object combination, as in \cite{kaul2024throne}, is infeasible for ReLAION-5B.

\ours is a fully automatic pipeline, designed to require no human labels. While this carries the risk of errors—which we test in our evaluation—it enables \ours to produce large-scale datasets that can be used for evaluation and fine-tuning of VLMs. 
To effectively detect systematic hallucinations, we propose two approaches:  \oursopt, based on images specifically generated to deceive the VLM, and \oursllm, based on LLM-generated queries. Subsequently, these image and text queries are used in an exploration phase to find real images in ReLAION-5B which trigger \hallu s.
In an exploitation step, we verify that these are not isolated errors by identifying semantically similar images that consistently deceive the VLM without containing the object. Finally, as different queries may yield overlapping image groups, we cluster the results (see Fig. \ref{fig:all_cluster} for samples from all clusters for the object ``dam'' found by \ours).
We summarize the workflow of \ours in Fig. \ref{fig:pipeline-overview}.

\subsection{\oursllm}\label{sec:oursllm}
\hallu s often arise from associations between the target \textit{object} and other objects present in an image. This can result from co-occurrence at the image level (i.e., objects that are frequently photographed together) or co-occurrence in texts. We leverage the fact that large language models (LLMs) are trained on extensive text corpora and are well-suited for generating lists of candidate prompts.
Specifically, we use \llama 3.1-70B with a carefully crafted system prompt to generate 50 text prompts, $q_1,\dots,q_{50}$, for each object. We ask the model to ``create prompts that could lead an AI model to falsely recognize the object due to the presence of spurious features, even though the object itself is not present in the images'' (see~\cref{sec:app_llm_prompt}). The LLM is instructed to avoid mentioning the object itself or any of its parts in the prompts and to avoid repeating queries.

While this approach to generating text queries is simple, it proves to be quite effective. However, it has two limitations: first, it is entirely agnostic to the specific VLM being used. Second, even if the generated text queries are conceptually capable of causing the VLM to hallucinate, our CLIP-similarity-based retrieval on ReLAION-5B (see~\cref{sec:explore-exploit}), may fail to retrieve the appropriate images. We address these limitations of \oursllm by using image queries generated through \oursopt, which are directly optimized to induce hallucinations in the VLM.

\subsection{\oursopt}\label{sec:oursopt}

The goal of \oursopt is to generate images that cause \hallu s in the VLM. %
As our pipeline is supposed to be fully automatic, we require an alternative to human verification to determine whether the image contains the object. For this purpose, we employ an open-world object detector---in this case, OWLv2 \cite{minderer2024scaling}---with a very low detection threshold. This minimizes the chance of missing the object when it is actually in the image. However, some images may be incorrectly flagged as containing the object. %
We discuss this trade-off in more detail in \cref{app:objectdetector}.

Typically, optimizing in pixel space results in adversarial samples; however, previous work \cite{augustin2024dig} has shown that meaningful solutions in the natural image manifold can be obtained by optimizing the input variables of a diffusion process. While prior works \cite{augustin2024dig, samuel2024generating} rely on multi-step diffusion processes—which are computationally intensive—we utilize recent advances in diffusion distillation~\cite{ren2024hyper} and use a single-step diffusion process, significantly reducing computational cost. We denote by $q(C)$ the output image of the diffusion process based on its conditioning $C$ (see~\cref{sec:app_diffusion} for details). Below, we describe the two objectives we optimize to obtain the desired query image $q(C)$.

We denote the text query ``Can you see a \obj in this image?'' by $\question{\obj}$ for the considered object class \obj. 
Since ``\yes'' corresponds to a single token, we can directly optimize the cross-entropy loss for the standard next-token prediction of ``\yes'' in the VLM based on the image $q(C)$:
\begin{equation}
L_\text{vlm}(C) =  -\log p_\text{vlm}\left( \text{``\yes''} \mid q(C), \question{\obj} \right).
\end{equation}
Similarly, we propose a loss function to penalize the object detector's confidence in detecting the object class \obj in the generated query image $q(C)$:
\begin{equation}
L_\text{det}(C) =  -\log \left( 1 - p_\text{det}\left(  \text{\obj} \mid q(C) \right) \right),
\end{equation}
where $p_\text{det}\left(  \text{\obj} \mid q(C) \right)$ is the object detector's confidence that the generated query image $q(C)$ contains the %
class \obj.

The final optimization problem for the conditioning $C$ generating the image $q(C)$ is then given as:
\begin{equation}
\label{eq:oursopt}
\begin{split}
    \min_C \, L_\text{vlm}(C) + L_\text{det}(C)
\end{split}
\end{equation}
This objective aims to make the VLM hallucinate the object in the resulting image $q(C)$, while the detection loss ensures that the optimization process does not simply insert the object into the image.

In practice, we initialize the optimization of $C$ in the diffusion process using real text prompts encoded by the diffusion model's text encoder. We start with the same 50 LLM-generated prompts from \oursllm, creating 50 image queries $q_1, \dots, q_{50}$. However, our optimization process steers the resulting image $q(C)$ away from images that either fail to fool the VLM or actually contain the object. Importantly, while the initial text prompts are derived purely from the LLM and are independent of the VLM, the image queries are directly optimized for the VLM, allowing us to uncover model-specific issues.
 
\begin{figure*}[htbp]
    \centering
    \footnotesize
    \setlength{\tabcolsep}{0.5pt} %
    \renewcommand{\arraystretch}{0.8}
    \begin{tabular}{c|cccccccccc}
         {\textbf{\obj}} & \multicolumn{9}{c}{\textbf{\oursllm}}\\
        \toprule
         {\textbf{ptarmigan}} &
        \multicolumn{10}{c}{ Cluster Size: 190, Query: A mountain valley with a few scattered trees and a stream.}\\
        \includegraphics[width=0.083\textwidth,height=0.075\textwidth]{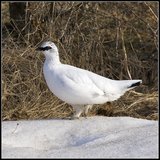} &
        \includegraphics[width=0.083\textwidth,height=0.075\textwidth]{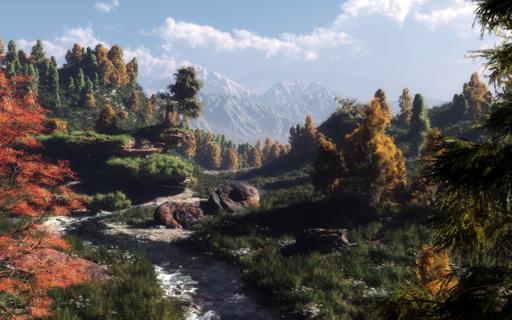} &
        \includegraphics[width=0.083\textwidth,height=0.075\textwidth]{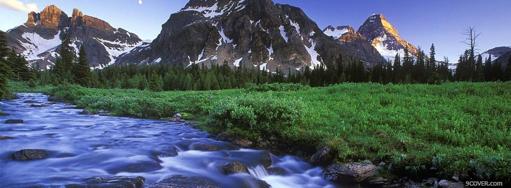} &
        \includegraphics[width=0.083\textwidth,height=0.075\textwidth]{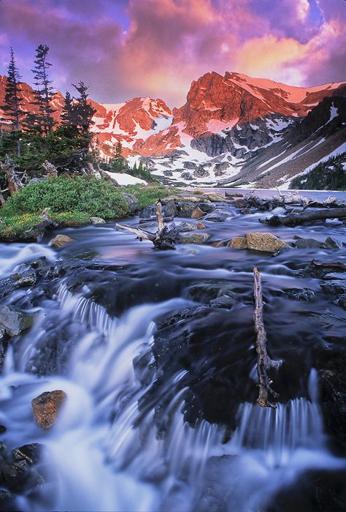} &
        \includegraphics[width=0.083\textwidth,height=0.075\textwidth]{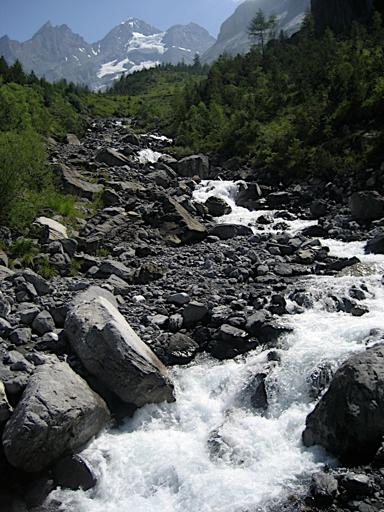} &
        \includegraphics[width=0.083\textwidth,height=0.075\textwidth]{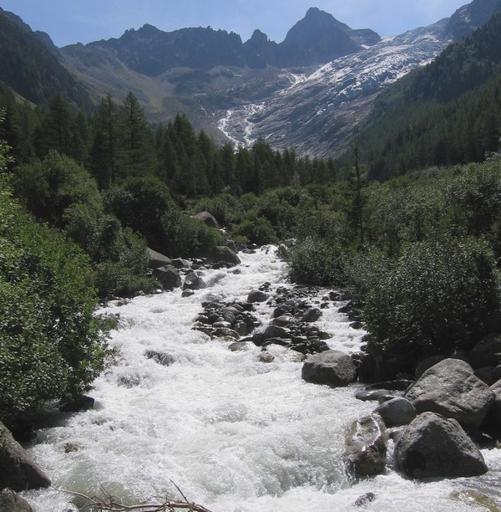} &
        \includegraphics[width=0.083\textwidth,height=0.075\textwidth]{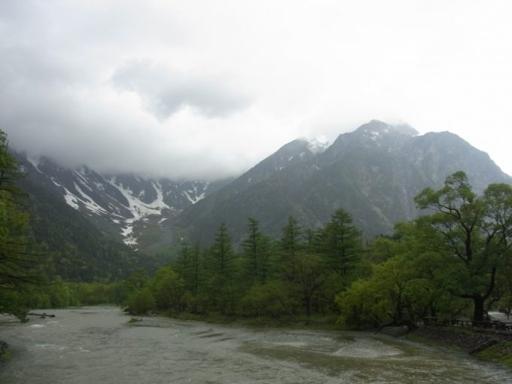} &
        \includegraphics[width=0.083\textwidth,height=0.075\textwidth]{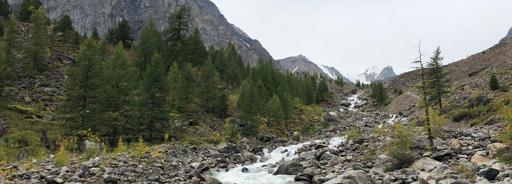} &
        \includegraphics[width=0.083\textwidth,height=0.075\textwidth]{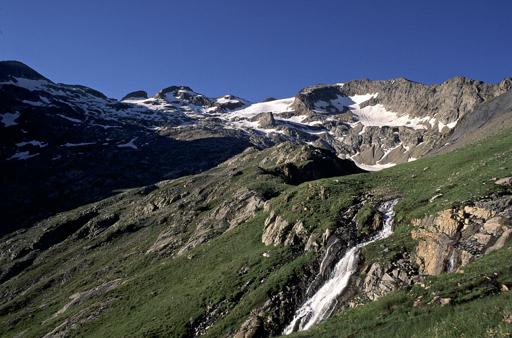} &
        \includegraphics[width=0.083\textwidth,height=0.075\textwidth]{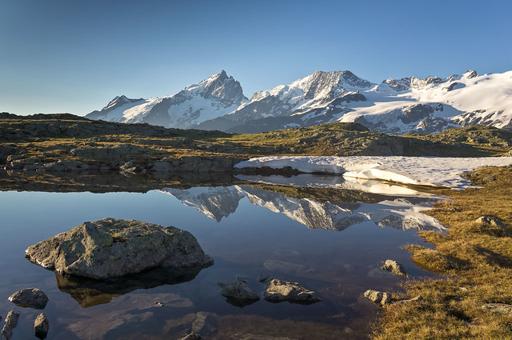} &
        \includegraphics[width=0.083\textwidth,height=0.075\textwidth]{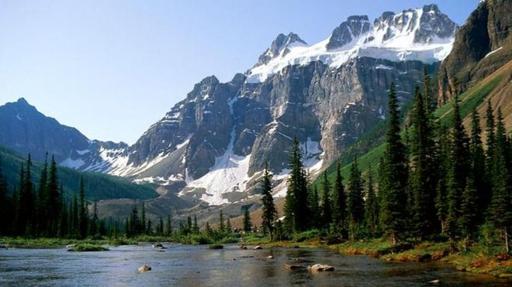} \\
        
        \midrule
         {\scriptsize\textbf{baumkuchen}} & 
        \multicolumn{10}{c}{Cluster Size: 389, Query: A traditional German Christmas pyramid with candles and ornaments. }\\
        \includegraphics[width=0.083\textwidth,height=0.075\textwidth]{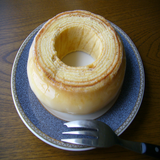} &
        \includegraphics[width=0.083\textwidth,height=0.075\textwidth]{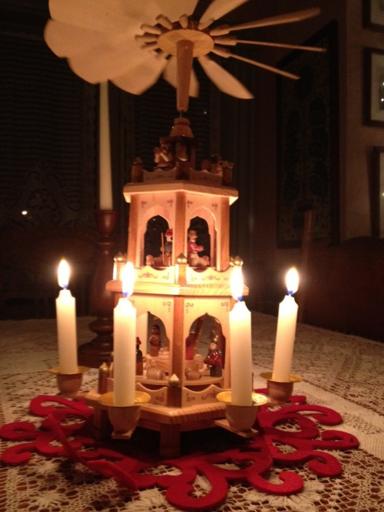} &
        \includegraphics[width=0.083\textwidth,height=0.075\textwidth]{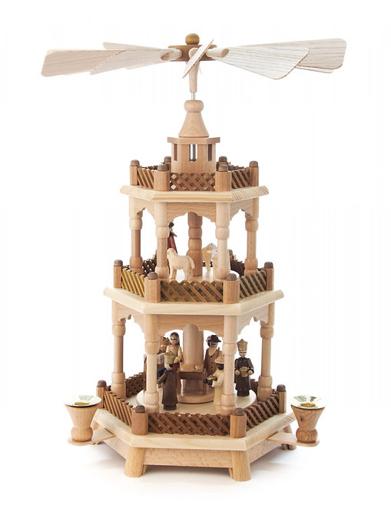} &
        \includegraphics[width=0.083\textwidth,height=0.075\textwidth]{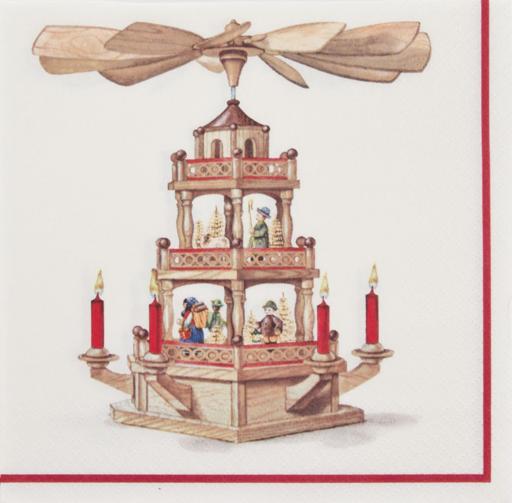} &
        \includegraphics[width=0.083\textwidth,height=0.075\textwidth]{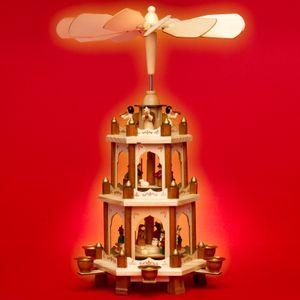} &
        \includegraphics[width=0.083\textwidth,height=0.075\textwidth]{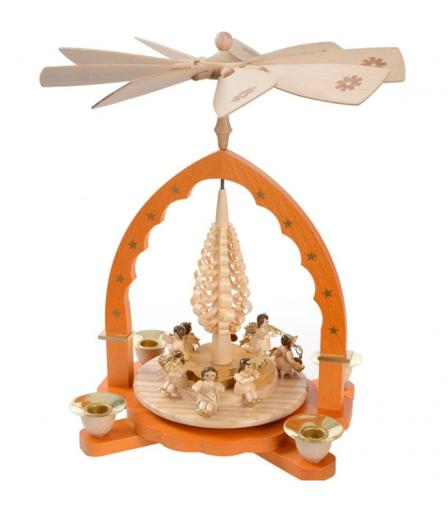} &
        \includegraphics[width=0.083\textwidth,height=0.075\textwidth]{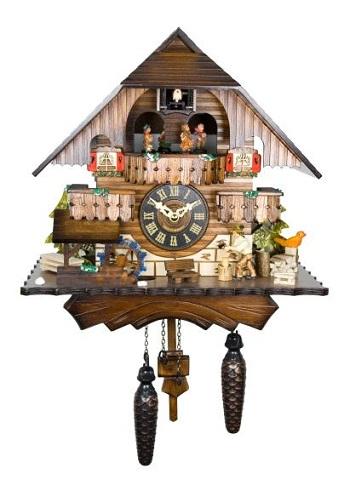} &
        \includegraphics[width=0.083\textwidth,height=0.075\textwidth]{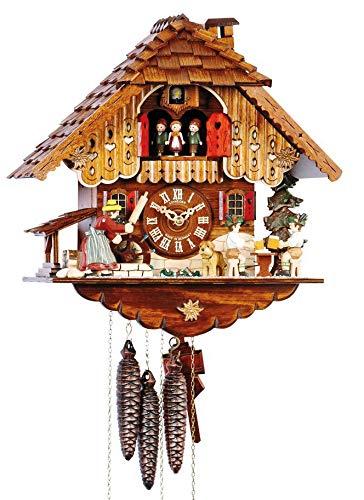} &
        \includegraphics[width=0.083\textwidth,height=0.075\textwidth]{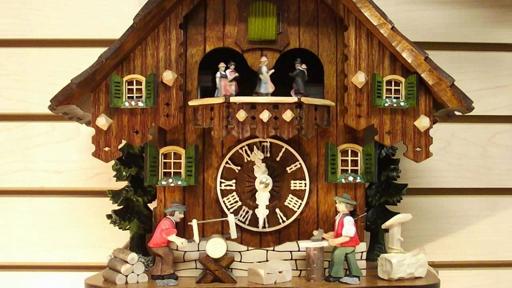} &
        \includegraphics[width=0.083\textwidth,height=0.075\textwidth]{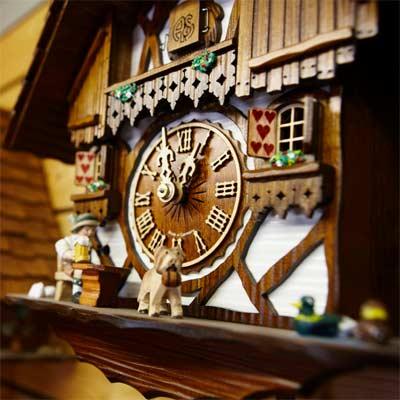} &
        \includegraphics[width=0.083\textwidth,height=0.075\textwidth]{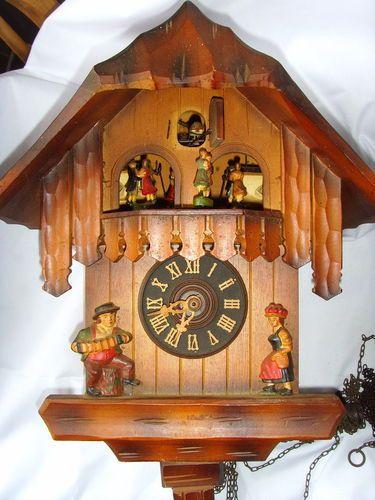}        \\
        \midrule
         {\textbf{cello}} &
        \multicolumn{10}{c}{Cluster Size: 62, Query: A music sheet with intricate notes and markings.}\\
        \includegraphics[width=0.083\textwidth,height=0.075\textwidth]{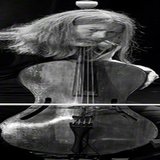} &
        \includegraphics[width=0.083\textwidth,height=0.075\textwidth]{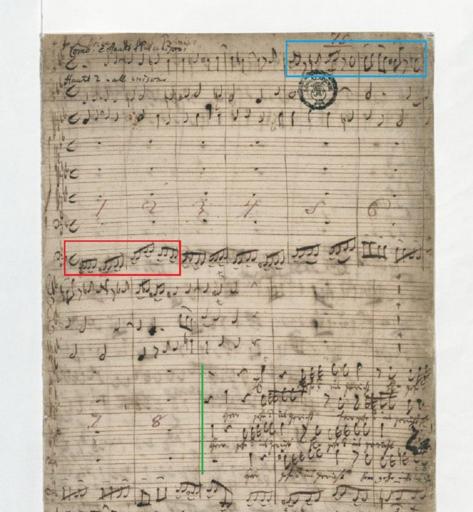} &
        \includegraphics[width=0.083\textwidth,height=0.075\textwidth]{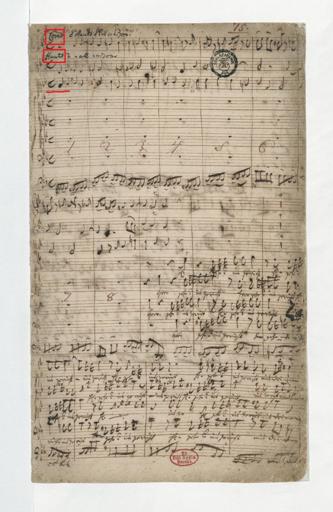} &
        \includegraphics[width=0.083\textwidth,height=0.075\textwidth]{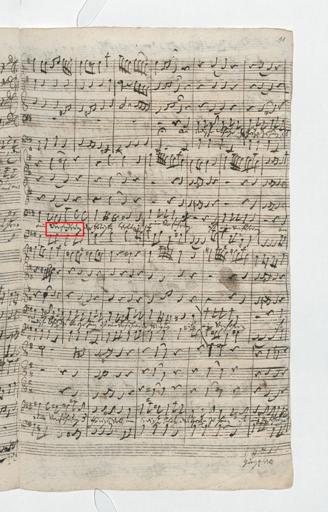} &
        \includegraphics[width=0.083\textwidth,height=0.075\textwidth]{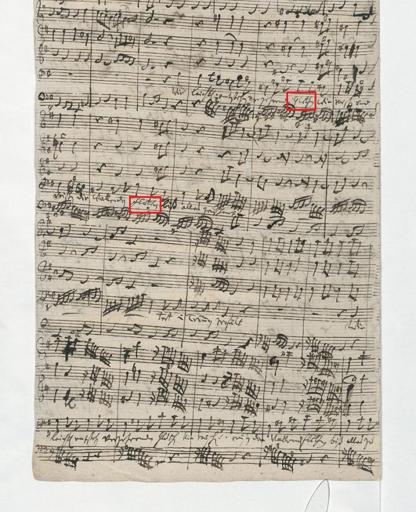} &
        \includegraphics[width=0.083\textwidth,height=0.075\textwidth]{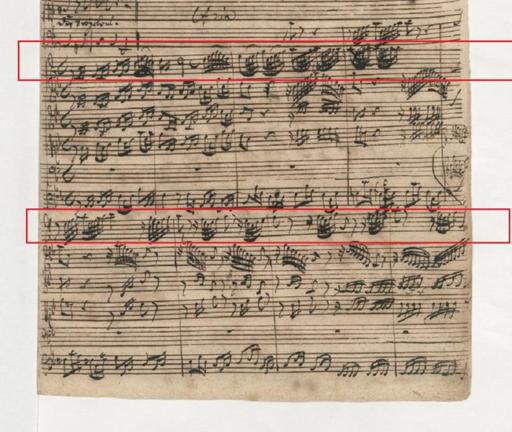} &
        \includegraphics[width=0.083\textwidth,height=0.075\textwidth]{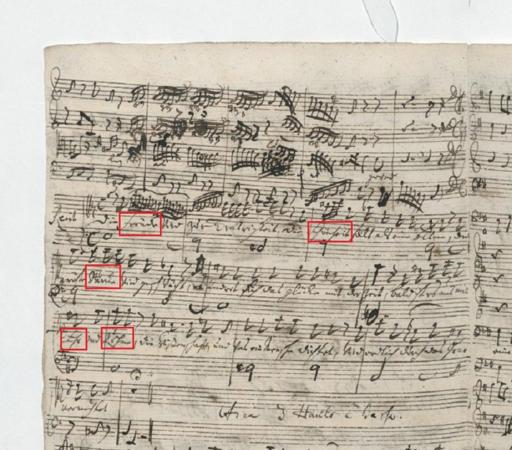} &
        \includegraphics[width=0.083\textwidth,height=0.075\textwidth]{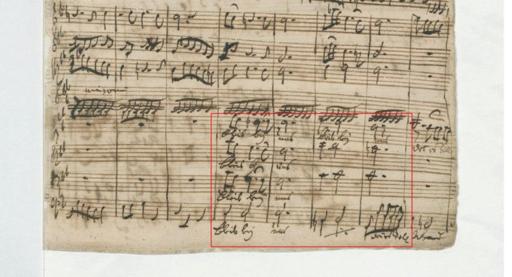} &
        \includegraphics[width=0.083\textwidth,height=0.075\textwidth]{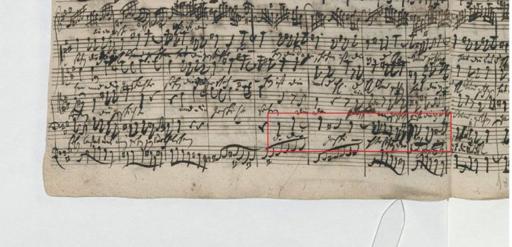} &
        \includegraphics[width=0.083\textwidth,height=0.075\textwidth]{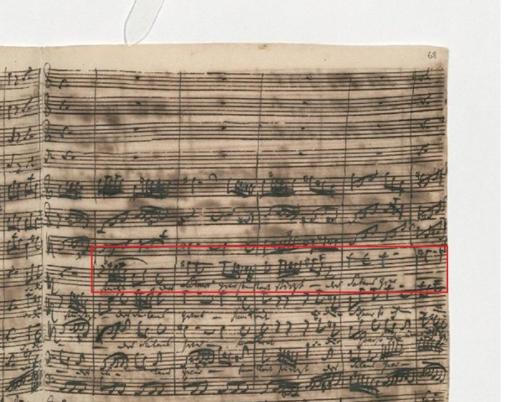} &
        \includegraphics[width=0.083\textwidth,height=0.075\textwidth]{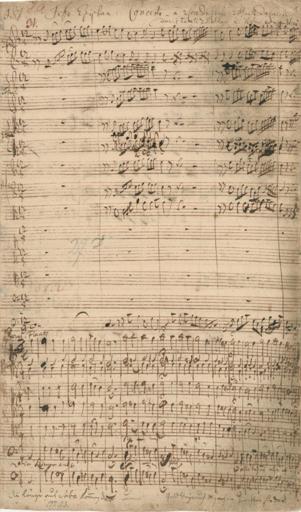} \\ 
        \bottomrule
        \multicolumn{9}{c}{}\\
        \end{tabular}

    \begin{tabular}{c|c|ccccccccc}
        {\textbf{\obj}} & \multicolumn{9}{c}{\textbf{\oursopt}}\\
        \toprule
        {\textbf{tench}} & Query &
        \multicolumn{7}{c}{Cluster Size: 170}\\
        \includegraphics[width=0.083\textwidth,height=0.075\textwidth]{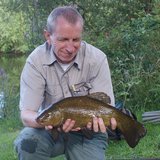} &
        \includegraphics[width=0.083\textwidth,height=0.075\textwidth]{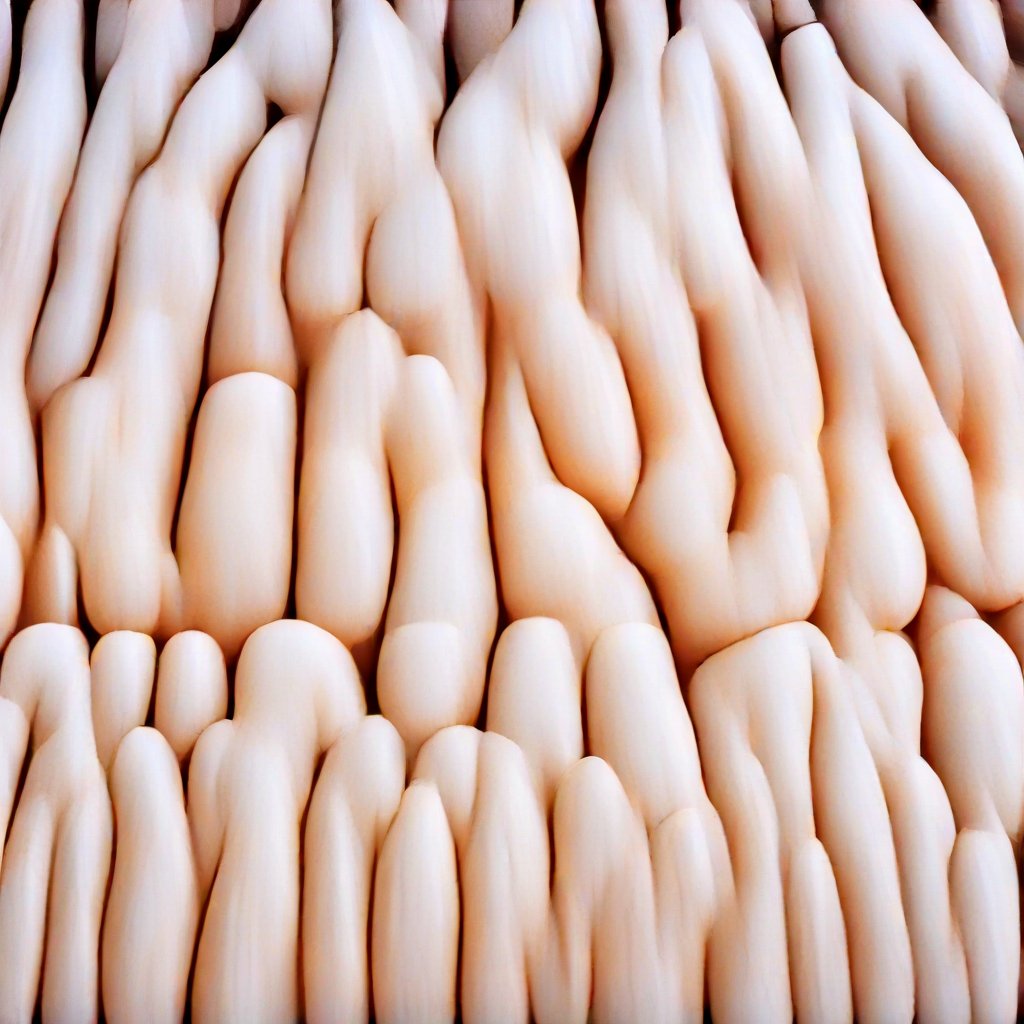} &
        \includegraphics[width=0.083\textwidth,height=0.075\textwidth]{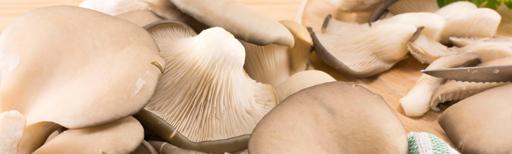} &
        \includegraphics[width=0.083\textwidth,height=0.075\textwidth]{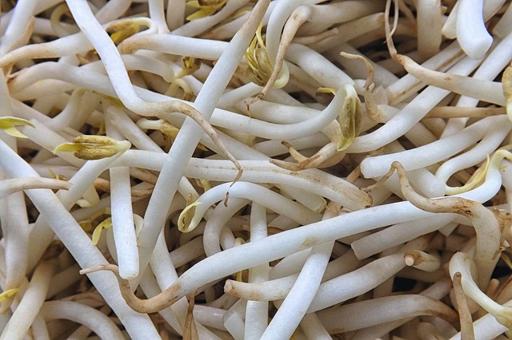} &
        \includegraphics[width=0.083\textwidth,height=0.075\textwidth]{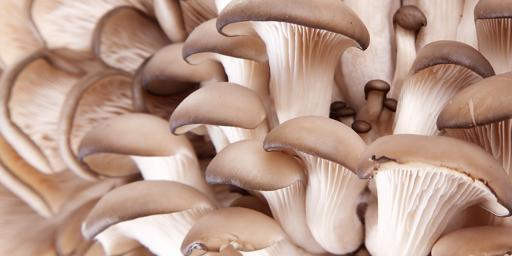} &
        \includegraphics[width=0.083\textwidth,height=0.075\textwidth]{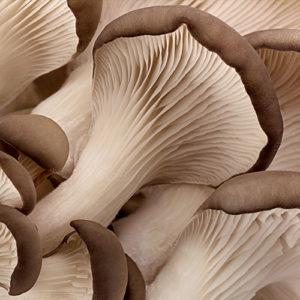} &
        \includegraphics[width=0.083\textwidth,height=0.075\textwidth]{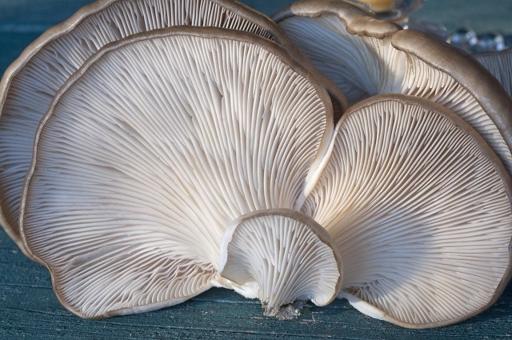} &
        \includegraphics[width=0.083\textwidth,height=0.075\textwidth]{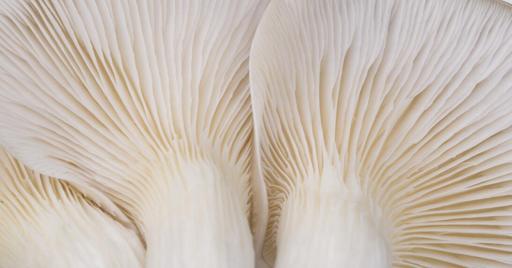} &
        \includegraphics[width=0.083\textwidth,height=0.075\textwidth]{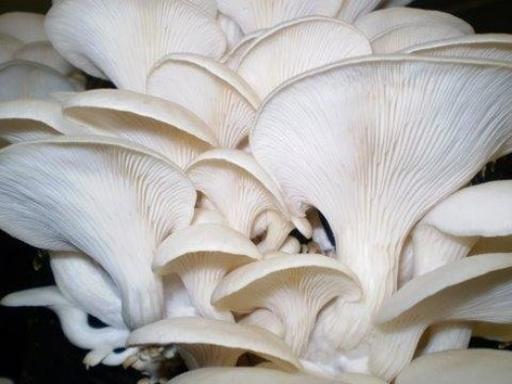} &
        \includegraphics[width=0.083\textwidth,height=0.075\textwidth]{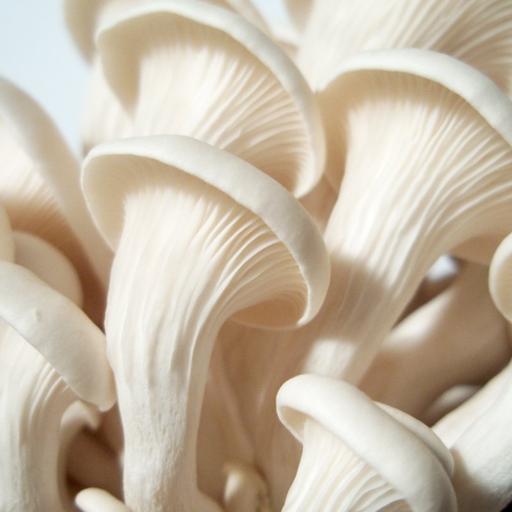} &
        \includegraphics[width=0.083\textwidth,height=0.075\textwidth]{images/retrieval/pali_img/si_tench/48_8.jpg} 
        \\
        \midrule
        {\textbf{leopard}}& Query &
        \multicolumn{7}{c}{Cluster Size: 276}\\
        \includegraphics[width=0.083\textwidth,height=0.075\textwidth]{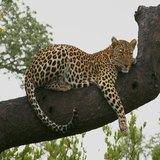} &
        \includegraphics[width=0.083\textwidth,height=0.075\textwidth]{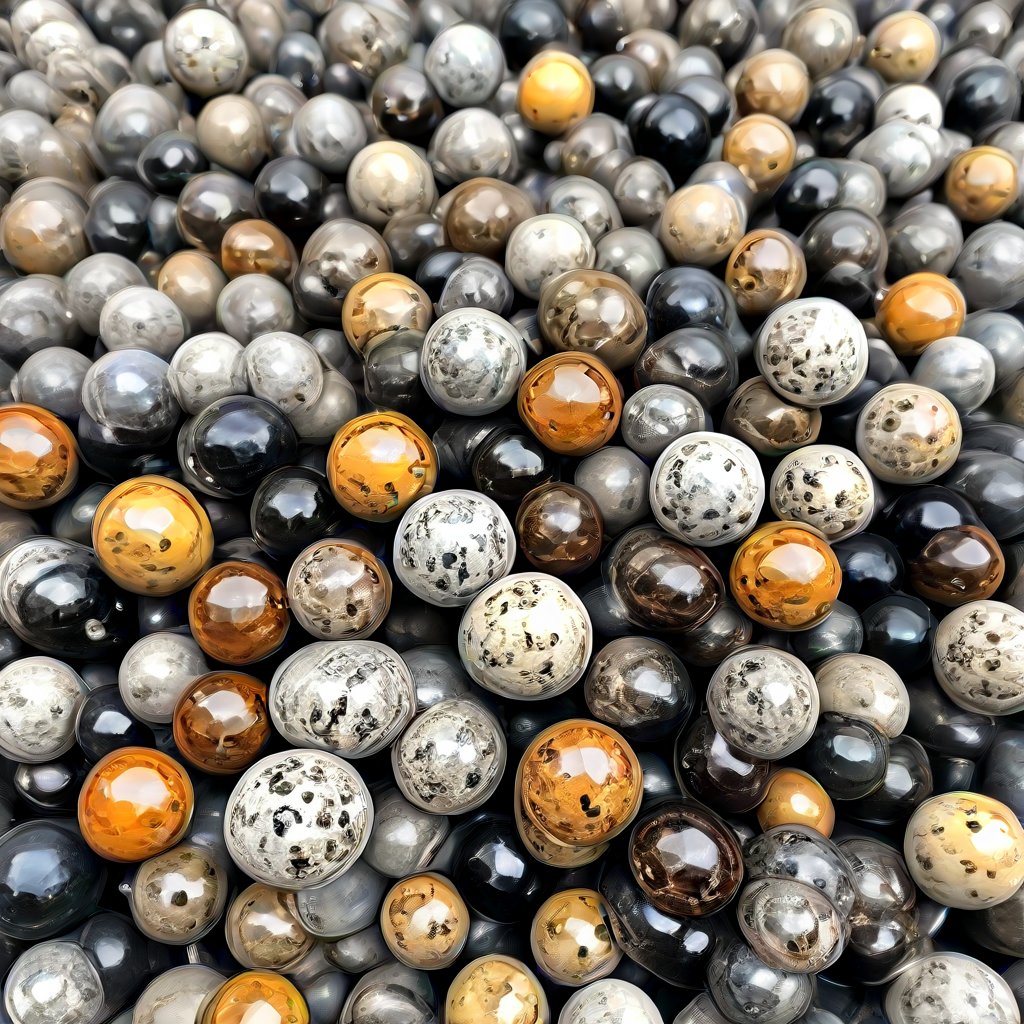} &
        \includegraphics[width=0.083\textwidth,height=0.075\textwidth]{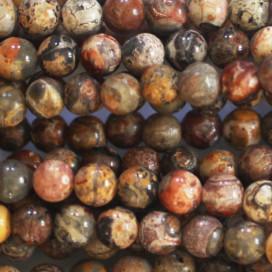} &
        \includegraphics[width=0.083\textwidth,height=0.075\textwidth]{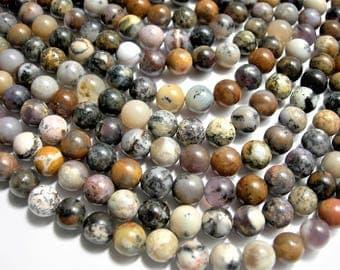} &
        \includegraphics[width=0.083\textwidth,height=0.075\textwidth]{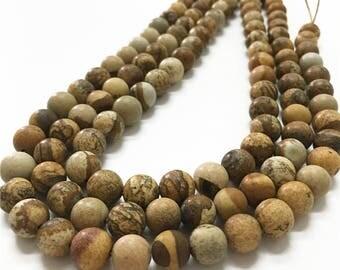} &
        \includegraphics[width=0.083\textwidth,height=0.075\textwidth]{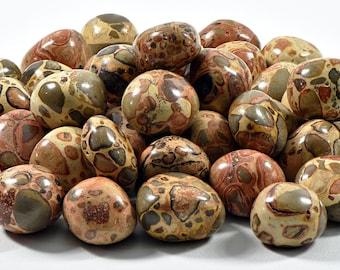} &
        \includegraphics[width=0.083\textwidth,height=0.075\textwidth]{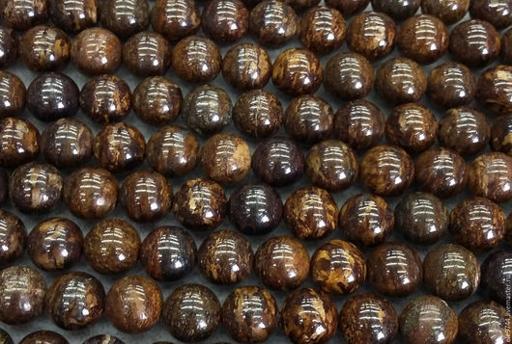} &
        \includegraphics[width=0.083\textwidth,height=0.075\textwidth]{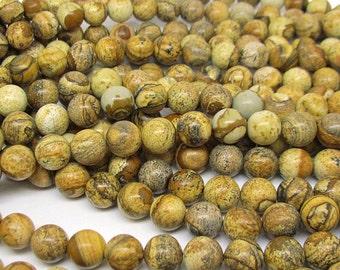} &
        \includegraphics[width=0.083\textwidth,height=0.075\textwidth]{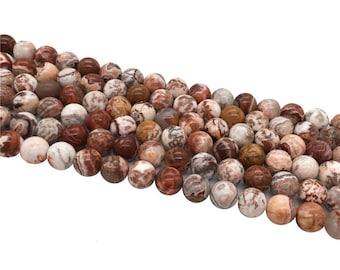} &
        \includegraphics[width=0.083\textwidth,height=0.075\textwidth]{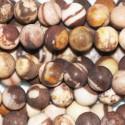} &
        \includegraphics[width=0.083\textwidth,height=0.075\textwidth]{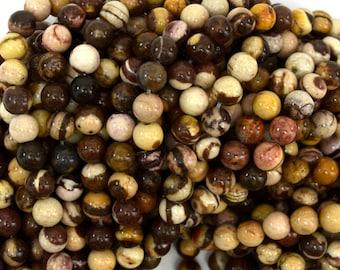}
        \\
        \midrule
         {\textbf{piano}} & Query &
        \multicolumn{7}{c}{Cluster Size: 150}\\
        \includegraphics[width=0.083\textwidth,height=0.075\textwidth]{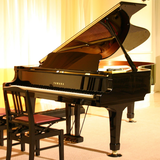} &
        \includegraphics[width=0.083\textwidth,height=0.075\textwidth]{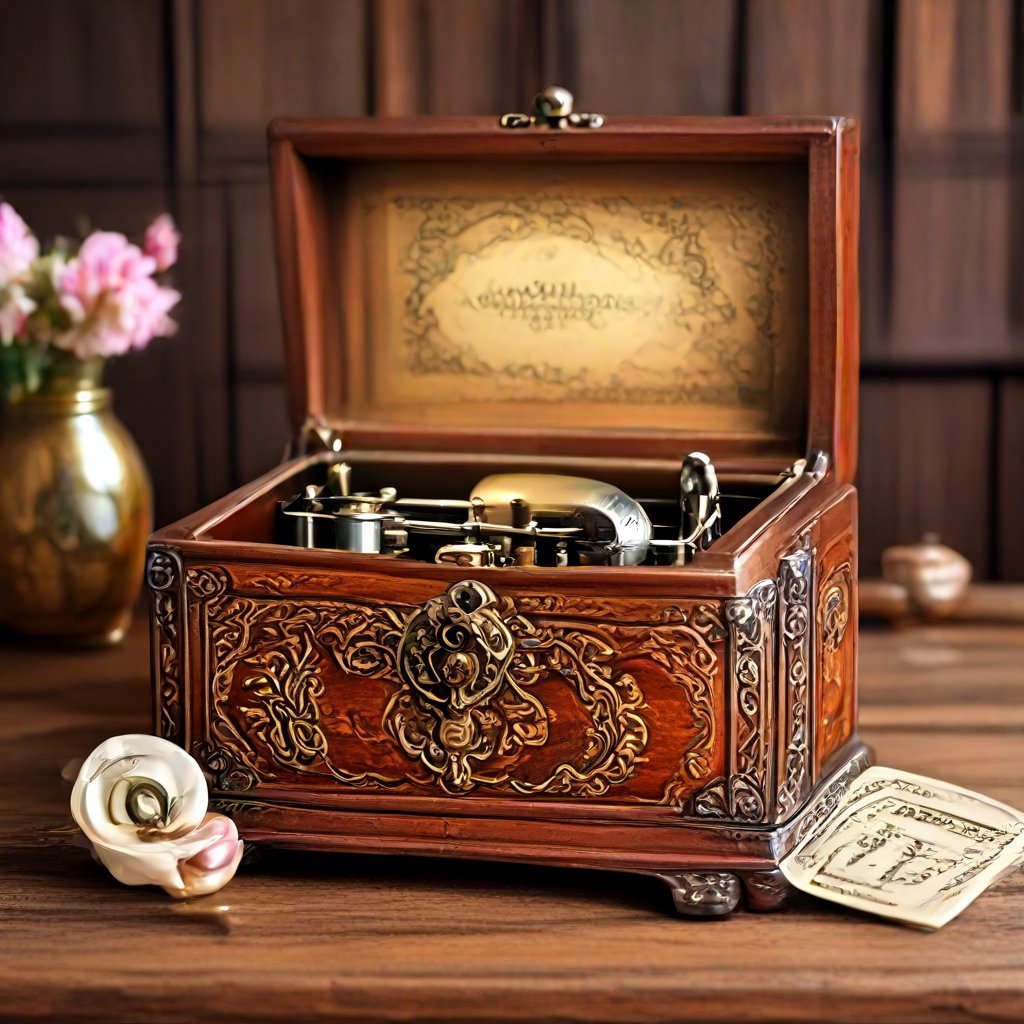} &
        \includegraphics[width=0.083\textwidth,height=0.075\textwidth]{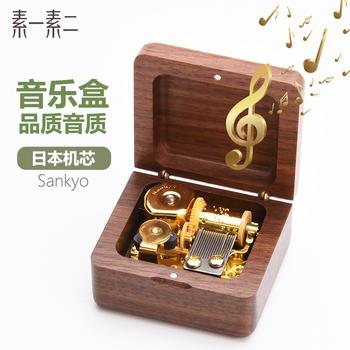} &
        \includegraphics[width=0.083\textwidth,height=0.075\textwidth]{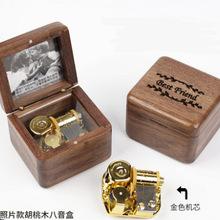} &
        \includegraphics[width=0.083\textwidth,height=0.075\textwidth]{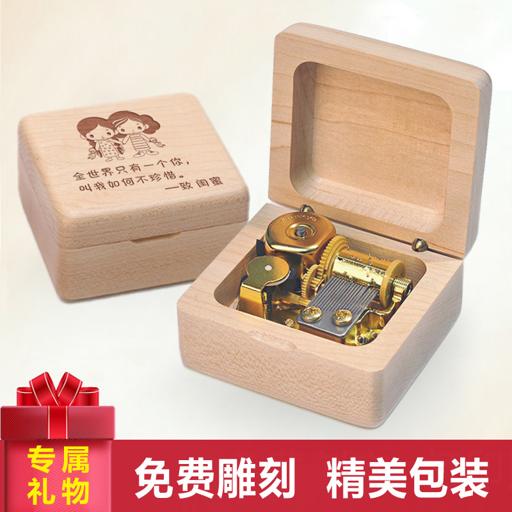} &
        \includegraphics[width=0.083\textwidth,height=0.075\textwidth]{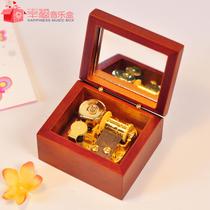} &
        \includegraphics[width=0.083\textwidth,height=0.075\textwidth]{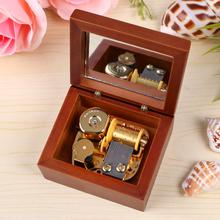} &
        \includegraphics[width=0.083\textwidth,height=0.075\textwidth]{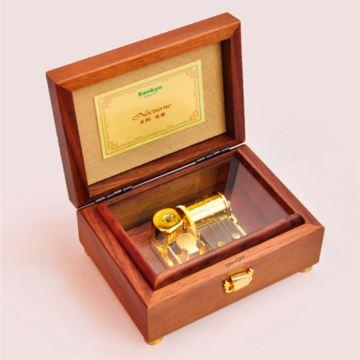} &
        \includegraphics[width=0.083\textwidth,height=0.075\textwidth]{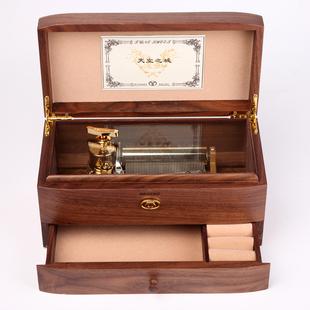} &
        \includegraphics[width=0.083\textwidth,height=0.075\textwidth]{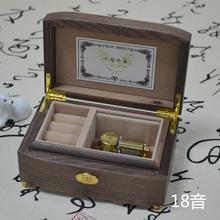} &
        \includegraphics[width=0.083\textwidth,height=0.075\textwidth]{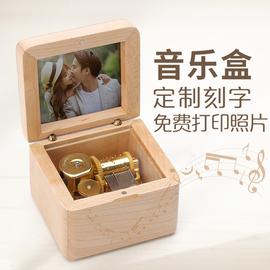} 
        \\
        \bottomrule

    \end{tabular}
    \caption{\textbf{Examples of systematic \hallu~clusters found by \ours for \pali:} We present six hallucination clusters, each for a different object—three identified by \oursllm and three by \oursopt. For each cluster, we show a sample of images and the total number of images. \textbf{For each of these images, \pali answers “yes” to “Can you see a \obj in this image?”} while the object detector reports a confidence below 0.1. None of the images actually contain the object. We also provide the text (\oursllm) and image queries (\oursopt) used for retrieval during exploration for the majority of the cluster.}
    
    \label{fig:retrieval}
    \vspace{-2mm}
\end{figure*}

Notably, the final queries $q(C)$ often differ substantially from the initial text prompt outputs (see discussion in Sec.~\ref{sec:exp-retrievalresults}), leading \oursopt to discover a larger variety of clusters compared to \oursllm %
as well as achieving a higher success rate in producing images that mislead the VLM, as shown in~\cref{tab:retrieval}. See~\cref{fig:retrieval} for examples of generated queries $q(C)$ for the objects tench, leopard, piano.

\subsection{Exploration, Exploitation and Clustering}\label{sec:explore-exploit}

Given the 50 text queries generated with \oursllm respectively 50 image queries generated with \oursopt (denoted as $q_{\text{text}}$ and $q_{\text{img}}$ in the overview of the pipeline in Fig. \ref{fig:pipeline-overview}), our goal is to find real images for which the VLM hallucinates the object \obj even though it is not visible in the image. Our pipeline is entirely source-data-free, requiring only object labels and no human supervision in subsequent stages, making it easily scalable to large datasets.

In the \textbf{exploration phase}, %
let $q_1, \dots, q_{50}$ denote the 50 text or image queries generated with \oursllm or \oursopt for a given object. 
For each query, we retrieve 20 images from ReLAION-5B using the fast CLIP kNN-index of \cite{beaumont-2022-clip-retrieval,douze2024faiss}, yielding a total of 1,000 images per object.

We then filter out all images where the object detector flags that the object %
is contained in the image. We use a very conservative threshold, as discussed above, to ensure that the object is not mistakenly contained in the image. %
While this reduces our success rate, ensuring that the object is not present is crucial for the pipeline; otherwise, the VLM does not hallucinate when replying with ``\yes''.
We evaluate our automatic object-detector-based approach against a human baseline in Section~\ref{sec:exp-retrievalresults}. Among the remaining images, we filter out all images that fail to trigger a \hallu in the VLM. %
We call this stage \textbf{exploration}, as we explore whether retrieval based on our text and image queries leads to successful hallucination on real images. Furthermore, the exploration phase aims to generate a diverse set of candidates for potential systematic hallucinations.

While the goal of the exploration phase is to provide a diverse set of candidates, the \textbf{exploitation phase} aims to achieve a high VLM acceptance rate and reveal systematic hallucinations. During this phase, we retrieve 50 images for each image candidate from the exploration phase, again using kNN-retrieval on ReLAION-5B. However, note that we are effectively retrieving a larger set of nearest neighbors as potential candidates since we filter out near duplicates using the perceptual metric DreamSim \cite{fu2023dreamsim} with a threshold of 0.9. This is necessary, as ReLAION-5B contain identical images or slight variations in resolution, viewpoint, or cropping. %
Our goal is to find semantically similar images in the sense of similar image composition and object content but not just small variations. While the exploration phase can use either text or image queries, the exploitation phase always uses image queries derived from the exploration phase. This is particularly useful since it results in a natural clustering of semantically similar images. To make this even more explicit, we further merge several of these pre-clusters, \ie images that are neighbors of the same source image, using agglomerative clustering with average linkage in the CLIP embedding space. Note that we overload notation by using \oursllm and \oursopt\ to refer to the results of the full pipeline, not just the initial query generation. All stages of the pipeline are %
summarized in Fig.~\ref{fig:pipeline-overview}.

\section{Experiments}\label{sec:exp}
For our experiments, we use a total of 380 object categories. Specifically, we include the 80 classes from COCO~\cite{lin2014coco}, 100 randomly selected object categories from Objects365~\cite{shao2019objects365} and 100 classes from ImageNet which have been used in Spurious ImageNet~\cite{neuhaus2023spurious}. Additionally, we sort the objects in OpenImages~\cite{kuznetsova2020open} based on their occurrence frequency and create four subsets, each containing 25 objects. The first subset corresponds to the most common objects in OpenImages. The second and third correspond to objects around the 10\% quantile and the median in occurrence frequency, respectively. The last subset contains the least common objects, some without a single occurrence. We use these subsets to assess the influence of an object's occurence frequency on the hallucination rate in App.~\ref{app:oi-frequency}. %
For our retrieval pipeline described in Sec.~\ref{sec:method}, we use \pali~\cite{beyer2024paligemma} and \llavanext~\cite{liu2024llavanext} in the Vicuna~\cite{peng2023vicuna} and Mistral~\cite{jiang2023mistral7b} variants as VLMs. As the large-scale image dataset, we utilize ReLAION-5B \cite{relaion5b, schuhmann2022laion} with a CLIP index for fast kNN retrieval~\cite{douze2024faiss,beaumont-2022-clip-retrieval}. See App.~\ref{app:reverse} for an experiment on the reverse task: VLM responds ``No'' although the object is visible.

\subsection{Results of \ours}\label{sec:exp-retrievalresults}

We report the results of \ours in Tab.~\ref{tab:retrieval}. %
\oursllm finds a total of $99.3$K images for \pali, $162.4$K images for \llavanext \vicuna, and $78.5$K images for \llavanext \mistral, while \oursopt obtains $221.7$K, $252.0$K, and $133.8$K. In addition to more images, \oursopt yields a larger number of clusters: $3895$/$4632$/$3229$ compared to $1892$/$3632$/$2001$ for \oursllm. 

\begin{table}[t]
\centering
\footnotesize
\setlength{\tabcolsep}{5pt} %

\begin{tabular}{ll|cc|cc}
\toprule
\makecell{Source\\Model} & \makecell{\ours\\Variant} & \makecell{Total\\Images} &  \makecell{Total\\Clusters} & \makecell{Avg Clstr\\per Object} & \makecell{Avg Imgs\\per Clstr}\\ 
\midrule
\multirow{2}{*}{PaliG} & LLM &  99.3K & 1892 & 5.0 & 52.5\\
& OPT & 221.7K & 3895 & 10.3 & 56.9 \\
\midrule
\multirow{2}{*}{\lnvic} & LLM  & 162.4K & 3632 & 9.6 & 44.7\\
& OPT & 252.0K & 4632 & 12.2 & 54.4 \\
\midrule
\multirow{2}{*}{\lnmis} & LLM & 78.5K & 2001 & 5.3 & 39.3\\
& OPT &   133.8K & 3229 & 8.5 & 41.5\\
\bottomrule
\end{tabular}

\caption{Retrieval results for \oursllm and \oursopt across \pali, \llavanext \vicuna, and \llavanext \mistral, accumulated over the 380 object categories from all datasets.}
\label{tab:retrieval}
\vspace{-1mm}
\end{table}

Some of the clusters for both approaches are shown in \cref{fig:retrieval} (and App.~\ref{app:retrieval}) along with one of the corresponding text or image queries, respectively. Considering the clusters found by \oursllm, the associations between objects and hallucinations are relatively clear and closely correspond to the text queries: a water cannon is often present in images of fireboats; the mountainsides are a natural habitat of ptarmigans; Baumkuchen is part of German Christmas traditions; and music sheets are linked to musical instruments like cellos. On the other hand, the results of \oursopt indicate that it can uncover %
hallucinations in a completely different context by moving away from the initial text queries during optimization. For these subgroups, potential causes are less obvious, e.g. at first sight, images of beads do not seem to be related in any way to the animal leopard. However, there exist so-called ``leopard beads'', i.e. beads showing a leopard pattern. While the beads in our cluster are not of this type, their existence might play the role of a confounder.

\begin{figure}[tb]
    \centering
    \includegraphics[width=0.9\columnwidth]{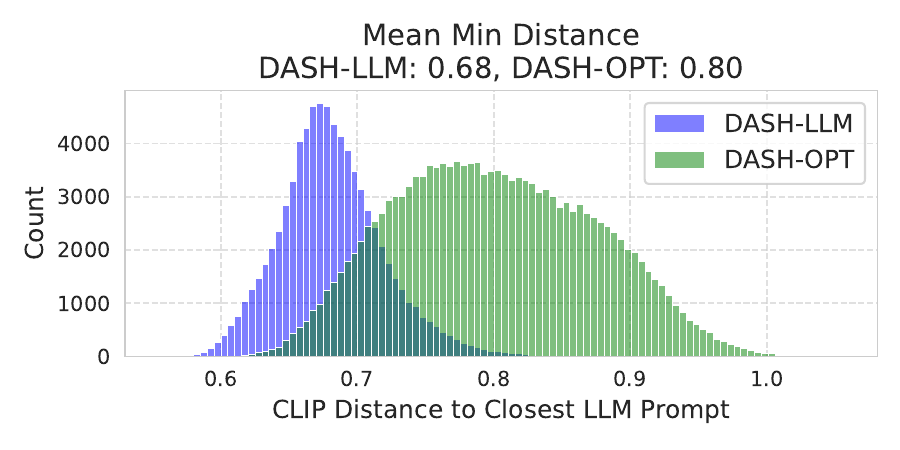}
    \vspace{-2mm}
    \caption{Histogram illustrating the minimum embedding distance from success images to the nearest LLM prompt for \oursllm and \oursopt. While both methods use these LLM prompts in their exploration stage, the image-based method is able to find unexpected hallucinations far away from the initial LLM prompts. }
    \label{fig:llm_vs_opt}
 \vspace{-1mm}
\end{figure}

\begin{figure}[htbp]
    \centering
    \footnotesize
    \setlength{\tabcolsep}{1pt} %
    \renewcommand{\arraystretch}{1} %
    \begin{tabular}{ccc|ccc}
        \multicolumn{6}{c}{\normalsize	\textbf{\oursllm} - Dam - 4 Clusters and 84 images}\\
        \toprule
        \multicolumn{3}{c|}{Cluster Size: 45} & \multicolumn{3}{c}{Cluster Size: 15} \\
        \includegraphics[width=0.075\textwidth,height=0.075\textwidth]{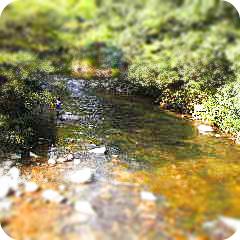} &
        \includegraphics[width=0.075\textwidth,height=0.075\textwidth]{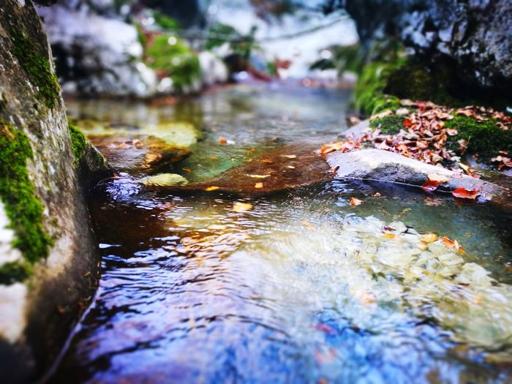} &
        \includegraphics[width=0.075\textwidth,height=0.075\textwidth]{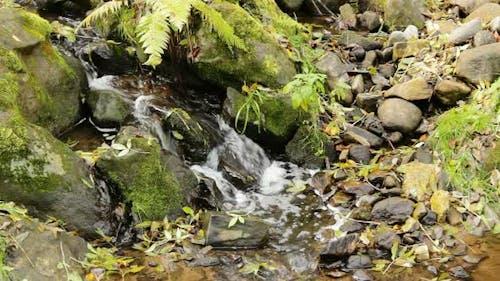} &
        \includegraphics[width=0.075\textwidth,height=0.075\textwidth]{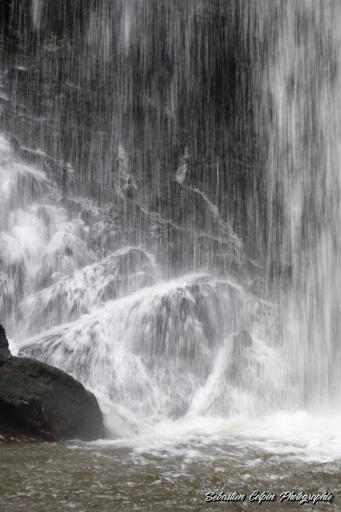} &
        \includegraphics[width=0.075\textwidth,height=0.075\textwidth]{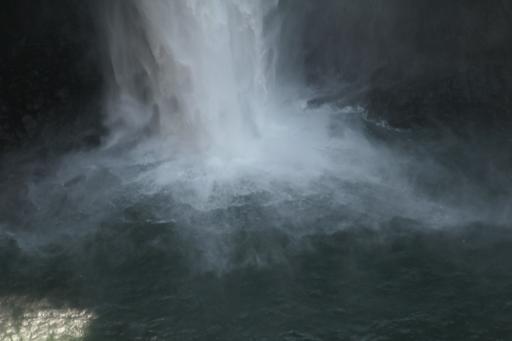} &
        \includegraphics[width=0.075\textwidth,height=0.075\textwidth]{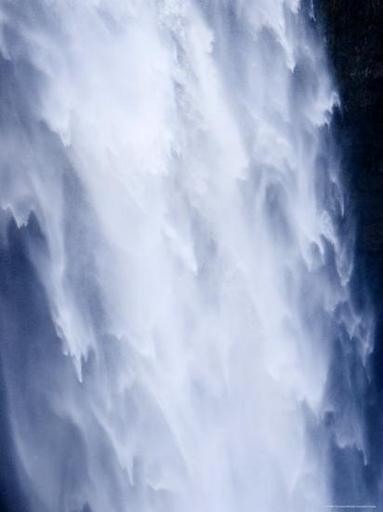} \\
        \multicolumn{3}{c|}{Cluster Size: 13} & \multicolumn{3}{c}{Cluster Size: 11}\\
        \includegraphics[width=0.075\textwidth,height=0.075\textwidth]{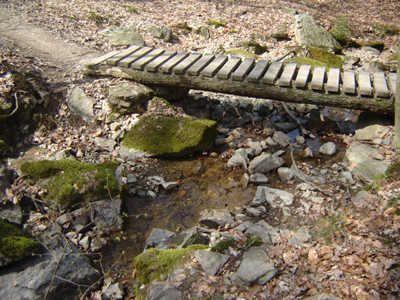} &
        \includegraphics[width=0.075\textwidth,height=0.075\textwidth]{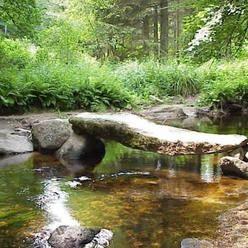} &
        \includegraphics[width=0.075\textwidth,height=0.075\textwidth]{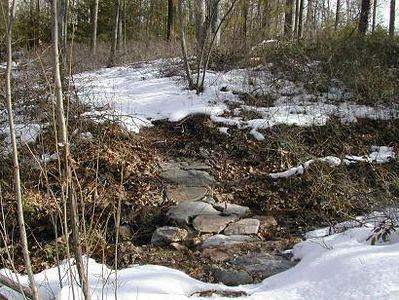} &
        \includegraphics[width=0.075\textwidth,height=0.075\textwidth]{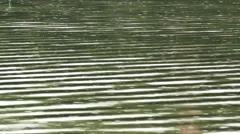} &
        \includegraphics[width=0.075\textwidth,height=0.075\textwidth]{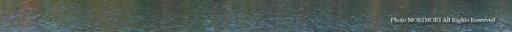} &
        \includegraphics[width=0.075\textwidth,height=0.075\textwidth]{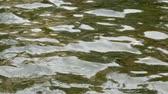} 
        \\
        \bottomrule
        \multicolumn{6}{c}{}\\
    \end{tabular}
\vspace{-1mm}
 
    \begin{tabular}{ccc|ccc}
        \multicolumn{6}{c}{\normalsize	\textbf{\oursopt}  - Dam - 10 Clusters and 186 Images}\\
        \toprule
        \multicolumn{3}{c|}{Cluster Size: 57} & \multicolumn{3}{c}{Cluster Size: 34} \\ 
        \includegraphics[width=0.075\textwidth,height=0.075\textwidth]{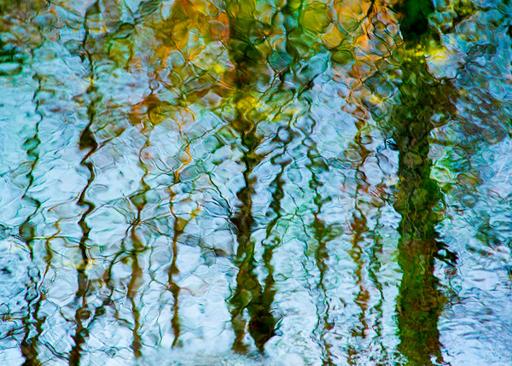} &
        \includegraphics[width=0.075\textwidth,height=0.075\textwidth]{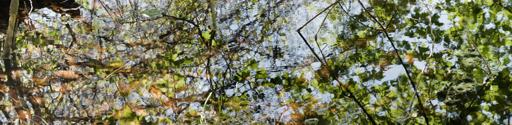} &
        \includegraphics[width=0.075\textwidth,height=0.075\textwidth]{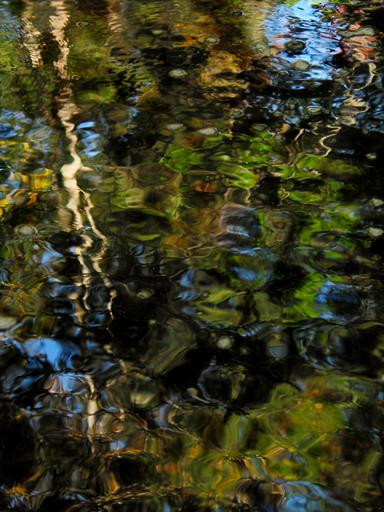} &
        \includegraphics[width=0.075\textwidth,height=0.075\textwidth]{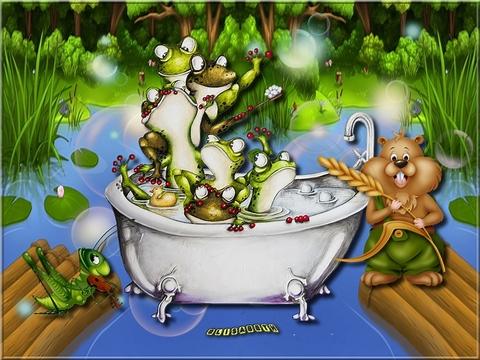} &
        \includegraphics[width=0.075\textwidth,height=0.075\textwidth]{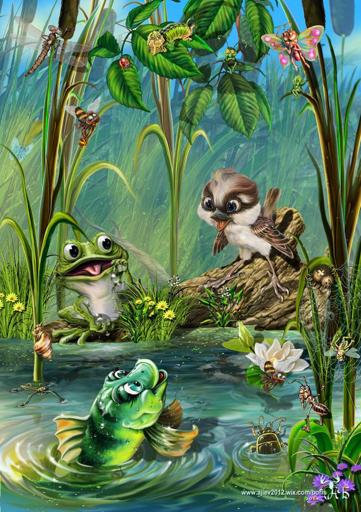} &
        \includegraphics[width=0.075\textwidth,height=0.075\textwidth]{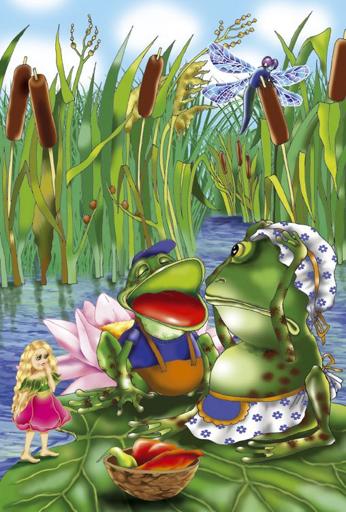} \\
    \multicolumn{3}{c|}{Cluster Size: 24} & \multicolumn{3}{c}{Cluster Size: 20}\\        
        \includegraphics[width=0.075\textwidth,height=0.075\textwidth]{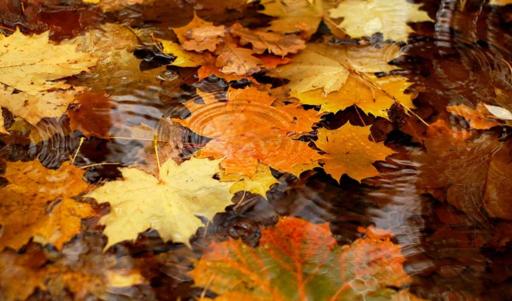} &
        \includegraphics[width=0.075\textwidth,height=0.075\textwidth]{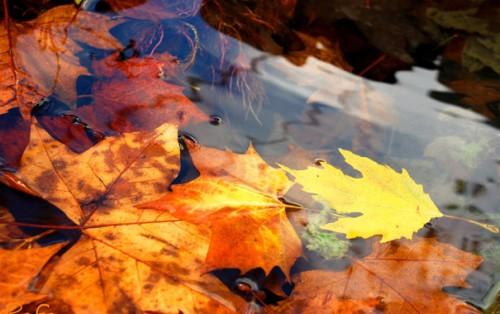} &
        \includegraphics[width=0.075\textwidth,height=0.075\textwidth]{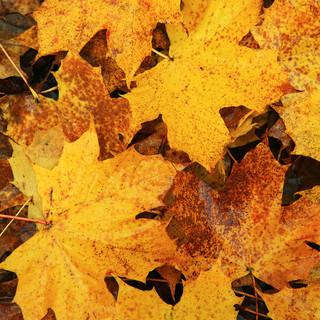} &
        \includegraphics[width=0.075\textwidth,height=0.075\textwidth]{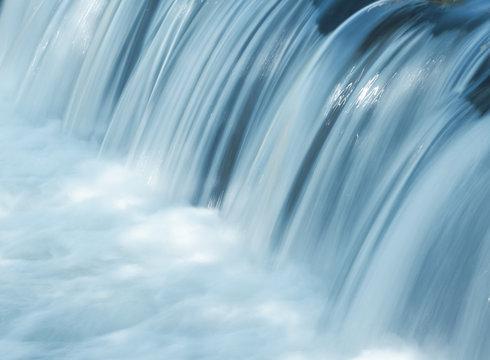} &
        \includegraphics[width=0.075\textwidth,height=0.075\textwidth]{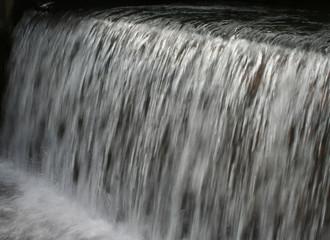} &
        \includegraphics[width=0.075\textwidth,height=0.075\textwidth]{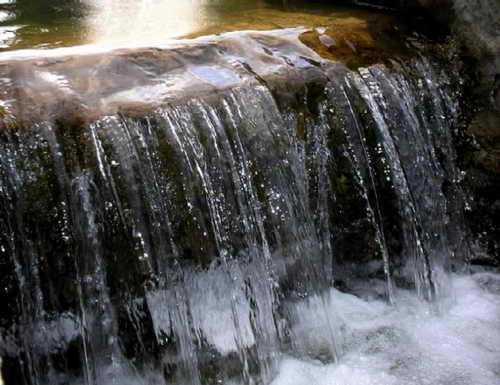} \\
        \multicolumn{3}{c|}{Cluster Size: 17} & \multicolumn{3}{c}{Cluster Size: 10} \\
        \includegraphics[width=0.075\textwidth,height=0.075\textwidth]{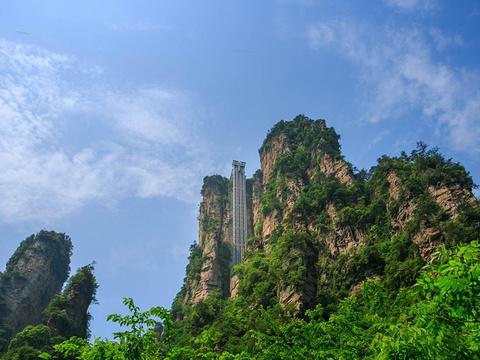} &
        \includegraphics[width=0.075\textwidth,height=0.075\textwidth]{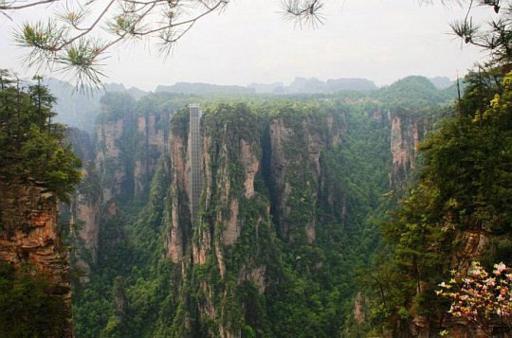} &
        \includegraphics[width=0.075\textwidth,height=0.075\textwidth]{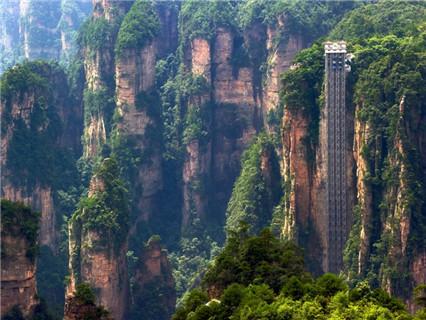} &
        \includegraphics[width=0.075\textwidth,height=0.075\textwidth]{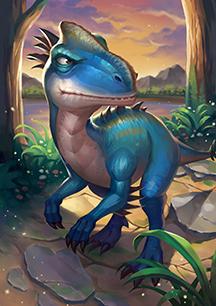} &
        \includegraphics[width=0.075\textwidth,height=0.075\textwidth]{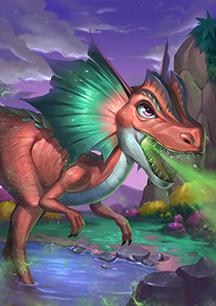} &
        \includegraphics[width=0.075\textwidth,height=0.075\textwidth]{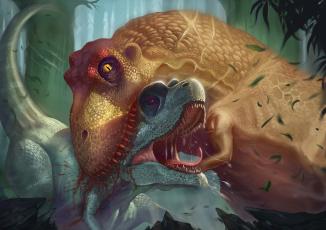} \\
        \multicolumn{3}{c|}{Cluster Size: 7 } & \multicolumn{3}{c}{Cluster Size: 6}\\
        \includegraphics[width=0.075\textwidth,height=0.075\textwidth]{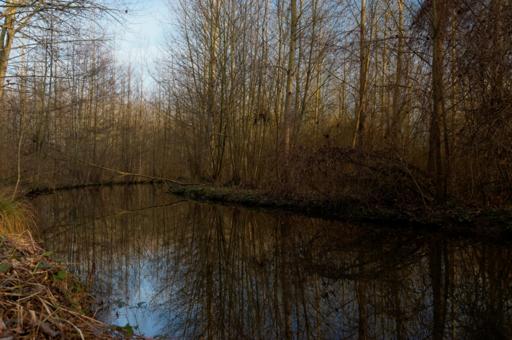} &
        \includegraphics[width=0.075\textwidth,height=0.075\textwidth]{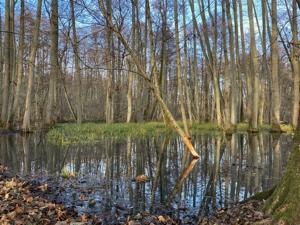} &
        \includegraphics[width=0.075\textwidth,height=0.075\textwidth]{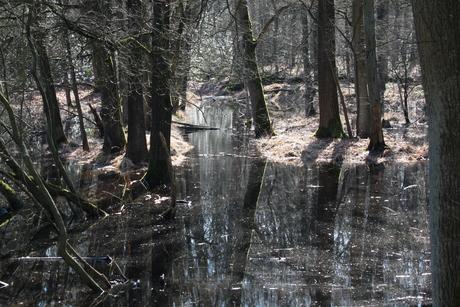} &
        \includegraphics[width=0.075\textwidth,height=0.075\textwidth]{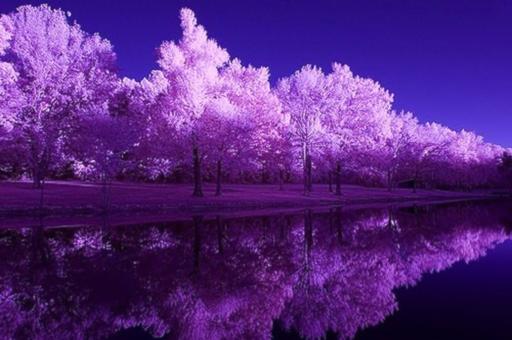} &
        \includegraphics[width=0.075\textwidth,height=0.075\textwidth]{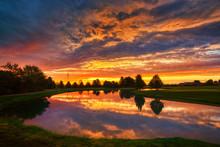} &
        \includegraphics[width=0.075\textwidth,height=0.075\textwidth]{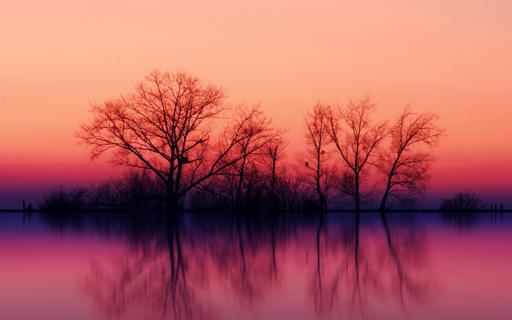} \\
        \multicolumn{3}{c|}{Cluster Size: 6} & \multicolumn{3}{c}{Cluster Size: 5}\\
        \includegraphics[width=0.075\textwidth,height=0.075\textwidth]{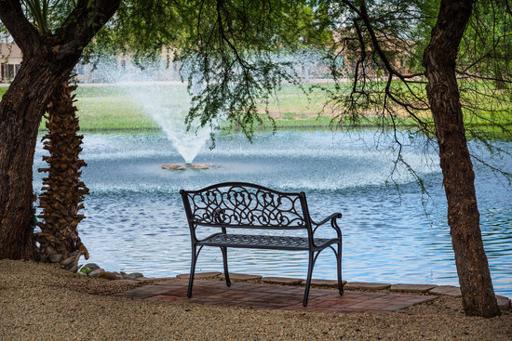} &
        \includegraphics[width=0.075\textwidth,height=0.075\textwidth]{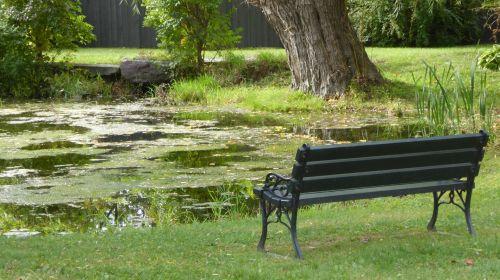} &
        \includegraphics[width=0.075\textwidth,height=0.075\textwidth]{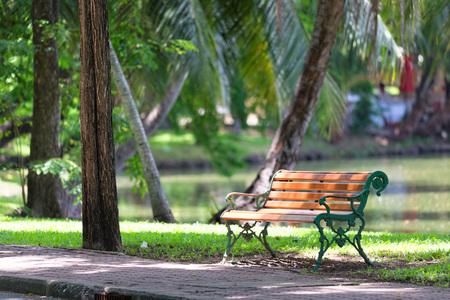} &
        \includegraphics[width=0.075\textwidth,height=0.075\textwidth]{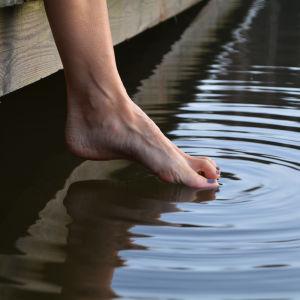} &
        \includegraphics[width=0.075\textwidth,height=0.075\textwidth]{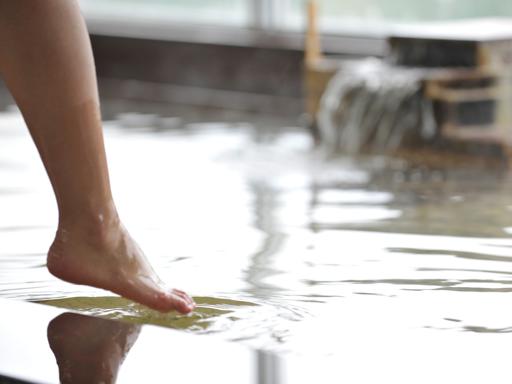} &
        \includegraphics[width=0.075\textwidth,height=0.075\textwidth]{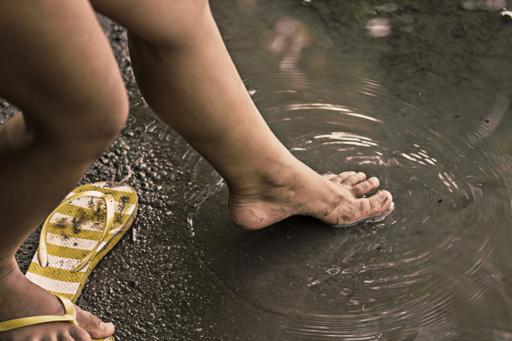} \\
        \bottomrule
    \end{tabular}
    \vspace{-1mm}
    \caption{All clusters for \oursllm and \oursopt for the object 'Dam' using \llavanext \vicuna. \oursopt identifies a larger total number of clusters and images, capturing a broader diversity of visuals. This demonstrates that \oursopt can uncover unexpected systematic hallucination patterns, such as cartoon frogs and dinosaurs, orange leaves, bare feet, or a park bench, whereas \oursllm tends to highlight failure modes more directly linked to the object, such as water associated with 'Dam'. }
    \vspace{-2mm}
    \label{fig:all_cluster}
\end{figure}

We further validate the larger exploration range of \oursopt over \oursllm in \cref{fig:llm_vs_opt}, where we illustrate the CLIP distance between the image embedding and the text embedding of the closest original text prompt. While both LLM and OPT use the same text prompts, the additional optimization of the OPT variant allows it to find images that are further away.
This can also be seen in \cref{fig:all_cluster}, where we visualize all clusters detected for the object Dam using \llavanext \vicuna. \oursllm results in 4 clusters with a total of 84 images, while \oursopt finds 10 clusters comprising 186 images. The clusters resulting from LLM queries represent natural waterfalls and water surfaces and are also detected using OPT queries. However, beyond those, \oursopt also surfaces a range of more diverse and unexpected patterns, including different colors (leaves/sunsets) and distribution shifts (comic frogs/comic dragons).

\noindent\textbf{%
Advantage of a larger retrieval pool:} In \cref{fig:ours_vs_ref}, we compare hallucinations found by \ours\ to their nearest neighbors in the curated dataset, which also contains the object. Note that, first, some of the objects (\eg, rubber boot) or types of images (\eg, maps) are not contained in the smaller reference datasets. Second, the most similar images that are contained might not be specific enough to fool the VLM, even when they are similar in object, color, and composition. For example, while Objects365 contains trees, only ``Baobab trees'' are detected as ``sausage''.  This emphasizes the necessity of a large dataset for searching hallucinations.

\noindent\textbf{Human Verification:} To validate the object detector in \ours, we use human supervision to verify the absence of the object. In particular, for all 380 objects found for \pali with \oursopt, we randomly select 10 images. We then label them as ``yes'' if the object is visible, ``no'' if it is absent, and ``ambiguous'' for corner cases. For example, it can be difficult to assess whether a close-up photo of a flat surface is a ``dessert table'' or a ``coffee table''. Across all images, we find that 5.2\% contain the object and 7.8\% are ambiguous.
For comparison, we perform the same annotation on the corresponding subset of POPE, i.e. image/question pairs where \pali responds ``\yes'', but the COCO ground truth indicates ``\no''. Among these 137 alleged false positives, 25.5\% contain the object, and 22.6\% are marked as ambiguous (examples in App.~\ref{app:pope-errors}). %
This demonstrates that our conservative threshold for the object detector yields a reliable automatic pipeline with less errors compared to POPE. 

\noindent\textbf{Transfer across prompts:} We check the influence of the type of the question on the evaluation of the \hallu s found by \ours by testing 11 prompts (``Is \obj in the image?'', ``Does this image contain a \obj?'', \ldots)  on \pali, \lnvicuna, and \lnmistral (results in App.~\ref{app:different-prompts}). \lnvicuna has an average ``\yes'' rate of $82.3\%\pm 6.7\%$, similarly \lnmistral $78.9\% \pm 7.0\%$ showing that the prompt has only a minor influence. For \pali, which was trained on this task on OpenImages, the ``\yes''-rate drops to $71.6\% \pm 18.7\%$ with higher variance as prompts similar to the training prompt (“Is OBJ in the image?”, 31\% ``\yes'') show a lower transfer. %

\subsection{Transfer to unseen VLMs}
\begin{table*}[htbp]
\centering
\footnotesize
\begin{tabular}{ll|c|ccc|cc|cc|c}
\toprule
\multicolumn{2}{l|}{\textbf{VLM Type}}   & \pali & \multicolumn{3}{c|}{---\llavanext---} & \multicolumn{2}{c|}{---\prismatic---} & \multicolumn{2}{c|}{---\qwen-VL---} &	\llama 3.2-VL \\

\multicolumn{2}{l|}{\textbf{Vision Encoder}} & \siglip & 	\clip & \clip &	\clip & \clip & \siglip & Custom & Custom & Custom\\

\multicolumn{2}{l|}{\textbf{LLM}} & \gemma & 	\vicuna & \mistral & \llama 3.0  & \vicuna & \vicuna & \qwen-7B & \qwen-72B &	\llama 3.1\\
\midrule
\multirow{2}{*}{PaliG}                 & LLM & -    & 0.49 & 0.31 & 0.26 & 0.60 & 0.44 & 0.18 & 0.18 & 0.09 \\
                                       & OPT & -    & 0.43 & 0.23 & 0.22 & 0.49 & 0.34 & 0.17 & 0.15 & 0.10 \\
\midrule
\multirow{2}{*}{\lnvic} & LLM & 0.34 & -    & 0.39 & 0.30 & 0.64 & 0.44 & 0.18 & 0.16 & 0.10 \\
                                       & OPT & 0.30 & -    & 0.33 & 0.27 & 0.59 & 0.38 & 0.15 & 0.13 & 0.09 \\
\midrule
\multirow{2}{*}{\lnmis} & LLM & 0.39 & 0.67 & -    & 0.42 & 0.67 & 0.51 & 0.25 & 0.23 & 0.14 \\
                                       & OPT & 0.35 & 0.66 & -    & 0.41 & 0.63 & 0.46 & 0.24 & 0.21 & 0.15 \\
\midrule
\multicolumn{2}{l|}{Average transfer}         & 0.35 & 0.56 & 0.31 & 0.31 & 0.60 & 0.43 & 0.19 & 0.18 & 0.11 \\
\doublemidrule
\multicolumn{2}{l|}{TPR-ICO}       & 0.81 & 0.87 & 0.83 & 0.81 & 0.83 & 0.77 & 0.85 & 0.85 & 0.80 \\
\bottomrule
\end{tabular}

\caption{\textbf{Transfer of \ours images (rows) to different VLMs (columns)} 
Different LLM backbones(\llavanext) and different vision encoders (\prismatic) have a significant impact on the vulnerability to \hallu s, but the LLM size (\qwen-VL) shows only a small effect.
TPR-\underline{ICO} is the true positive rate calculated on ground-truth validation data from \underline{I}mageNet, \underline{C}OCO, and \underline{O}bjects365 corresponding to the employed object categories. }
\label{tab:transfer}

\end{table*}

\ours searches specifically for images causing hallucinations for a target model. In this section, we investigate how the found systematic hallucinations transfer to unseen VLMs.  \cref{tab:transfer} reports the transfer rate, defined as the proportion of images found by \ours (for \pali, \llavanext\ \vicuna, and \llavanext\ \mistral) that trigger a \hallu~ in the model. %
We use these transfer rates to quantify the impact of the LLM backbone, vision encoder, and model scale on \hallu s. %
In addition, to quantify the detection-hallucination trade-off, we evaluate the true positive rate on a subset of images from COCO, Object365, and ImageNet showing the object (see App.~\ref{app:tpr}) which we call TPR-ICO. 

\noindent\textbf{Influence of LLM backbone:}
We examine the LLM backbone's influence on the model's vulnerability to systematic hallucinations by considering three versions of \llavanext, based on the LLMs \vicuna, \mistral, and \llama. Apart from the LLM, these three share the same vision backbone, architecture, and training procedure. Thus, differences in their robustness to hallucinations can be attributed to the LLM. On images found for \pali, the \vicuna variant has the highest transfer rates (\oursllm $49\%$ and \oursopt $43\%$), followed by \mistral ($31\%$ and $23\%$) and \llama ($26\%$ and $22\%$). The TPR-ICO indicates (\vicuna $87\%$, \mistral $83\%$, \llama $81\%$) that the models with higher transfer rates in general reply with ``\yes'' more frequently. Overall, one can conclude that the LLM backbone significantly impacts the vulnerability to \hallu s.

\noindent\textbf{Influence of vision encoder:}
Similar to the LLM backbone, we compare three Prismatic~\cite{karamcheti2024prismatic} models that share the same LLM but use different ViT-L~\cite{dosovitskiy2020image} vision encoders, namely \clip~\cite{radford2021clip} and \siglip~\cite{zhai2023siglip}. %
On average, \clip %
results in a higher transfer rate of $60\%$ while the \siglip variant is only fooled by $43\%$ of the images, showing that the choice of the vision encoder has a major impact. However, smaller hallucination transfer rates also come with a smaller TPR-ICO, \ie \siglip ($77\%$) is less likely to respond with ``\yes'', even on data containing the object, than \clip ($83\%$).%

\noindent\textbf{Model size:} We evaluate \qwen-VL~\cite{Qwen2VL} both with \qwen-7B~\cite{yang2024qwen2} and 72B as language backbone. Overall, the size of the used LLM seems to have only a small effect on the VLM's vulnerability to hallucinations  ( $18\%$ to $19\%$ ) but as both models have the same TPR-ICO of 85\%, model scaling can improve the detection-hallucination tradeoff.

\begin{figure}[tb]
    \centering
    \footnotesize
    \setlength{\tabcolsep}{1pt} %
    \renewcommand{\arraystretch}{1} %
    \begin{tabular}{ccc|ccc}
    \toprule
    \multicolumn{3}{c|}{coil spring} &
    \multicolumn{3}{c}{shipping box} \\
    \includegraphics[width=0.075\textwidth,height=0.075\textwidth]{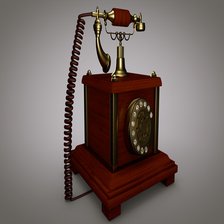} &
    \includegraphics[width=0.075\textwidth,height=0.075\textwidth]{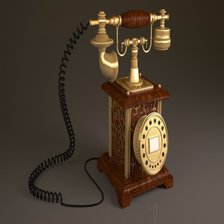} &
    \includegraphics[width=0.075\textwidth,height=0.075\textwidth]{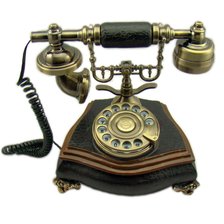} &
    \includegraphics[width=0.075\textwidth,height=0.075\textwidth]{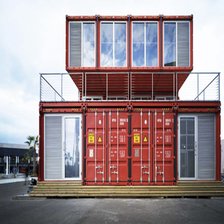} &
    \includegraphics[width=0.075\textwidth,height=0.075\textwidth]{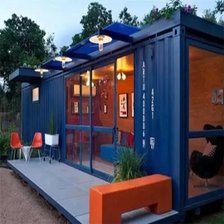} &
    \includegraphics[width=0.075\textwidth,height=0.075\textwidth]{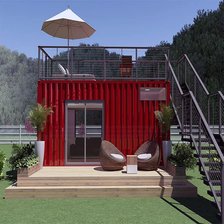} \\
    \multicolumn{3}{c|}{balance beam} &
    \multicolumn{3}{c}{vase} 
    \\
    \includegraphics[width=0.075\textwidth,height=0.075\textwidth]{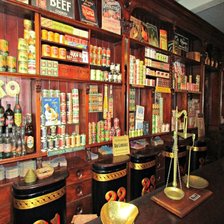} &
    \includegraphics[width=0.075\textwidth,height=0.075\textwidth]{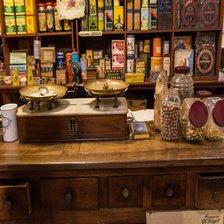} &
    \includegraphics[width=0.075\textwidth,height=0.075\textwidth]{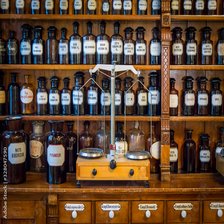} &
    \includegraphics[width=0.075\textwidth,height=0.075\textwidth]{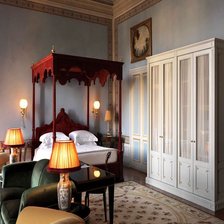} &
    \includegraphics[width=0.075\textwidth,height=0.075\textwidth]{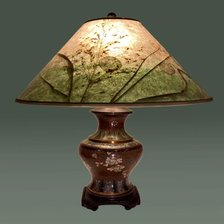} &
    \includegraphics[width=0.075\textwidth,height=0.075\textwidth]{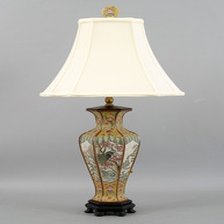} 
    \\
    \multicolumn{3}{c|}{watch} &
    \multicolumn{3}{c}{postcard} 
    \\
    \includegraphics[width=0.075\textwidth,height=0.075\textwidth]{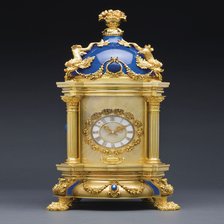} &
    \includegraphics[width=0.075\textwidth,height=0.075\textwidth]{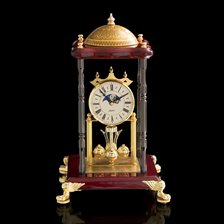} &
    \includegraphics[width=0.075\textwidth,height=0.075\textwidth]{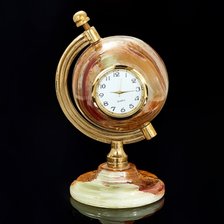} &
    \includegraphics[width=0.075\textwidth,height=0.075\textwidth]{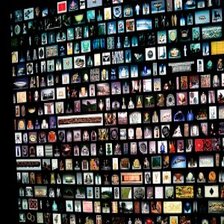} &
    \includegraphics[width=0.075\textwidth,height=0.075\textwidth]{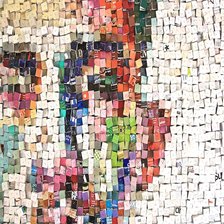} &
    \includegraphics[width=0.075\textwidth,height=0.075\textwidth]{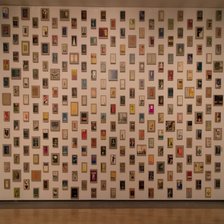} 
    \\    
    \bottomrule     
    \end{tabular}
    \caption{\textbf{Object hallucination benchmark \ours-B:} examples from the negative set of \oursb (images and object label) where GPT-4o-mini, the best scoring model on \oursb (see Tab.~\ref{app:benchmark}) hallucinates the object even though it is not present in the image.%
    }
    \vspace{-2mm}
    \label{fig:qwen_llama_intersection}
\end{figure}

\subsection{\ours-B: Object Hallucination Benchmark}
As the popular POPE \cite{li2023evaluating} benchmark for object hallucinations seems saturated and contains significant label noise, a novel hard object hallucination benchmark is needed to measure further progress. For this purpose we use all images found by \ours which transfer to both \qwen-VL and Llama 3.2-VL (see App.~\ref{app:benchmark} details). We verify those images for a selection of 70 objects by consensus of two human labelers and limit the minimal and maximal number of images to 3 and 50, respectively. This results in 1341 images, see  \cref{fig:qwen_llama_intersection} for examples.
We use this as our set of negatives and add the same amount of images containing the objects. As performance measure, we use accuracy over all 2642 images. Tab.~\ref{tab:dashb}, contains the results for four models not used in the generation of \ours-B. Given the label noise (see App.~\ref{app:pope-errors}), the high true negative rates (TNR) on POPE ($96.0\%\pm1.2\%$) suggest that the benchmark is saturated and provides limited insight regarding \hallu s. In contrast, the models exhibit a significantly lower TNR on \oursb ($48.6\%\pm10.1\%$). We show results on \oursb for more models in Tab.~\ref{app:benchmark}.

\begin{table}
    \centering
    \setlength{\tabcolsep}{2pt}
    \footnotesize
    \begin{tabular}{c||c|c|c|c}
        \textbf{Models} & PaliG2-3B & Ovis2-8B & LLaVa-OneVision & 4o-mini\\ 
        \midrule 
        \oursb Acc. & $68.9\%$ & $71.4\%$& $75.1\%$ & $86.3\%$ \\
        \midrule
        \oursb TNR & $40.9\%$ & $44.8\%$ & $60.1\%$ & 76.7\%\\
        POPE TNR & $97.3\%$ & $94.9\%$ &$95.8\%$ & -\footnotemark[2]
    \end{tabular}
    \caption{\textbf{Object Hallucination Benchmark \oursb:}
    Compared to POPE, \ours-B is not saturated as demonstrated by the substantially lower true negative rates (TNR). More results in Tab.~\ref{tab:app-benchmark}.
   }
    \vspace{-2mm}    
    \label{tab:dashb}
\end{table}

\footnotetext[2]{In our POPE evaluation, GPT-4o-mini only provided valid replies for $77.5\%$ of the images and achieved a TNR of $92.7\%$ among those.} 

\subsection{Fine-Tuning on \ours}
We test the usage of the images found by \ours for fine-tuning \pali in a small proof-of-concept (details in App.~\ref{app:mitigation}): First, we merge the results of \oursllm and \oursopt for \pali and ensure that none of the images of \oursb are part of this train set. 
Per object, we sample $200$ random images from this set and additionally $400$ positive samples, i.e. images that contain the object. During fine-tuning, the models is trained to output ``\no'' on \ours images and predict ``\yes'' on the positive samples (see App.~\ref{app:mitigation} for more details on the fine-tuning setup).

\cref{tab:ft} contains the results after finetuning: Accuracy is significantly improved on \oursb ($+11.6\%$) due to an increase in TNR and TPR (see App.~\ref{app:mitigation}) and also increases on Amber Existence~\cite{wang2023llm} ($+2.2\%$) as well as R-Bench~\cite{rbench} ($+0.3\%$). However, we observe a small decrease of accuracy on POPE. We also evaluate two captioning tasks (Flickr30k, COCO) and two VQA tasks (TextVQA, VQAv2) and report the averaged results. The minor performance drop (captioning $-1.8$, VQA $-0.9\%$) is expected as we are fine-tuning the model on a different task. Generally, the \ours data should be integrated into a curriculum learning scheme, as one of several tasks, which is out of scope for this work.

\begin{table}
    \centering
    \setlength{\tabcolsep}{2pt}
    \footnotesize
    \begin{tabular}{c||c|c|c|c||c|c}
         & \ours-B & Amber Ex. & R-Bench & POPE & Caption & VQA\\
        \midrule
        PaliG & $56.4\%$ & $93.2\%$ & $79.9\%$ & $\mathbf{87.2\%}$ & $\mathbf{101.0}$ & $\mathbf{70.4\%}$\\
        +ft  & $\mathbf{68.0\%}$ & $\mathbf{95.4\%}$ & $\mathbf{80.2\%}$ & $86.4\%$ & $99.2$ & $69.5\%$
        
    \end{tabular}
    \caption{\textbf{Fine-tuning on \ours:} 
    Our fine-tuning strategy improves performance on \oursb, Amber Existence and R-Bench.} 
    \label{tab:ft}
    \vspace{-2mm}
\end{table}

\section{Conclusion and Limitations} 

We have demonstrated that \ours is an effective, automatic pipeline for identifying systematic hallucinations in VLMs. Contrary to the belief that object hallucinations are no longer an issue, we find that they persist extensively when using VLMs in an open world scenario. \yn{To address this, we propose \ours-B for a more rigorous assessment of these errors.} Our initial experiments on mitigation suggest that \ours could be a valuable addition to the training pipeline of future VLMs.

\noindent\textbf{Limitations:} we note that achieving exhaustive coverage of all systematic hallucinations with \ours is not possible, as this would require a fully exhaustive approach. Even with such an approach, and despite ReLAION-5B providing significant coverage of natural images, some images remain underrepresented. As a result, even when we identify an image where the VLM hallucinates, there may not be enough semantically similar images in ReLAION-5B to consider it a systematic hallucination. 
For the most advanced VLMs, our conservative threshold for the object detector could pose a limitation, potentially.

\section*{Acknowledgments}
We are grateful for support by the DFG, Project number
390727645, and the Carl Zeiss Foundation, project “Certification and Foundations of Safe Machine Learning Systems in Healthcare” and thank the IMPRS-IS for supporting YN.

{
    \small
    \bibliographystyle{ieeenat_fullname}
    \bibliography{main}
}

\clearpage
\appendix

 \section{Overview}
We give an overview over the contents of the Appendix.

\begin{itemize}
    \item In \cref{sec:app_llm_prompt}, we present additional details about the creation of initial LLM queries for \oursllm, including the prompts used to create the queries with \llama.
    \item In \cref{sec:app_diffusion}, we break down the optimization of the \oursopt image queries in the latent space of the distilled SDXL model.
    \item In \cref{sec:app_exploration_exploitation_retrieval}, we give additional details about the ReLAION exploration/exploitation retrieval.
    \item In \cref{app:oi-frequency}, we investigate the influence of an object's occurence frequency on hallucination rates.
    \item In \cref{app:retrieval}, we present additional results for \ours on \pali, \llavanext \vicuna and \mistral. In \cref{sec:app_ours_vs_reference}, we also present \ours results on ReLAION next to the most similar images from COCO and Objects365.
    \item In \cref{app:objectdetector}, we further explore the performance of the object detector in \ours to filter out images containing the object.

    \item In \cref{app:pope-errors}, we show examples of COCO annotation errors and discuss their effect on the POPE benchmark.
    \item In \cref{app:transfer}, we present extended results about the transfer between \ours images to other VLMs and also present information about all VLMs used in the paper in \cref{sec:app_vlm_models} and the true positive rate calculation in \cref{app:tpr}
    \item In \cref{app:benchmark}, we describe the image selection process and discuss different metrics for our proposed benchmark \oursb. Additionally, we report more results on a range of VLMs.
    \item In \cref{app:mitigation}, we give further details about the mitigation finetuning using \ours.
    \item In \cref{app:reverse}, we provide a proof of concept for a possible application of our pipeline to the reverse task and discuss problems.
    \item In \cref{app:different-prompts}, we examine the generalization of \ours results to different prompts than the one used in our experiments.
    
\end{itemize}

\section{\oursllm Prompt}\label{sec:app_llm_prompt}
The prompts supplied to \llama-3.1-70B~\cite{dubey2024llama3herdmodels} to create the queries for \oursllm\ are given in ~\cref{fig:llm-prompt} and~\ref{fig:llm-prompt_followup}. We also use the same queries to initialize the generation of the image queries in \oursopt. To generate the queries, we use the system prompt provided in Figure~\ref{fig:llm-prompt}. We then pass the input ``object: \obj'' to the LLM, which generates an initial list of 50 queries. Since we noticed that these initial queries can sometimes contain references to the object or duplicates, we use a simplified version of chain-of-thought prompting~\cite{wei2022chain}. After the LLM generates the initial list of 50 queries, we pass the follow-up prompt provided in Figure~\ref{fig:llm-prompt_followup} to the model, which responds with an updated list of 50 queries.

\begin{figure*}[p] %
  \centering
  \begin{minipage}{\textwidth}
    \begin{lstlisting}
As an AI language model assistant, your task is to provide descriptive captions for images showing spurious features.

A spurious feature is a visual element that frequently co-occurs with a given object in images and may cause AI models to incorrectly recognize the object, even when it is not present.

Task Overview:

You will be given:
- An object.

Your job is to:

1. Think of potential spurious features: Identify objects, scenes, or elements that frequently co-occur with the given object in images. These should not include any parts or components of the object itself.

2. Generate 50 unique and diverse prompts describing images that contain only these spurious features, without including the object itself or any of its parts.

Important Guidelines:

- Do Not Mention the Object Name or Any Part of It: Avoid any direct or indirect references to the object name. If the object name is a composite or compound word, do not include any part of the object name in the prompts. For example, if the object is "firetruck," do not use "fire" or "truck" in the prompts.

- Do Not Mention Parts of the Object: Do not include any parts or components of the object in the prompts. For example, if the object is "mountainbike," do not use "handlebar," "gear shift," or "saddle" in the prompts.

- Do Not Include the Object Name in Written Text: Do not create prompts that refer to written text containing the object name or any part of it. For example, avoid descriptions like "a sign that says 'hummingbird'."

- Focus on Spurious Features: Use features that are likely correlated with the object due to frequent co-occurrence in images.

- Combining Elements: You may combine elements if they logically make sense to appear together in one image. Do not combine elements unlikely to co-occur.

- Ensure Diversity: Each prompt should be unique and cover different aspects of the spurious features.

- Avoid Repetition: Do not repeat prompts or make minor variations of the same prompt.

- Style and Detail: Write clear, creative, and descriptive prompts. Keep each prompt concise.

- Language and Grammar: Use proper grammar and spelling.

- Content Restrictions: Do not include offensive, sensitive, or inappropriate content.

- Avoid Bias: Ensure prompts are inclusive and free from cultural, gender, or racial bias.

- Verification: Before submitting, review the prompts to ensure they comply with all guidelines.


    \end{lstlisting}
  \end{minipage}
  \caption{\oursllm prompt for generating the text queries (1/3)}
  \label{fig:llm-prompt}
\end{figure*}

\begin{figure*}[p] %
  \ContinuedFloat
  \centering
  \begin{minipage}{\textwidth}
    \begin{lstlisting}

Examples:

For the object "hummingbird":

- Correct Prompts:
  - "Close-up of a bird feeder hanging in a lush garden."
  - "A garden filled with vibrant red flowers."
  - "Green foliage glistening after a rainfall."
  - "A bird feeder surrounded by blooming plants."
  - "Red tubular flowers swaying in the breeze."

- Incorrect Prompts (Do Not Use):
  - "A hummingbird hovering near a flower."
  - "Close-up of a hummingbird's wings in motion."
  - "A small bird with iridescent feathers perched on a branch."
  - "A sign with the word 'hummingbird' in a botanical garden."

For the object "firetruck":

- Correct Prompts:
  - "A fire station with bright red doors."
  - "Close-up of a spinning emergency siren light."
  - "Firefighters conducting a training drill."
  - "A tall ladder reaching up the side of a building."
  - "Protective gear hanging neatly in a station locker room."

- Incorrect Prompts (Do Not Use):
  - "A bright red firetruck parked on the street."
  - "Children waving at a passing firetruck."
  - "A sign that reads 'Fire Station No. 1'."
  - "A red truck with emergency equipment."
  - Using the words "fire" or "truck" in the prompts.

For the object "mountainbike":

- Correct Prompts:
  - "A winding trail cutting through a dense forest."
  - "A helmet resting on a tree stump beside a path."
  - "Sunlight filtering through trees along a forest trail."
  - "A backpack leaning against a wooden signpost on a hillside."
  - "A group of friends hiking through mountainous terrain."

- Incorrect Prompts (Do Not Use):
  - "A mountainbike leaning against a tree."
  - "Close-up of a mountainbike's gears."
  - "A cyclist adjusting the saddle of a mountainbike."
  - "A sign that says 'Mountainbike Trail Ahead'."
  - Using the words "mountain" or "bike" in the prompts.
  - Mentioning parts like "handlebar," "gear shift," or "saddle."


    \end{lstlisting}
  \end{minipage}
  \caption{\oursllm prompt for generating the text queries (2/3)}
\end{figure*}

\begin{figure*}[p] %
\ContinuedFloat
  \centering
  \begin{minipage}{\textwidth}
    \begin{lstlisting}

Formatting Instructions:

- Start each prompt on a new line, numbered sequentially from 1 to 50.

- The format should be:

  1: <prompt_1>
  2: <prompt_2>
  3: <prompt_3>
  ...
  50: <prompt_50>

User Input Format:

The user will provide the object in the following format:

object: <object name>

Your Response:

- Return exactly 50 prompts per user request.

- Ensure that the last line of your response starts with:

  50: <prompt_50>

- Under no circumstances should you include any content in your response other than the 50 prompts. Do not include explanations, apologies, or any additional text.

Summary:

- Do not mention the object name or any part of it. If the object name is a composite or compound word, do not include any part of it in the prompts.

- Do not mention parts or components of the object.

- Do not create prompts that refer to written text containing the object name or any part of it.

- Focus on spurious features that frequently co-occur with the object.

- You may combine elements if they logically co-occur in an image.

- Ensure diversity and uniqueness in the prompts.

- Use proper language and avoid any inappropriate content.

- Review all prompts for compliance before submitting.

- Under no circumstances should you include any content in your response other than the 50 prompts. Do not include explanations, apologies, or any additional text.

Remember, the goal is to create prompts that could lead an AI model to falsely recognize the object due to the presence of spurious features, even though the object itself is not present in the images.
    \end{lstlisting}
  \end{minipage}
  \caption{\oursllm prompt for generating the text queries (3/3)}
\end{figure*}

\begin{figure*}[p] %
  \centering
  \begin{minipage}{\textwidth}
    \begin{lstlisting}
Please review the list of prompts you previously generated and check for any mistakes or deviations from the guidelines. Identify any prompts that do not fully comply with the instructions. Then, generate a new list of 50 prompts that strictly adhere to all the guidelines provided.

Important Guidelines:

- Do not mention the object name or any part of it. If the object name is a composite or compound word, do not include any part of the object name in the prompts.
- Do not mention parts or components of the object.
- Do not create prompts that refer to written text containing the object name or any part of it.
- Focus on spurious features that frequently co-occur with the object.
- You may combine elements if they logically co-occur in an image.
- Ensure diversity and uniqueness in the prompts.
- Use proper language and avoid any inappropriate content.
- Review all prompts for compliance before submitting.
- Under no circumstances should you include any content in your response other than the 50 prompts. Do not include explanations, apologies, or any additional text.

Formatting Instructions:

- Start each prompt on a new line, numbered sequentially from 1 to 50.
- The format should be:

  1: <prompt_1>
  2: <prompt_2>
  3: <prompt_3>
  ...
  50: <prompt_50>

- Ensure that the last line of your response starts with:

  50: <prompt_50>

Remember, your goal is to create prompts that could lead an AI model to falsely recognize the object due to the presence of spurious features, even though the object itself is not present in the images.

Now, generate the corrected list of 50 prompts.
    \end{lstlisting}
  \end{minipage}
  \caption{\oursllm follow-up prompt for generating the text queries}
  \label{fig:llm-prompt_followup}
\end{figure*}

\section{\oursopt Optimization}
\begin{figure*}[phtb]
    \centering
    \small
    \setlength{\tabcolsep}{1pt} %
    \renewcommand{\arraystretch}{1} %
    \begin{tabular}{cccccc}
    \toprule
     Initialization  & Step 5 & Step 10 & Step 15  & Step 20 & Step 25 \\
    \midrule
    \multicolumn{6}{c}{"Leopard" - Prompt: "A close-up of a rock's cracks and fissures."} \\
    \makecell{VLM: "no"\\$p_\text{yes}: 0.04$\\$p_\text{det}: 0.00$} &
   \makecell{VLM: "no"\\$p_\text{yes}: 0.05$\\$p_\text{det}: 0.00$} &
    \makecell{VLM: "no"\\$p_\text{yes}: 0.06$\\$p_\text{det}: 0.00$} &
    \makecell{VLM: "yes"\\$p_\text{yes}: 0.75$\\$p_\text{det}: 0.00$} &
    \makecell{VLM: "yes"\\$p_\text{yes}: 0.77$\\$p_\text{det}: 0.00$} &
    \makecell{VLM: "yes"\\$p_\text{yes}: \mathbf{0.79}$\\$p_\text{det}: 0.00$} 
    \\
    \includegraphics[width=0.12\textwidth,height=0.12\textwidth]{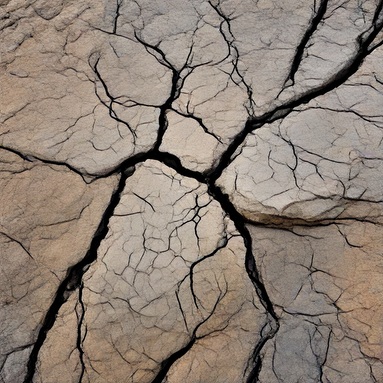} & 
   \includegraphics[width=0.12\textwidth,height=0.12\textwidth]{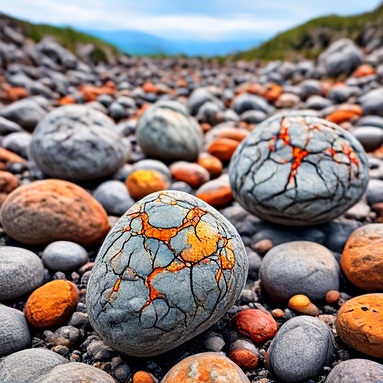} & 
   \includegraphics[width=0.12\textwidth,height=0.12\textwidth]{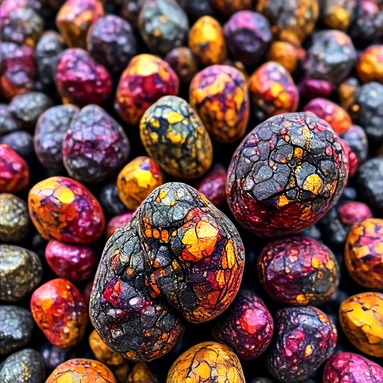} & 
   \includegraphics[width=0.12\textwidth,height=0.12\textwidth]{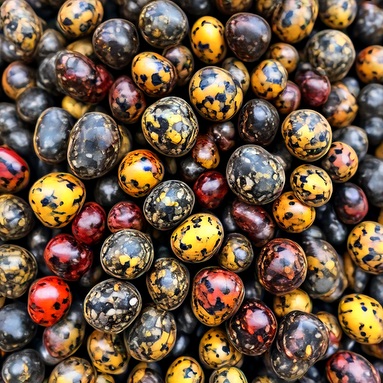} & 
   \includegraphics[width=0.12\textwidth,height=0.12\textwidth]{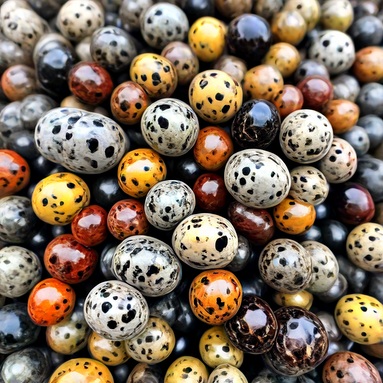} & 
   \includegraphics[width=0.12\textwidth,height=0.12\textwidth]{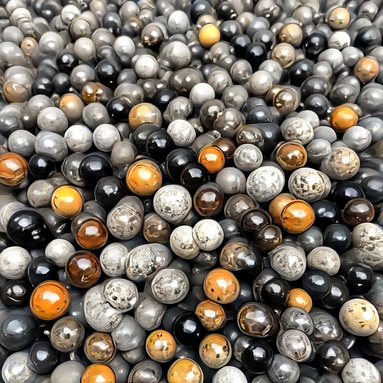} 
   \\
   \midrule
    \multicolumn{6}{c}{"Bathtub" - Prompt: "A set of scented lotions arranged on a shelf."} \\
    \makecell{VLM: "no"\\$p_\text{yes}: 0.03$\\$p_\text{det}: 0.00$} &
   \makecell{VLM: "no"\\$p_\text{yes}: 0.06$\\$p_\text{det}: 0.00$} &
    \makecell{VLM: "no"\\$p_\text{yes}: 0.04$\\$p_\text{det}: 0.00$} &
    \makecell{VLM: "no"\\$p_\text{yes}: 0.08$\\$p_\text{det}: 0.00$} &
    \makecell{VLM: "yes"\\$p_\text{yes}: \mathbf{0.71}$\\$p_\text{det}: 0.00$} &
    \makecell{VLM: "yes"\\$p_\text{yes}: 0.47$\\$p_\text{det}: 0.00$} 
    \\
    \includegraphics[width=0.12\textwidth,height=0.12\textwidth]{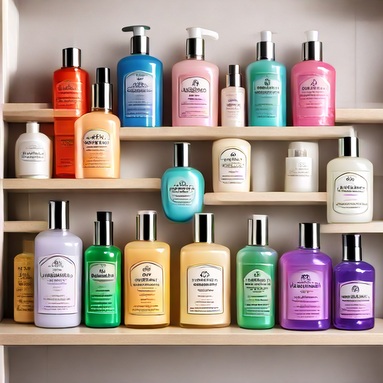} & 
   \includegraphics[width=0.12\textwidth,height=0.12\textwidth]{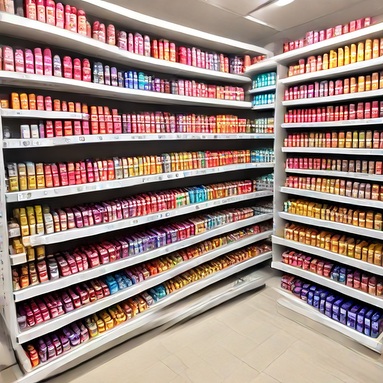} & 
   \includegraphics[width=0.12\textwidth,height=0.12\textwidth]{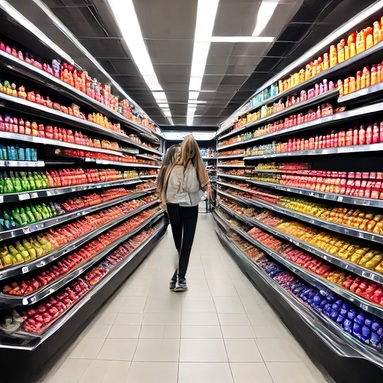} & 
   \includegraphics[width=0.12\textwidth,height=0.12\textwidth]{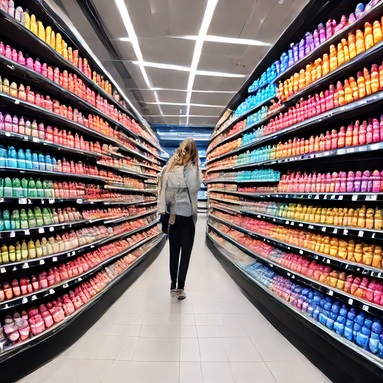} & 
   \includegraphics[width=0.12\textwidth,height=0.12\textwidth]{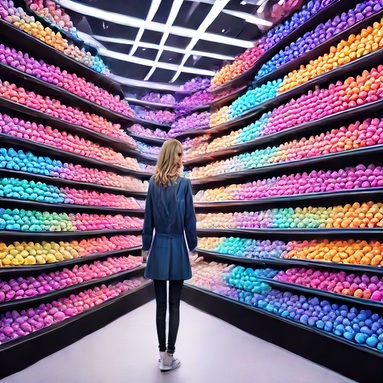} & 
   \includegraphics[width=0.12\textwidth,height=0.12\textwidth]{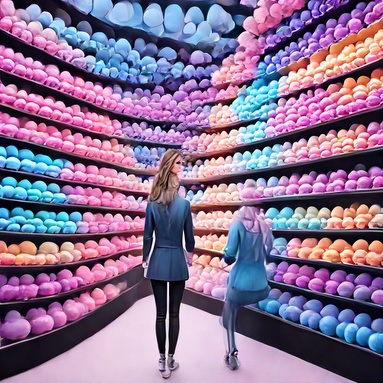} 
   \\
   \bottomrule
    \end{tabular}
    \caption{Optimization trajectories for \oursopt for \pali. For each example, we present the object label, the \oursllm query used to initialize the generation, as well as the answer and "yes" probability from the VLM and the probability from the detector. Through our optimization process, we can uncover model-specific "unknown unknowns," such as the "beads" (see \cref{fig:retrieval} for retrieved images) or the "bath bombs" (see \cref{fig:app_retrieval_pali_opt}). Since the last image
    is not necessarily the best, we select the image with the lowest loss as the query.
 }
    \label{fig:app_optim_progress}
\end{figure*}

\begin{figure*}[phtb]
    \centering
    \small
    \setlength{\tabcolsep}{1pt} %
    \renewcommand{\arraystretch}{1} %
    \begin{tabular}{cc|cc|cc}
    \toprule
     Initialization  & \oursopt & Initialization  & \oursopt  & Initialization  & \oursopt\\
    \midrule
    \multicolumn{2}{c|}{\makecell{\pali \, - "Gar"\\"A water's edge with a\\ few rocks and pebbles."}} &
    \multicolumn{2}{c|}{\makecell{\pali \, - "Cabbage Butterfly"\\"A white flower with a yellow center\\and a delicate, lacy texture."}}  &
    \multicolumn{2}{c}{\makecell{\llavanext \vicuna \, - "Pill bottle"\\"A pair of slippers next to\\a piece of furniture."}}  
    \\
    \makecell{VLM: "no"\\$p_\text{yes}: 0.03$\\$p_\text{det}: 0.00$} &
    \makecell{VLM: "yes"\\$p_\text{yes}: 0.68$\\$p_\text{det}: 0.00$} &
    \makecell{VLM: "no"\\$p_\text{yes}: 0.01$\\$p_\text{det}: 0.00$} &
    \makecell{VLM: "yes"\\$p_\text{yes}: 0.63$\\$p_\text{det}: 0.00$} &
    \makecell{VLM: "no"\\$p_\text{yes}: 0.21$\\$p_\text{det}: 0.00$} &
    \makecell{VLM: "yes"\\$p_\text{yes}: 0.87$\\$p_\text{det}: 0.00$}
    \\
    \includegraphics[width=0.14\textwidth,height=0.14\textwidth]{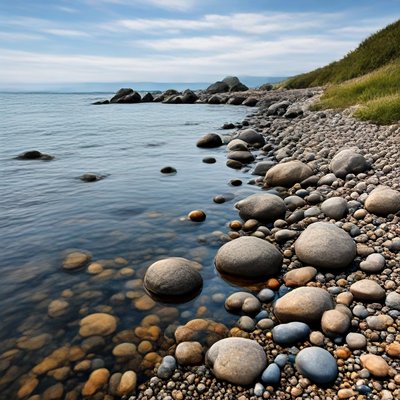} & 
    \includegraphics[width=0.14\textwidth,height=0.14\textwidth]{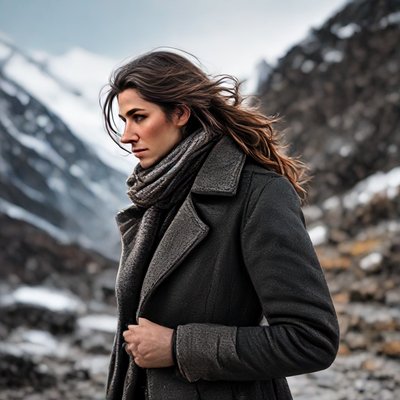} & 
    \includegraphics[width=0.14\textwidth,height=0.14\textwidth]{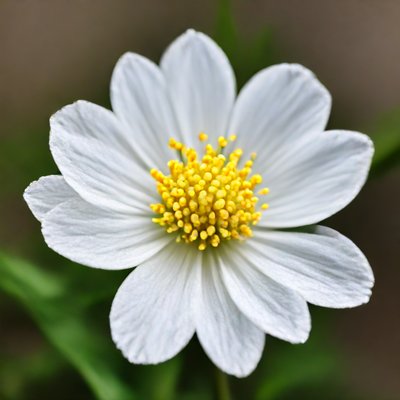} & 
   \includegraphics[width=0.14\textwidth,height=0.14\textwidth]{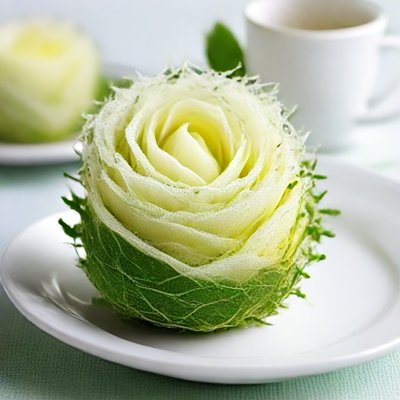} &
    \includegraphics[width=0.14\textwidth,height=0.14\textwidth]{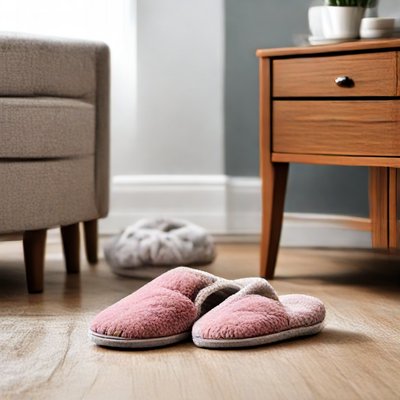} & 
   \includegraphics[width=0.14\textwidth,height=0.14\textwidth]{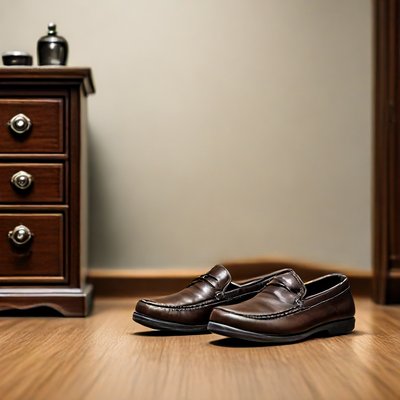} 
   \\
   \midrule
    \multicolumn{2}{c|}{\makecell{\llavanext \vicuna \, - "Gondola"\\"A beautiful Murano glass vase on\\display in a shop window."}}  &
    \multicolumn{2}{c|}{\makecell{\llavanext \mistral \, "Beehive"\\"A person holding a frame in\\a field of blooming flowers."}}  &
    \multicolumn{2}{c}{\makecell{\llavanext \mistral \, "Fortune Cookie"\\"A vibrant street festival\\with dragon dancers."}} 
    \\
    \makecell{VLM: "no"\\$p_\text{yes}: 0.15$\\$p_\text{det}: 0.00$} &
    \makecell{VLM: "yes"\\$p_\text{yes}: 0.61$\\$p_\text{det}: 0.00$} &
    \makecell{VLM: "no"\\$p_\text{yes}: 0.03$\\$p_\text{det}: 0.00$} &
    \makecell{VLM: "yes"\\$p_\text{yes}: 0.52$\\$p_\text{det}: 0.00$} &
    \makecell{VLM: "no"\\$p_\text{yes}: 0.18$\\$p_\text{det}: 0.07$} &
    \makecell{VLM: "yes"\\$p_\text{yes}: 0.85$\\$p_\text{det}: 0.00$}
    \\
    \includegraphics[width=0.14\textwidth,height=0.14\textwidth]{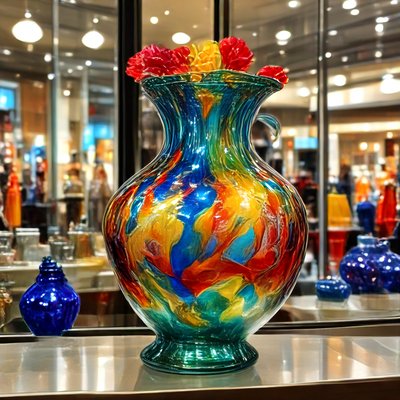} & 
   \includegraphics[width=0.14\textwidth,height=0.14\textwidth]{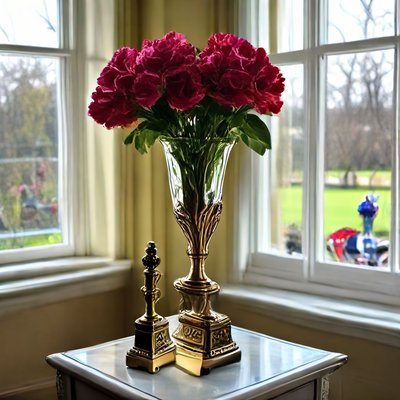}  &
    \includegraphics[width=0.14\textwidth,height=0.14\textwidth]{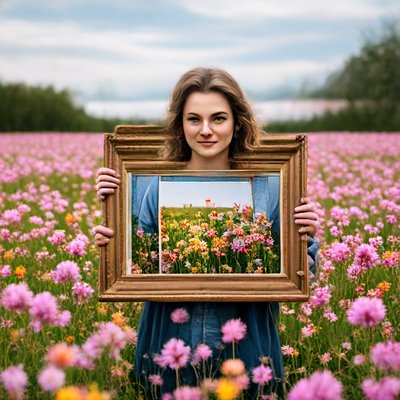} & 
    \includegraphics[width=0.14\textwidth,height=0.14\textwidth]{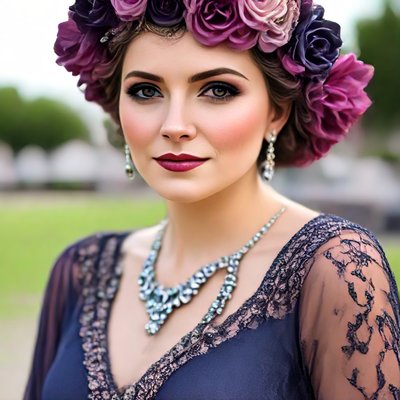} & 
    \includegraphics[width=0.14\textwidth,height=0.14\textwidth]{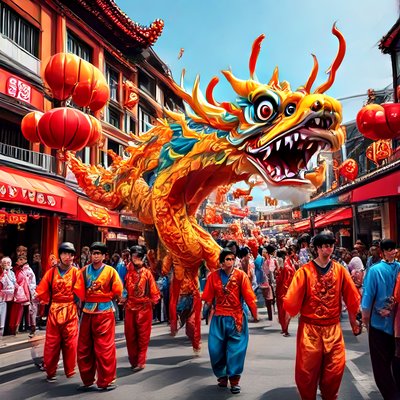} & 
   \includegraphics[width=0.14\textwidth,height=0.14\textwidth]{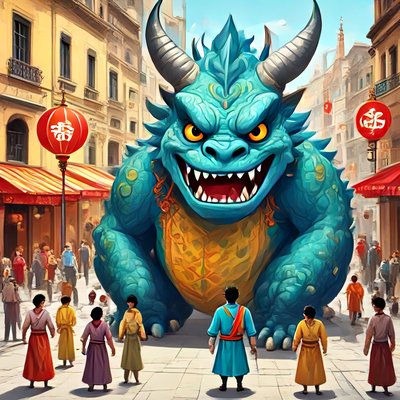} 
   \\
   \bottomrule

    \end{tabular}
    \caption{We show examples of \oursopt query images after optimization together with the initialization generated from the text query. Our optimization is able to generate images that make VLM hallucinate from non-successful prompts without generating the object. }
    \label{fig:app_optim_ab}
\end{figure*}

\label{sec:app_diffusion}

\noindent\textbf{Optimization examples:}
In \cref{fig:app_optim_progress}, we present the optimization trajectory of \oursopt for two images. In \cref{fig:app_optim_ab}, we provide additional examples where we show only the initialization (\ie, the image generated by SDXL using the text query from \oursllm without any optimization) and the final query image produced by \oursopt after optimization. These examples illustrate that \oursopt is capable of generating unexpected \hallu s. For instance, it introduces beads for ``leopard,'' which are absent from the original caption that merely describes ``a rock's cracks and fissures''. Similarly, we demonstrate the transformation of ``a set of scented lotions ... on a shelf'' into a scene of a person shopping for ``bathing bombs'' when optimizing for hallucinations related to the object ``bathtub''.

The corresponding retrieved images, which validate that these phenomena are not limited to synthetic data but also occur with real images, can be found in \cref{fig:retrieval} for ``leopard'' and \cref{fig:app_retrieval_pali_opt} for ``bathtub''.

\noindent\textbf{Implementation details:}
For \oursopt, we use the distilled version of Stable Diffusion XL (SDXL) \cite{podell2023sdxl} from \cite{ren2024hyper}. In particular, we use the single-step SDXL U-Net together with the Latent Consistency Model (LCM) scheduler~\cite{luo2023latent}, setting the start timestep to 800.

For optimization, we use the Adam optimizer~\cite{kingma2014adam} for 25 steps with a step size of $0.1$, applying a linear warmup over the first 3 steps. The gradient is clipped to an $L_2$ norm of 0.1 at every step. When using a deterministic scheduler like the single-step LCM scheduler, three variables determine the output of the diffusion process. The first is the Gaussian random latent drawn at the start of the generation. The second and third are the text encodings of the user prompt generated by the two different CLIP text encoders in SDXL. We optimize all three variables and additionally apply a step size factor of $0.1$ for the random latent. For the random latent, we also employ the chi-square latent regularization method from \cite{samuel2024norm}. Note that the text encodings are initialized using the text queries from \oursllm (\cref{sec:app_llm_prompt}). As optimization loss, we use \cref{eq:oursopt}. Note that $p_\text{det}\left(  \text{\obj} \mid q(C) \right)$ is computed as the maximum confidence overall bounding boxes and thresholded to $0$ for detection probabilities smaller than $0.05$. Since the optimization problem is highly non-convex, the last image is not necessarily the one with the best overall loss, and hence, we use the one with the lowest loss over all generated images as the query for \oursopt.

The optimization takes around 50 seconds for \pali\ and one minute for the \llavanext\ models on an NVIDIA A100 GPU with 80GB of memory.

\section{Retrieval process, exploration and exploitation}\label{sec:app_exploration_exploitation_retrieval}

The ReLAION-5B~\cite{relaion5b, schuhmann2022laion} index, which we use for retrieval during the exploration and exploitation stages, is based on OpenCLIP ViT-H~\cite{ilharco2021openclip}. During retrieval, we apply DreamSim~\cite{fu2023dreamsim} to remove near-duplicate images with a similarity score greater than $0.9$, as LAION is estimated to contain up to 30\% duplicated data~\cite{webster2023duplication}. 
For clustering in the exploitation phase, we first group all images retrieved for the same image during the exploration phase into pre-clusters. These pre-clusters are then merged using agglomerative clustering to form the final clusters. We employ average linkage based on DreamSim distances (note that in the main paper we wrongly stated that we use the CLIP embedding distance instead - however, note that DreamSim is based on fine-tuned CLIP and DINO embeddings - we correct this in the final version), with a maximum allowed merge threshold of $0.6$.

\section{Impact of object occurence frequency on object hallucinations}\label{app:oi-frequency}

\begin{figure*}[htbp]
    \centering
    
    \begin{subfigure}[t]{0.32\textwidth}
        \centering
        \includegraphics[width=\textwidth]{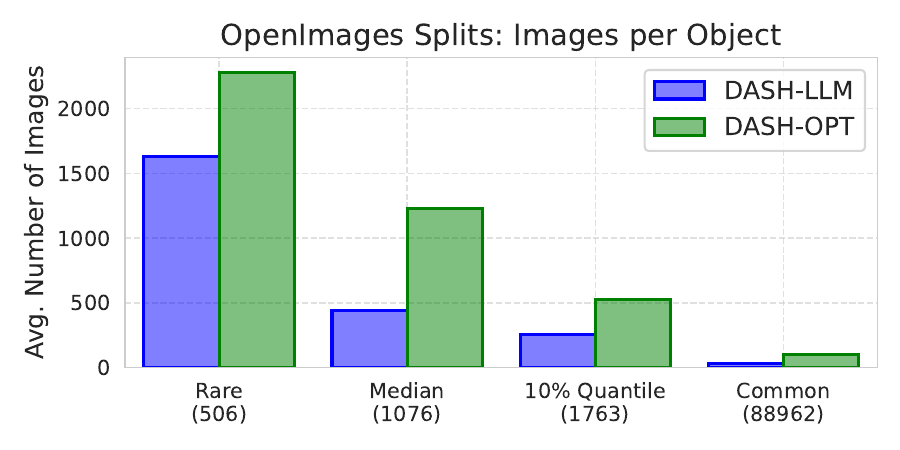}
        \caption{\pali}
    \end{subfigure}
    \hfill
    \begin{subfigure}[t]{0.32\textwidth}
        \centering
        \includegraphics[width=\textwidth]{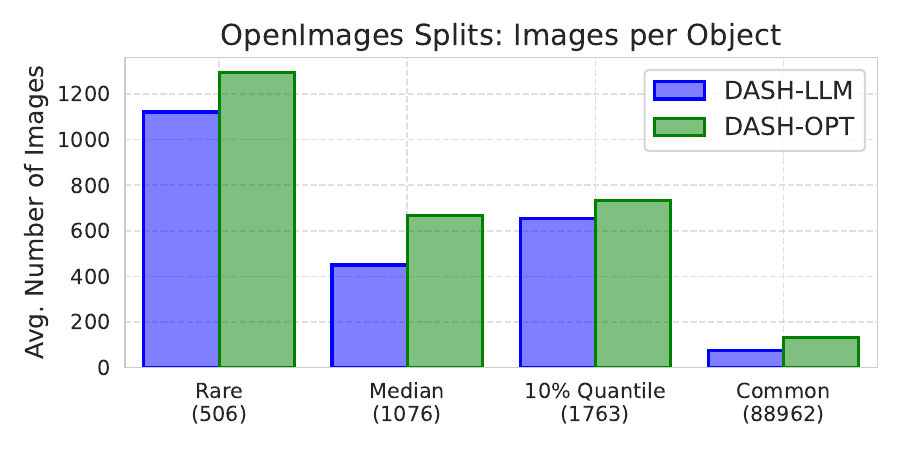}
        \caption{\llavanext \vicuna}
    \end{subfigure}
    \hfill
    \begin{subfigure}[t]{0.32\textwidth}
        \centering
        \includegraphics[width=\textwidth]{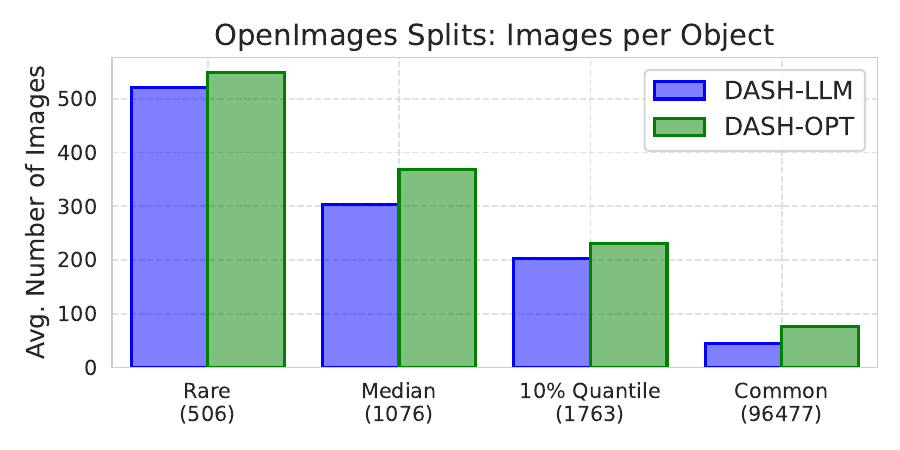}
        \caption{\llavanext \mistral}
    \end{subfigure}
    
    \caption{\textbf{Influence of object frequencies:}
    Histogram showing the average number of success images per object category across the OpenImages splits for \oursllm and OPT with \pali, \lnvicuna and \lnmistral. The average number of training examples per class in the full 9M OpenImages dataset is indicated in parentheses. The plot reveals that rarer concepts are more susceptible to %
    \hallu s, whereas common concepts with tens of thousands of examples are much less prone to such errors.}
    \label{fig:app_oi_frequencies_all}
\end{figure*}

\yn{
We run \ours on subsets of OpenImages with different occurrence frequencies and show the average number of images per object for each split found by \ours for \pali \llavanext \vicuna, and \llavanext \mistral in \cref{fig:app_oi_frequencies_all}. The results for \pali are particularly interesting, as the model was trained on a similar task (``Is there a \emph{object} in the image?'') on this dataset. Overall, it is easier to find systematic hallucinations for objects that are very rare (on average $506$ images) and gets harder if they occur more frequently. Especially for the frequent objects, the optimized queries help to find more of the rarer hallucinations, resulting in significantly more images per object for \oursopt on the corresponding splits compared to \oursllm.
The observed trends for \pali also hold true for  \llavanext \vicuna and \llavanext \mistral. Both are much more vulnerable on rare objects, and object frequency seems to be a strong indicator of an object's vulnerability although the \llavanext models are not trained on OpenImages. However, it is possible that the distribution of images in OpenImages is similar to that of other large-scale datasets, such as those used to train the CLIP~\cite{radford2021clip} models employed in \llavanext.}

\section{\ours Results Extended}
\label{app:retrieval}

\begin{figure*}[htbp]
    \centering
    \small
    \setlength{\tabcolsep}{1pt} %
    \renewcommand{\arraystretch}{1} %
\begin{tabular}{c|cccccccccc}
        \toprule
        \textbf{Barracouta} & 
        \multicolumn{10}{c}{ Cluster Size: 174, Query: A coastal town with a bustling harbor.}\\
        \includegraphics[width=0.083\textwidth,height=0.083\textwidth]{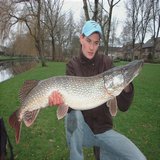} &
        \includegraphics[width=0.083\textwidth,height=0.083\textwidth]{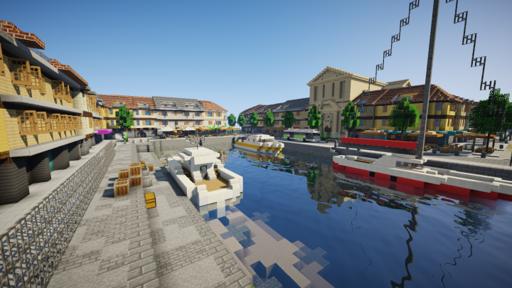} &
        \includegraphics[width=0.083\textwidth,height=0.083\textwidth]{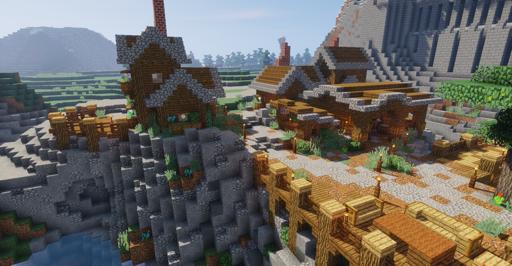} &
        \includegraphics[width=0.083\textwidth,height=0.083\textwidth]{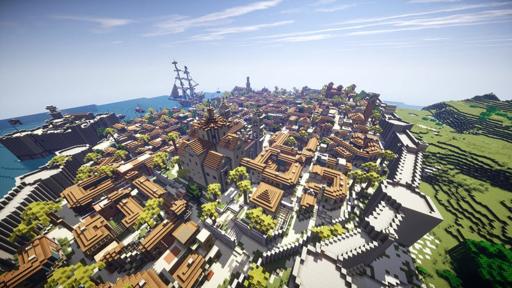} &
        \includegraphics[width=0.083\textwidth,height=0.083\textwidth]{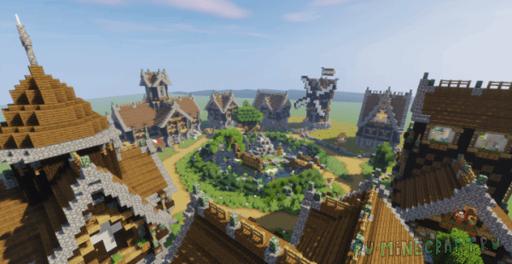} &
        \includegraphics[width=0.083\textwidth,height=0.083\textwidth]{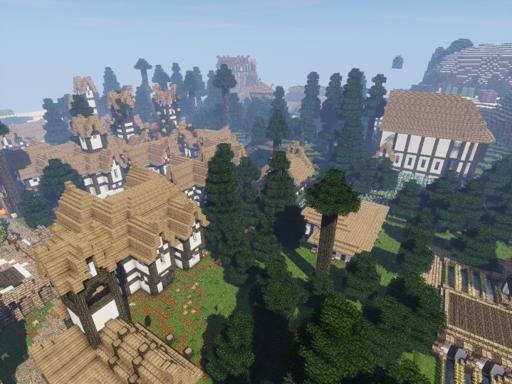} &
        \includegraphics[width=0.083\textwidth,height=0.083\textwidth]{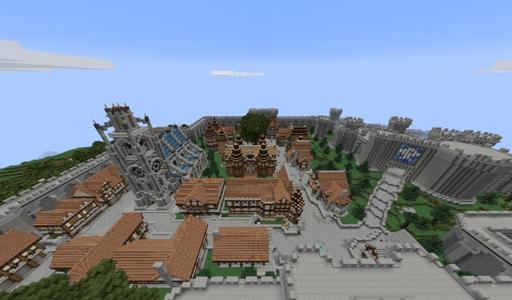} &
        \includegraphics[width=0.083\textwidth,height=0.083\textwidth]{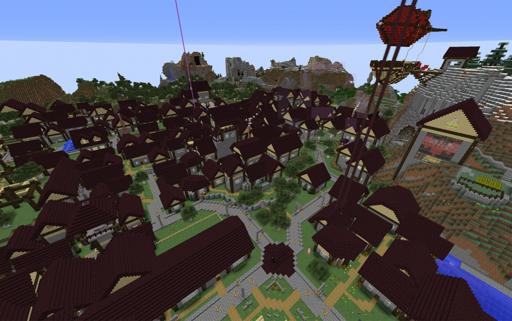} &
        \includegraphics[width=0.083\textwidth,height=0.083\textwidth]{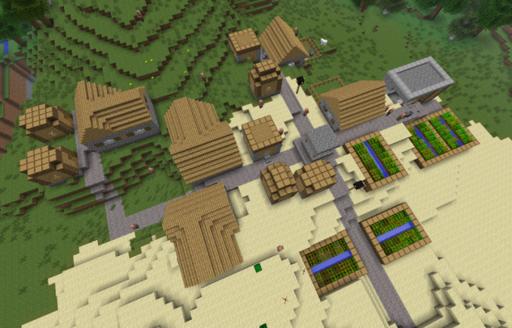} & 
        \includegraphics[width=0.083\textwidth,height=0.083\textwidth]{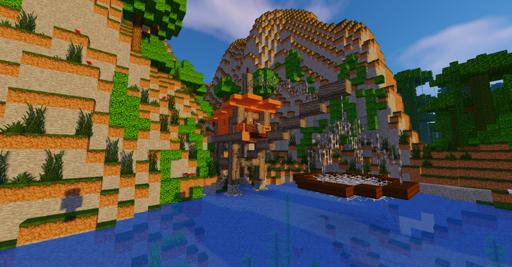} &
        \includegraphics[width=0.083\textwidth,height=0.083\textwidth]{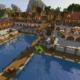} 
        \\
        \midrule
        \textbf{Fireboat} &
        \multicolumn{10}{c}{Cluster Size: 93, Query: A water cannon spraying a crowd of people.}\\
\bottomrule
    \end{tabular}
    \caption{\oursllm \pali - Please see \cref{sec:app_add_qualitative} for a description.}
    \label{fig:app_retrieval_pali_llm}
\end{figure*}

\begin{figure*}[htbp]
    \centering
    \small
    \setlength{\tabcolsep}{1pt} %
    \renewcommand{\arraystretch}{1} %
    \begin{tabular}{c|c|ccccccccc}
        \toprule
        \textbf{Bathtub} & \textbf{Query} &
        \multicolumn{8}{c}{Cluster Size: 31}\\
        \includegraphics[width=0.083\textwidth,height=0.083\textwidth]{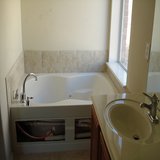} &
        \includegraphics[width=0.083\textwidth,height=0.083\textwidth]{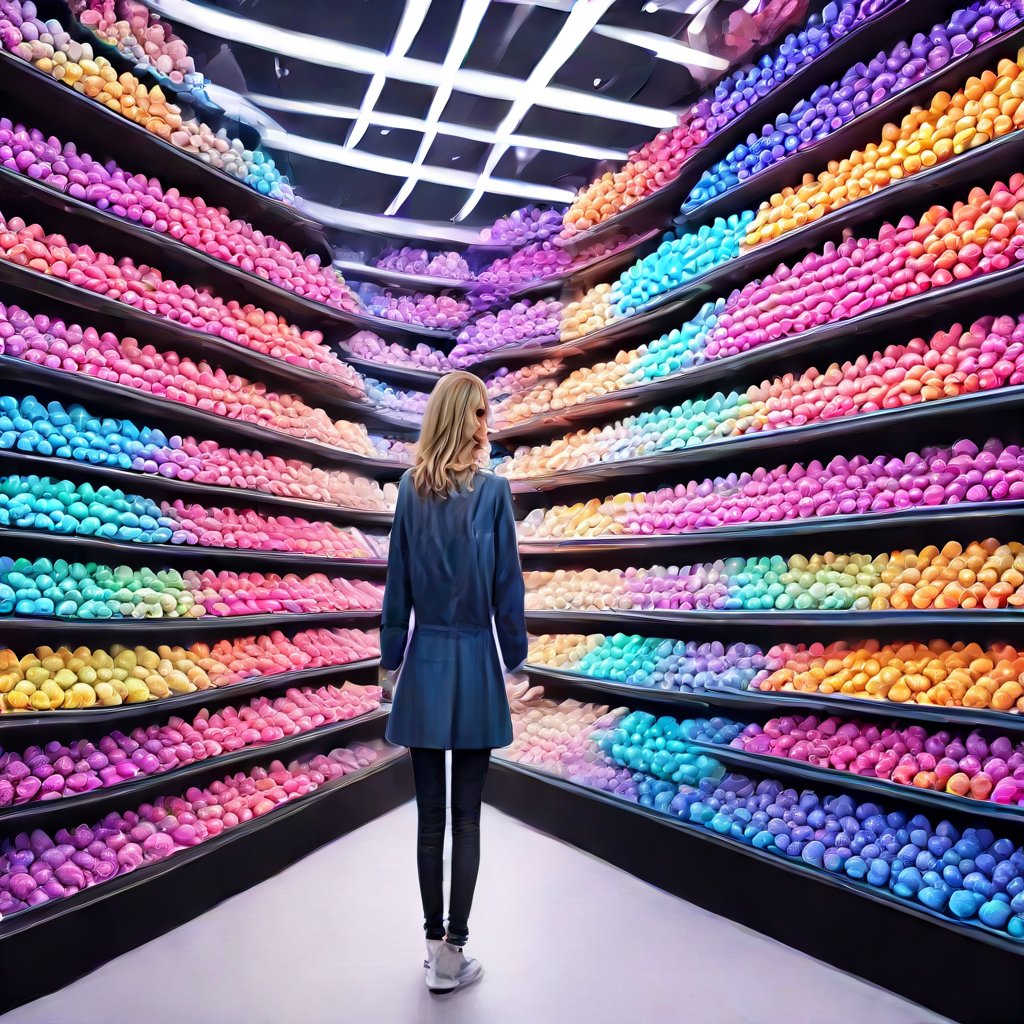}
        &
        \includegraphics[width=0.083\textwidth,height=0.083\textwidth]{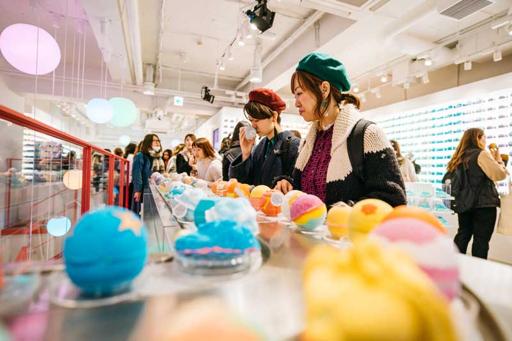} &
        \includegraphics[width=0.083\textwidth,height=0.083\textwidth]{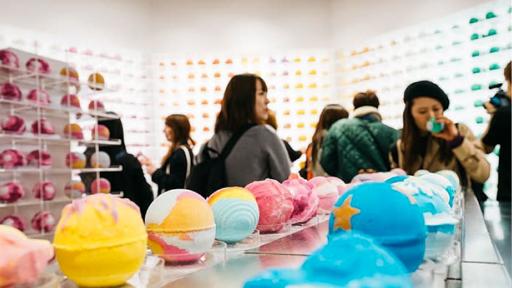} &
        \includegraphics[width=0.083\textwidth,height=0.083\textwidth]{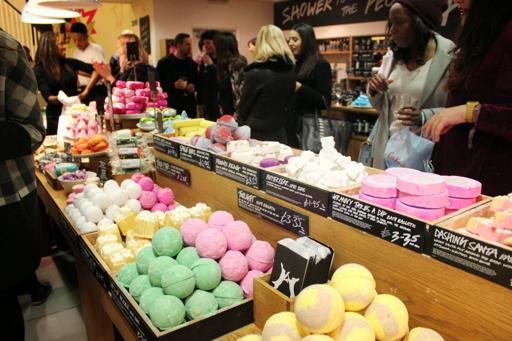} &
        \includegraphics[width=0.083\textwidth,height=0.083\textwidth]{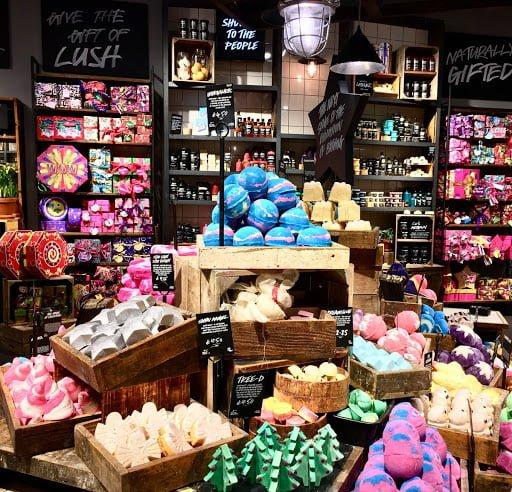} &
        \includegraphics[width=0.083\textwidth,height=0.083\textwidth]{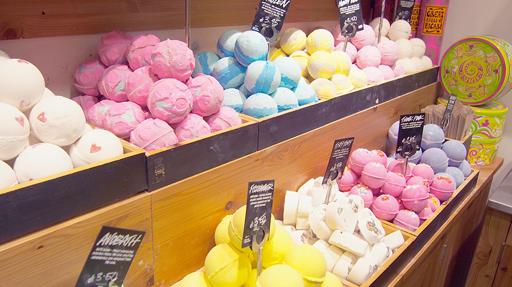} &
        \includegraphics[width=0.083\textwidth,height=0.083\textwidth]{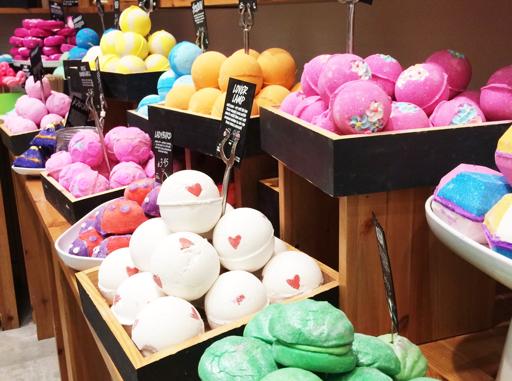} &
        \includegraphics[width=0.083\textwidth,height=0.083\textwidth]{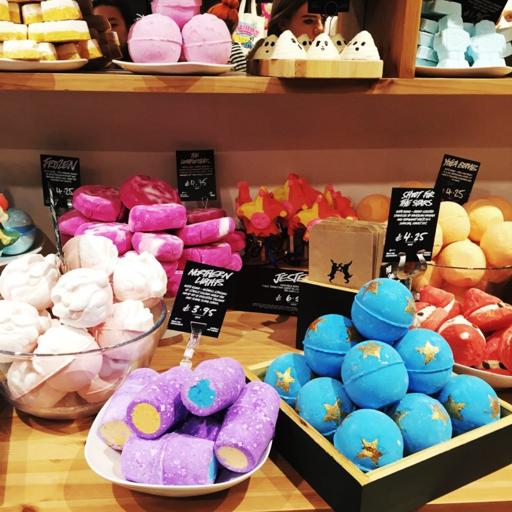} &
        \includegraphics[width=0.083\textwidth,height=0.083\textwidth]{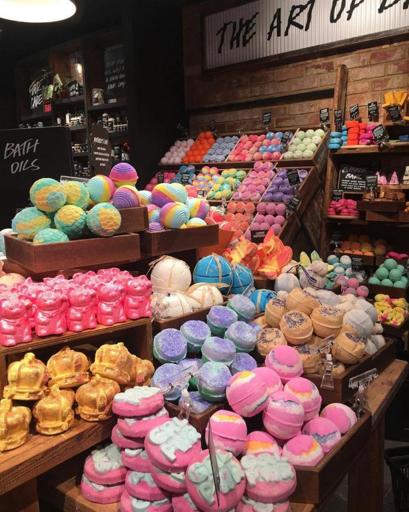}  &
        \includegraphics[width=0.083\textwidth,height=0.083\textwidth]{images/retrieval/pali_img/si_bathtub/0_9.jpg}  
        \\
        \midrule
        \textbf{Puck} & \textbf{Query}  &
        \multicolumn{8}{c}{Cluster Size: 74}\\
\bottomrule
    \end{tabular}
    \caption{\oursopt \pali - Please see \cref{sec:app_add_qualitative} for a description.}
    \label{fig:app_retrieval_pali_opt}
\end{figure*}

\begin{figure*}[htbp]
    \centering
    \small
    \setlength{\tabcolsep}{1pt} %
    \renewcommand{\arraystretch}{1} %
    \begin{tabular}{c|cccccccccc}
        \toprule
        \textbf{\tiny Academic Gown} &
        \multicolumn{10}{c}{Cluster Size: 69, Query: A university seal emblazoned on a stone wall.}\\
        \includegraphics[width=0.083\textwidth,height=0.083\textwidth]{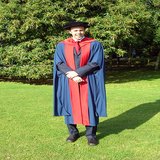} &
        \includegraphics[width=0.083\textwidth,height=0.083\textwidth]{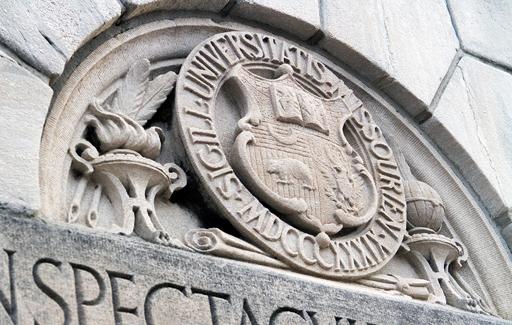} &
        \includegraphics[width=0.083\textwidth,height=0.083\textwidth]{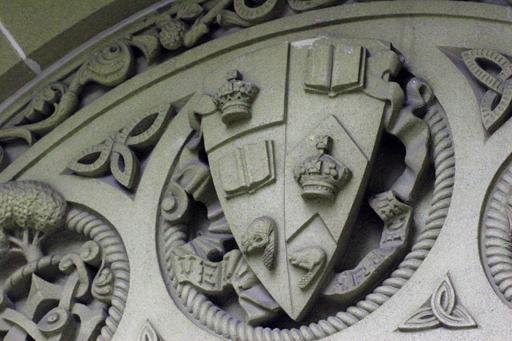} &
        \includegraphics[width=0.083\textwidth,height=0.083\textwidth]{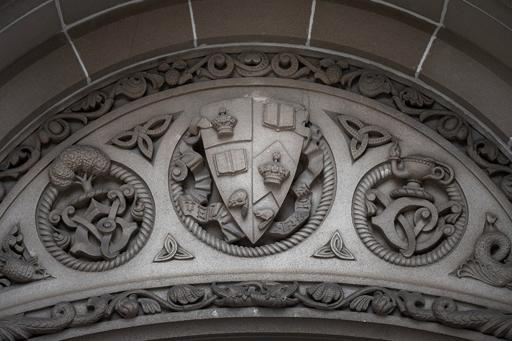} &
        \includegraphics[width=0.083\textwidth,height=0.083\textwidth]{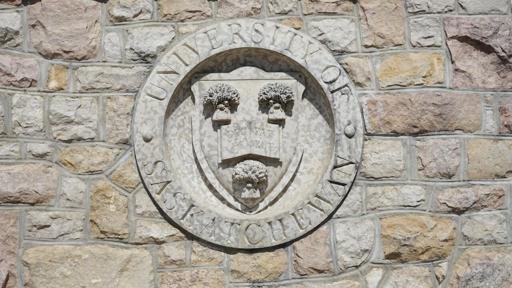} &
        \includegraphics[width=0.083\textwidth,height=0.083\textwidth]{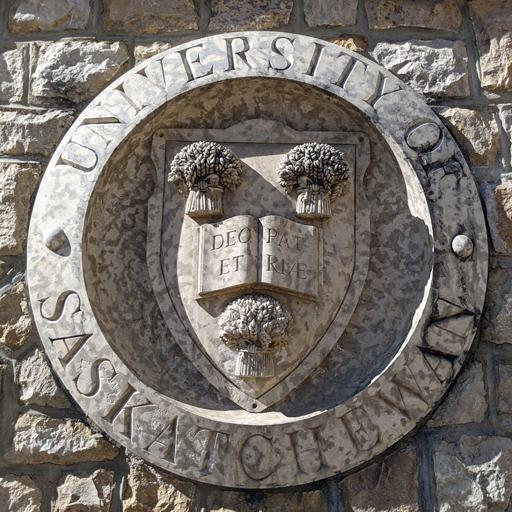} &
        \includegraphics[width=0.083\textwidth,height=0.083\textwidth]{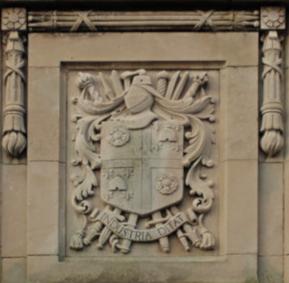} &
        \includegraphics[width=0.083\textwidth,height=0.083\textwidth]{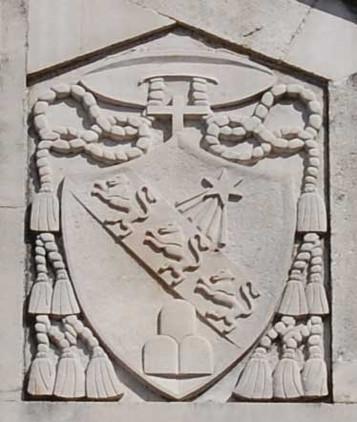} &
        \includegraphics[width=0.083\textwidth,height=0.083\textwidth]{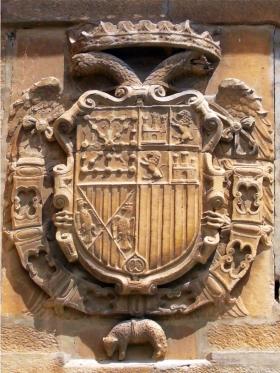} &
        \includegraphics[width=0.083\textwidth,height=0.083\textwidth]{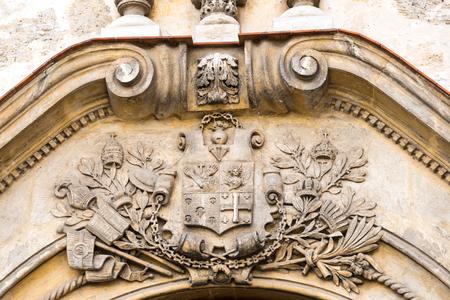} &
        \includegraphics[width=0.083\textwidth,height=0.083\textwidth]{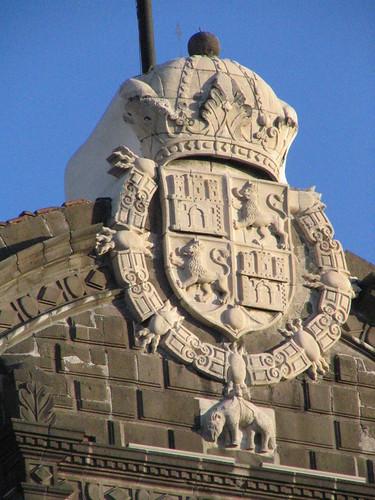} 
        \\
        \midrule
        \textbf{\scriptsize Balance Beam} &
        \multicolumn{10}{c}{Cluster Size: 100, Query: An athlete's feet in grip soles standing on a mat.}\\
\bottomrule

    \end{tabular}
    \caption{\oursllm - \llavanext \vicuna - Please see \cref{sec:app_add_qualitative} for a description.}
    \label{fig:app_retrieval_lnv_llm}
\end{figure*}

\begin{figure*}[htbp]
    \centering
    \small
    \setlength{\tabcolsep}{1pt} %
    \renewcommand{\arraystretch}{1} %
    \begin{tabular}{c|c|ccccccccc}
        \toprule
        \textbf{\scriptsize Airplane Wing} & \textbf{Query} & 
        \multicolumn{9}{c}{ Cluster Size: 53}\\
        \includegraphics[width=0.083\textwidth,height=0.083\textwidth]{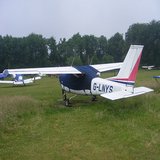} &
        \includegraphics[width=0.083\textwidth,height=0.083\textwidth]{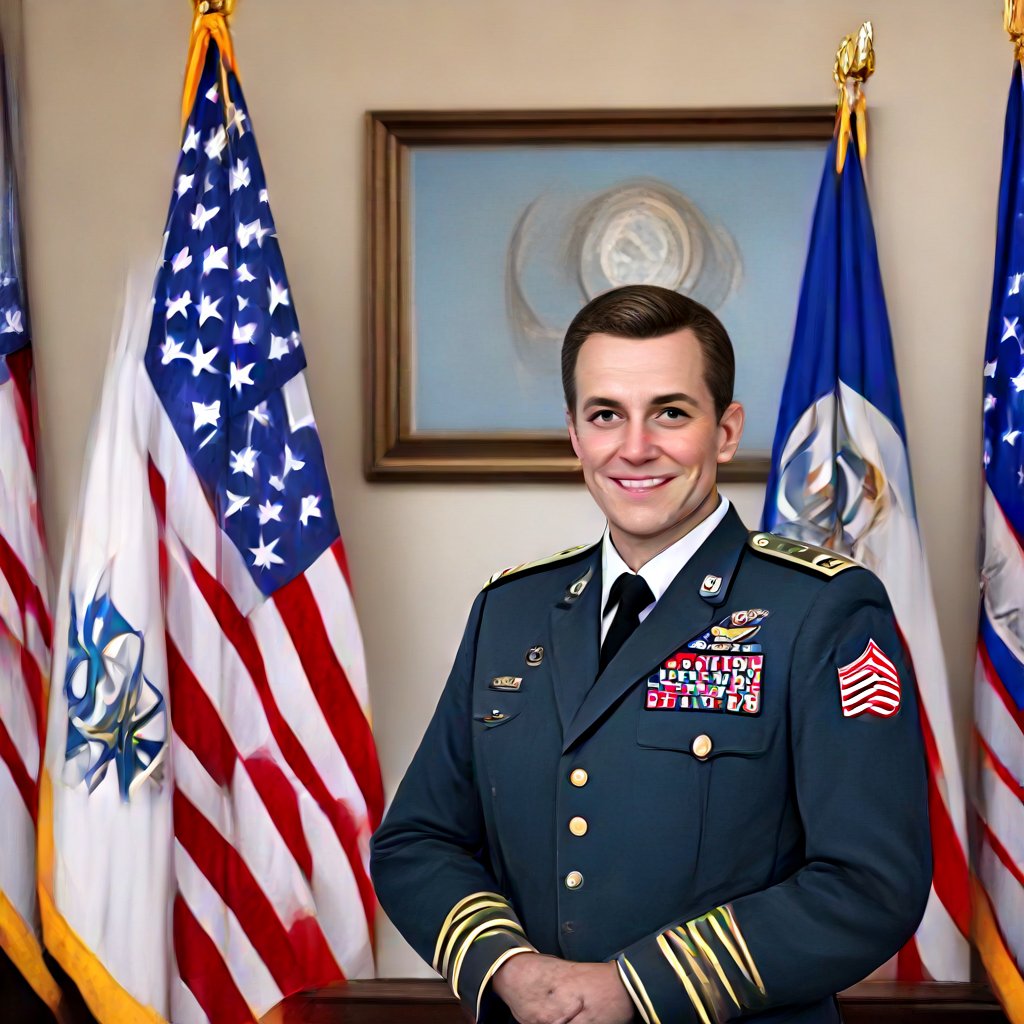}
        &
        \includegraphics[width=0.083\textwidth,height=0.083\textwidth]{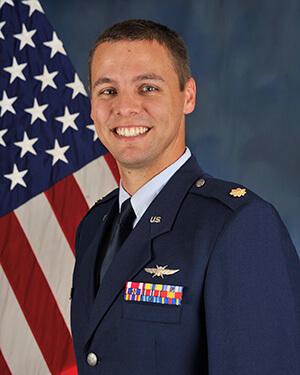} &
        \includegraphics[width=0.083\textwidth,height=0.083\textwidth]{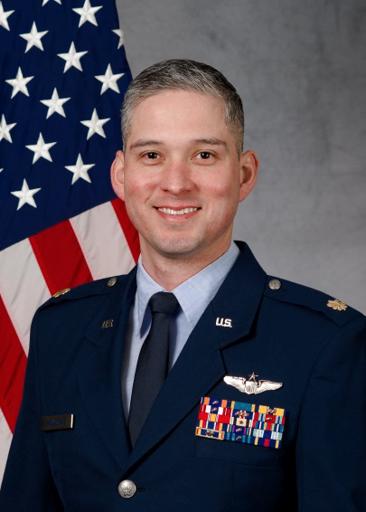} &
        \includegraphics[width=0.083\textwidth,height=0.083\textwidth]{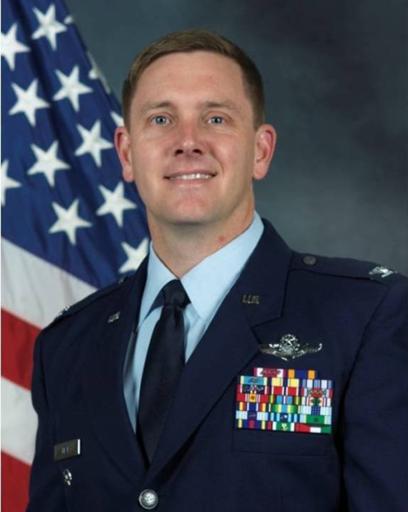} &
        \includegraphics[width=0.083\textwidth,height=0.083\textwidth]{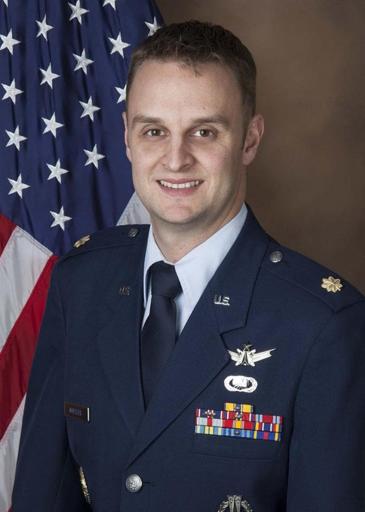} &
        \includegraphics[width=0.083\textwidth,height=0.083\textwidth]{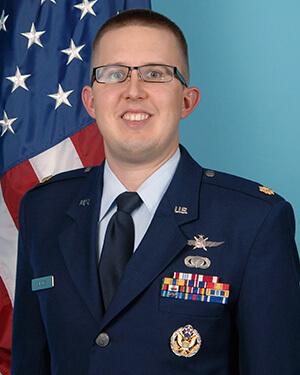} &
        \includegraphics[width=0.083\textwidth,height=0.083\textwidth]{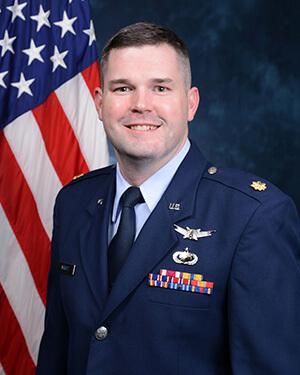} &
        \includegraphics[width=0.083\textwidth,height=0.083\textwidth]{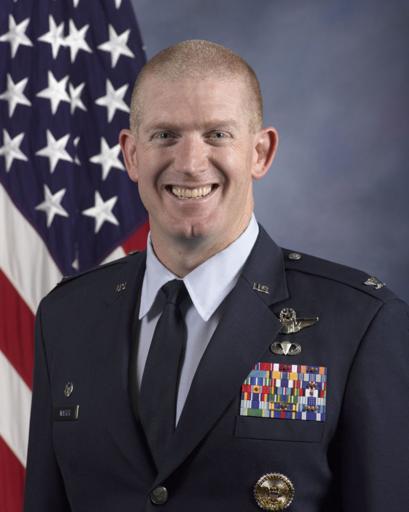} &
        \includegraphics[width=0.083\textwidth,height=0.083\textwidth]{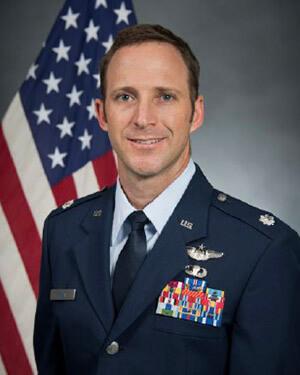} &
        \includegraphics[width=0.083\textwidth,height=0.083\textwidth]{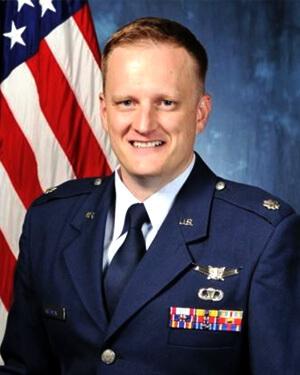} 
        \\
        \midrule
        \textbf{\scriptsize Bathing Cap} & \textbf{Query} &
        \multicolumn{9}{c}{Cluster Size: 43}\\
\bottomrule
    \end{tabular}
    \caption{\oursopt \llavanext \vicuna - Please see \cref{sec:app_add_qualitative} for a description.}
    \label{fig:app_retrieval_lnv_opt}
\end{figure*}

\begin{figure*}[htbp]
    \centering
    \small
    \setlength{\tabcolsep}{1pt} %
    \renewcommand{\arraystretch}{1} %
    \begin{tabular}{c|cccccccccc}
        \toprule
        \textbf{Band Aid} &
        \multicolumn{10}{c}{Cluster Size: 326, Query: A person holding a injured wrist.} \\
        \includegraphics[width=0.083\textwidth,height=0.083\textwidth]{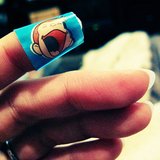} &
        \includegraphics[width=0.083\textwidth,height=0.083\textwidth]{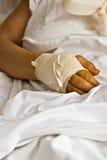} &
        \includegraphics[width=0.083\textwidth,height=0.083\textwidth]{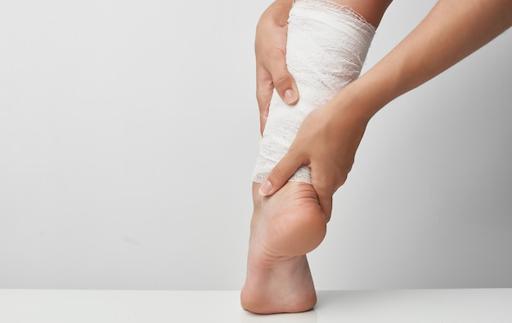} &
        \includegraphics[width=0.083\textwidth,height=0.083\textwidth]{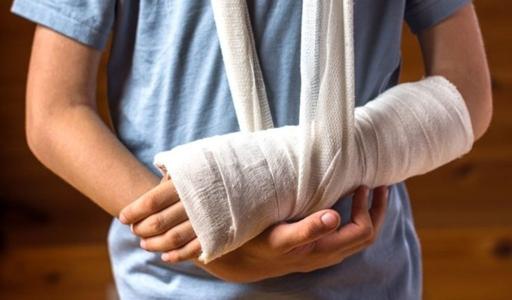} &
        \includegraphics[width=0.083\textwidth,height=0.083\textwidth]{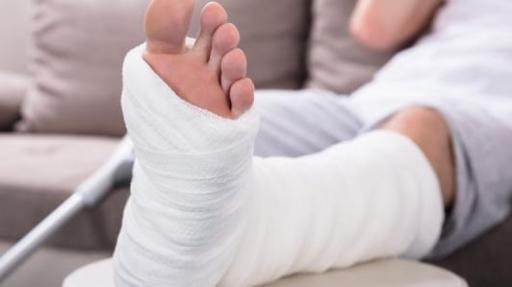} &
        \includegraphics[width=0.083\textwidth,height=0.083\textwidth]{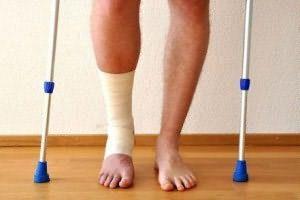} &
        \includegraphics[width=0.083\textwidth,height=0.083\textwidth]{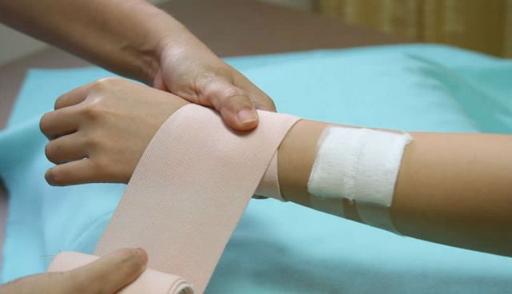} &
        \includegraphics[width=0.083\textwidth,height=0.083\textwidth]{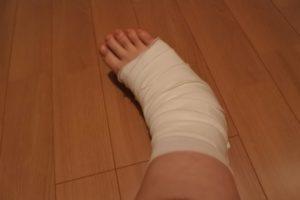} &
        \includegraphics[width=0.083\textwidth,height=0.083\textwidth]{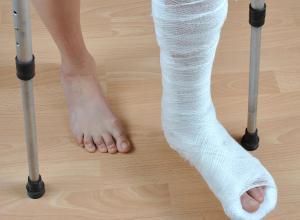} &
        \includegraphics[width=0.083\textwidth,height=0.083\textwidth]{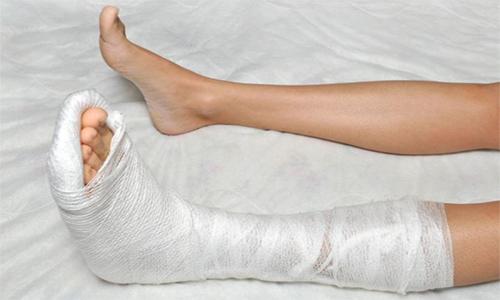} &
        \includegraphics[width=0.083\textwidth,height=0.083\textwidth]{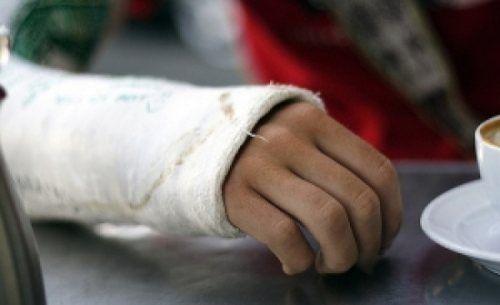} \\
        \midrule
        \textbf{Gondola} &
        \multicolumn{10}{c}{Cluster Size: 411, Query: A charming little shop selling handmade crafts.} \\
\bottomrule
    \end{tabular}
    \caption{\oursllm  \llavanext Mistral - Please see \cref{sec:app_add_qualitative} for a description.}
    \label{fig:app_retrieval_lnm_llm}
\end{figure*}

\begin{figure*}[htbp]
    \centering
    \small
    \setlength{\tabcolsep}{1pt} %
    \renewcommand{\arraystretch}{1} %
    \begin{tabular}{c|c|ccccccccc}
        \toprule
        \textbf{Agama} & \textbf{Query} & \multicolumn{9}{c}{Cluster Size: 237} \\
        \includegraphics[width=0.083\textwidth,height=0.083\textwidth]{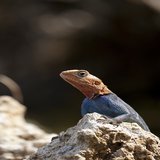} &
        \includegraphics[width=0.083\textwidth,height=0.083\textwidth]{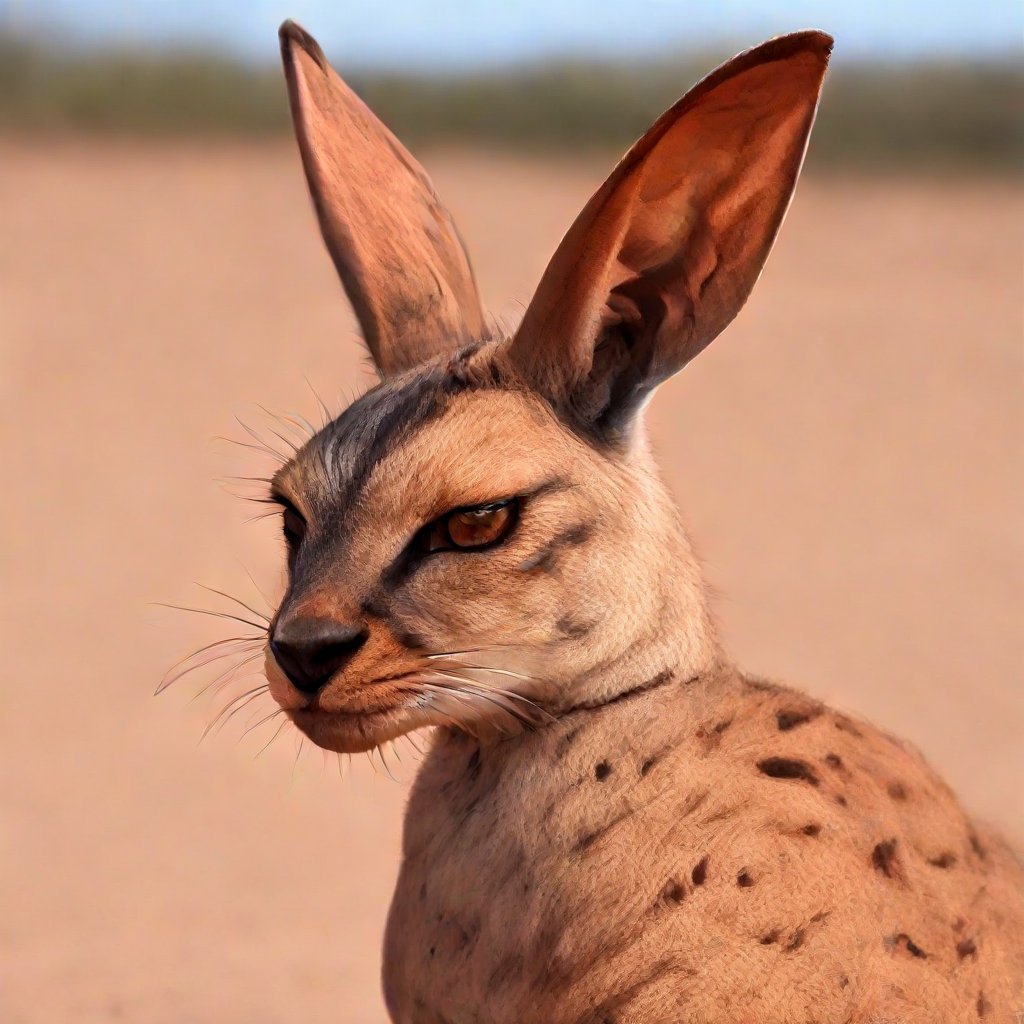}
        &
        \includegraphics[width=0.083\textwidth,height=0.083\textwidth]{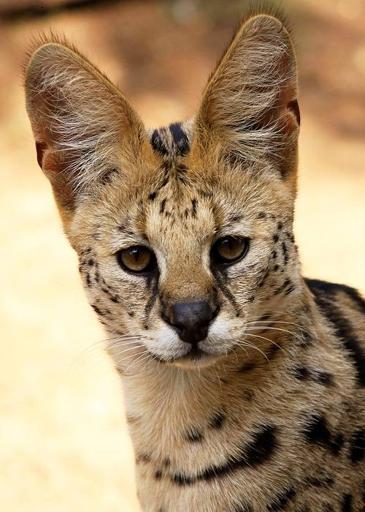} &
        \includegraphics[width=0.083\textwidth,height=0.083\textwidth]{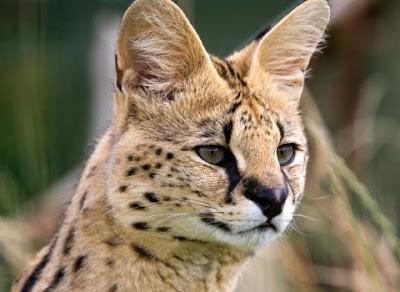} &
        \includegraphics[width=0.083\textwidth,height=0.083\textwidth]{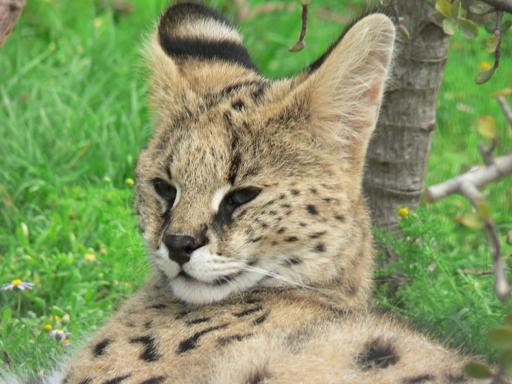} &
        \includegraphics[width=0.083\textwidth,height=0.083\textwidth]{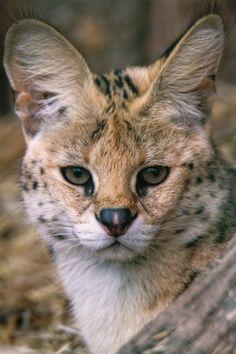} &
        \includegraphics[width=0.083\textwidth,height=0.083\textwidth]{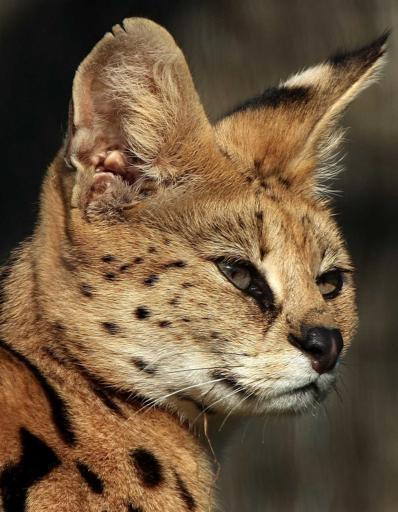} &
        \includegraphics[width=0.083\textwidth,height=0.083\textwidth]{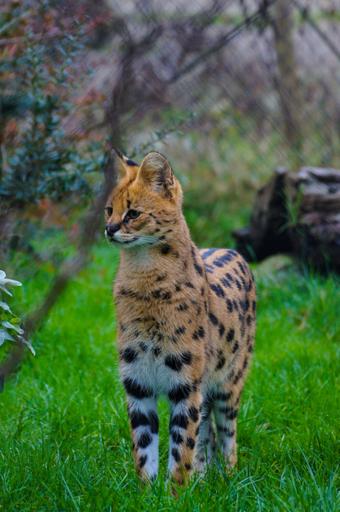} &
        \includegraphics[width=0.083\textwidth,height=0.083\textwidth]{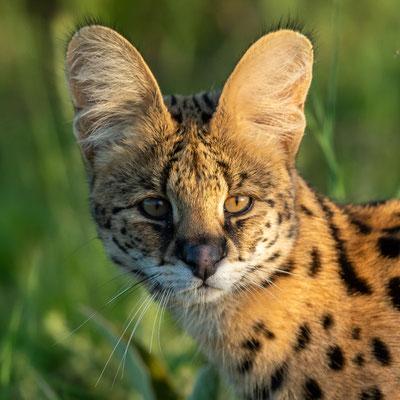} &
        \includegraphics[width=0.083\textwidth,height=0.083\textwidth]{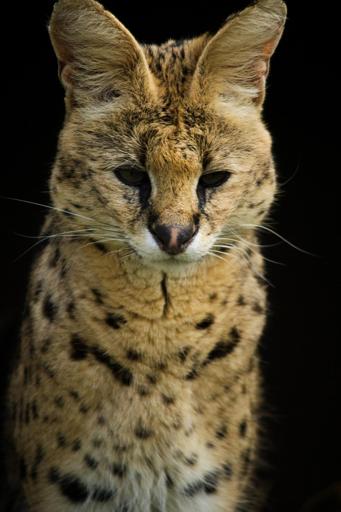} &
        \includegraphics[width=0.083\textwidth,height=0.083\textwidth]{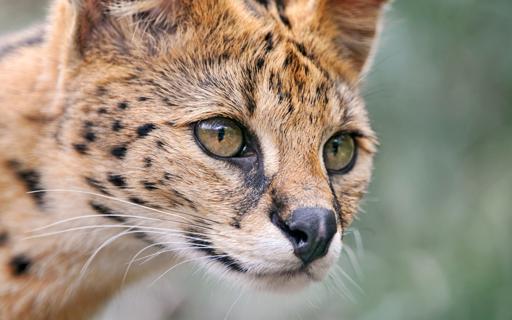} \\
        \midrule
        \textbf{\scriptsize Bulletproof Vest} & \textbf{Query} & \multicolumn{9}{c}{Cluster Size: 63} \\
\bottomrule
    \end{tabular}
    \caption{\oursopt \llavanext \mistral - Please see \cref{sec:app_add_qualitative} for a description.}
    \label{fig:app_retrieval_lnm_opt}
\end{figure*}

\subsection{Additional qualitative examples}\label{sec:app_add_qualitative}
In Figures \ref{fig:app_retrieval_pali_llm} to \ref{fig:app_retrieval_lnm_opt} we present additional retrieval results, similar to those from \cref{fig:retrieval} for \oursllm and \oursopt. In particular, we include results for \llavanext \vicuna and \mistral. As these Figures demonstrate, all 3 VLMs suffer from a substantial amount of type II hallucinations.

In \cref{fig:app_retrieval_pali_llm}, we show the clusters of images generated using \oursllm for \pali. 
The examples illustrate how the LLM-generated queries lead to images that are logically connected to the object in a semantic sense. For instance, for the object ``Barracouta,'' we observe coastal towns and harbors built in Minecraft, likely reflecting the object's marine context. For ``Fireboat'', we find images of the police using water cannons, which are often commonly found on a ``Fireboat''.
In \cref{fig:app_retrieval_pali_opt}, we present examples of clusters identified using \oursopt for the same VLM and various objects, highlighting cases of ``unknown unknowns''. For instance, for the object ``Bathtub,'' the cluster includes colorful images of bath bombs, rather than bathtubs, suggesting that the model has learned to associate the object label with related items rather than the physical object itself. 
Similarly, for the object ``Puck,'' instead of hockey pucks, the cluster prominently features images of stacked oranges and other spherical objects, reflecting a semantic confusion between shape and context. For ``Sulphur Butterfly,'' the cluster contains ornamental decorations and holiday-themed items, diverging significantly from the actual insect.

The clusters shown in \cref{fig:app_retrieval_lnv_llm} illustrate examples generated using \oursllm with \llavanext \vicuna as the VLM. These examples reflect expected yet interesting associations generated by the model based on LLM-guided queries. For instance, for the object ``Academic Gown,'' the cluster includes university seals and architectural elements from academic institutions, which are logically associated with the concept of academia but deviate visually from the actual object. Similarly, for ``Chain Mail,'' the model identifies medieval swords and weaponry, which are contextually related to chain mail in historical settings. The object ``Fountain Pen'' generates a cluster dominated by handwritten scripts and paper stacks, reinforcing a semantic association with writing and stationery. 
The cluster for ``Coral Fungus'' features lichen-covered tree bark and textures, highlighting a broader misinterpretation of the object’s actual form and an emphasis on natural growth patterns. Finally, for ``Postcard,'' the cluster predominantly displays boardwalks and scenic ocean views, which align with common themes of postcards.

In \cref{fig:app_retrieval_lnv_opt}, we showcase clusters generated using \oursopt with \llavanext \vicuna, highlighting more unexpected results where the VLM demonstrates surprising or unintended associations. 
For ``Dogsled,'' the cluster contains images of snowshoes and other winter-related gear, which are contextually linked to snowy environments but do not represent the object itself. 
The object ``Strawberry'' leads to a cluster featuring images of festive door decorations and wreaths. This unexpected association likely arises from the model's inability to separate the red and green color palette of strawberries from decorative elements.
For ``Beehive,'' the cluster includes a surprising array of human portraits, particularly women in colorful settings. This suggests that the VLM may associate the term ``beehive'' with a hairstyle rather than the physical structure created by bees.

In \cref{fig:app_retrieval_lnm_llm}, we present examples of clusters generated using \oursllm with \llavanext \mistral.
For the object ``Band Aid,'' the cluster contains images of people holding or bandaging injured wrists, reflecting a logical semantic association with the concept of injury and care. Similarly, for ``Gondola,'' the cluster features small shops, which aligns with a broader cultural and contextual understanding of gondolas as part of scenic, tourist-driven environments.
The object ``Dumbbell'' leads to a cluster of colorful exercise balls, emphasizing fitness and gym-related settings, likely derived from contextual overlaps. 
For ``Lighter,'' the cluster showcases dimly lit rooms with a smoky haze in video games, reflecting a plausible connection to the object's typical use in dark settings. The ``Lighthouse'' cluster includes solitary piers and fishing-related environments, reinforcing the model's interpretation of the lighthouse’s association with remote coastal locations.

In \cref{fig:app_retrieval_lnm_opt}, we present clusters generated using \oursopt with \llavanext \mistral.
For the object ``Agama,'' the cluster prominently features various wild cats.
For ``Bulletproof Vest,'' the cluster includes images of surveillance and monitoring rooms with large screens, likely due to the association of vests with security and law enforcement. 
The object ``Horizontal Bar'' leads to a cluster filled with water bottles and similar cylindrical objects, reflecting a superficial visual similarity in shape but entirely unrelated semantics. 
For ``Shallot,'' the cluster displays images of modern kitchens and industrial food preparation areas, suggesting that the VLM has learned to associate the object with its culinary context rather than its specific visual characteristics. For ``Bluehead,'' instead of the fish species, the cluster includes images of blue-themed furniture and interior designs, driven by the color association rather than the object itself.
For ``Bird,'' the cluster prominently features butterflies and flowers, illustrating a misalignment between the object category and the broader semantic associations of natural imagery. Lastly, the clusters for ``Hat'' and ``Balloon'' show out-of-distribution images that are not logically connected to the object.

\begin{figure*}[htbp]
    \centering
    \footnotesize
    \setlength{\tabcolsep}{1pt} %
    \renewcommand{\arraystretch}{1} %
    \begin{tabular}{c|cc|cc|cc|cc|cc|cc}
    \multicolumn{13}{c}{\bf{COCO}}\\
    \toprule
    &
    \multicolumn{2}{c|}{{bicycle}} &
    \multicolumn{2}{c|}{{apple}} &
    \multicolumn{2}{c|}{{cat}} &
    \multicolumn{2}{c|}{bear} &
    \multicolumn{2}{c|}{carrot} &
    \multicolumn{2}{c}{cell phone}
    \\
    \midrule
    \multirow{1}{*}[9mm]{\adjustbox{valign=c}{\rotatebox[origin=c]{90}{\ours}}} &

    \includegraphics[cframe=blue 15pt,cframe=blue 15pt,width=0.075\textwidth,height=0.075\textwidth]{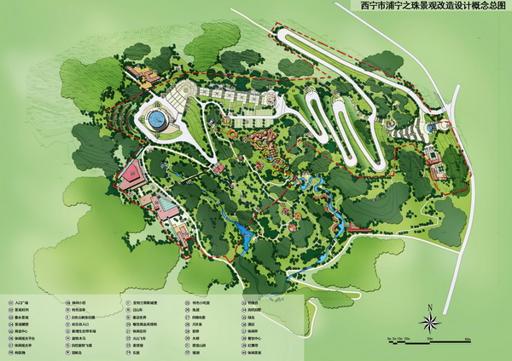} &
    \includegraphics[cframe=blue 15pt,width=0.075\textwidth,height=0.075\textwidth]{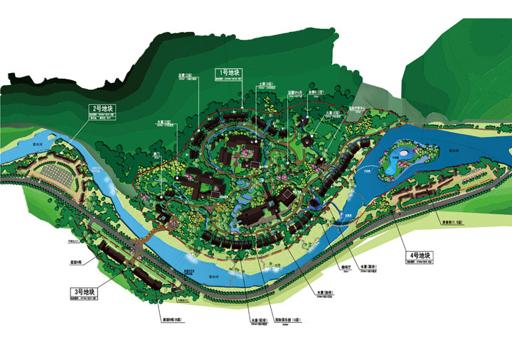} &
    \includegraphics[cframe=blue 15pt,width=0.075\textwidth,height=0.075\textwidth]{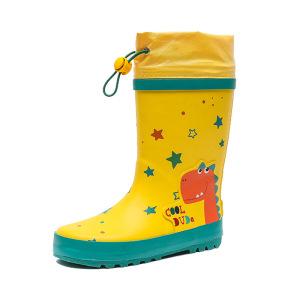} &
    \includegraphics[cframe=blue 15pt,width=0.075\textwidth,height=0.075\textwidth]{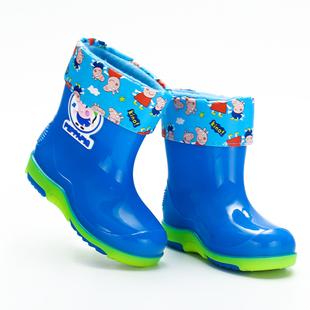} &
    \includegraphics[cframe=blue 15pt,width=0.075\textwidth,height=0.075\textwidth]{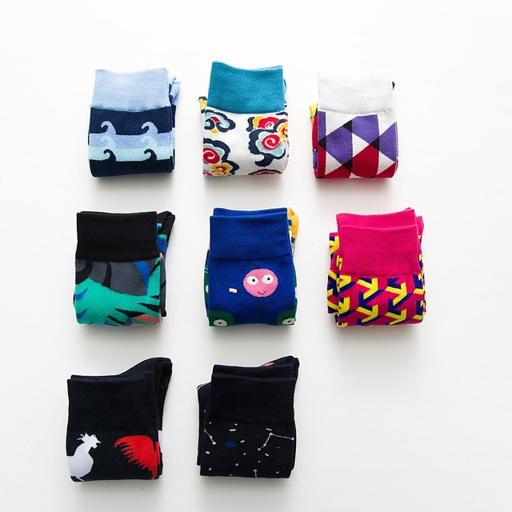} &
    \includegraphics[cframe=blue 15pt,width=0.075\textwidth,height=0.075\textwidth]{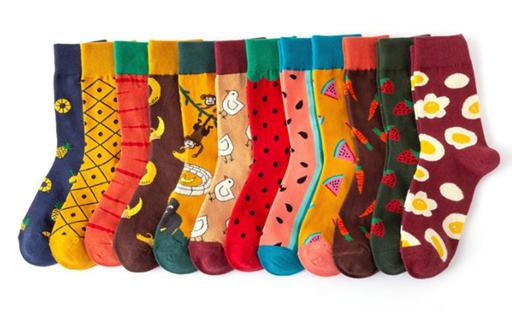} &
    \includegraphics[cframe=blue 15pt,width=0.075\textwidth,height=0.075\textwidth]{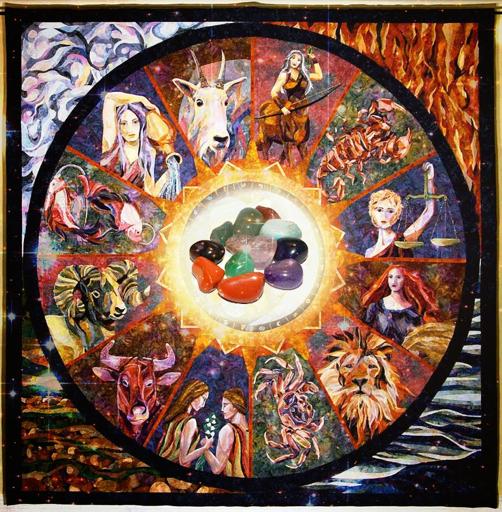} &
    \includegraphics[cframe=blue 15pt,width=0.075\textwidth,height=0.075\textwidth]{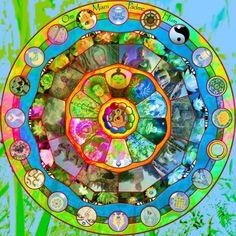} &
    \includegraphics[cframe=blue 15pt,width=0.075\textwidth,height=0.075\textwidth]{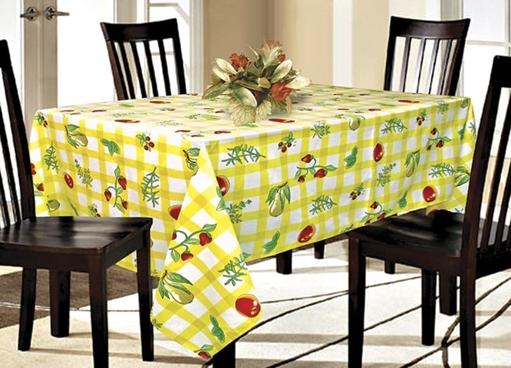} &
    \includegraphics[cframe=blue 15pt,width=0.075\textwidth,height=0.075\textwidth]{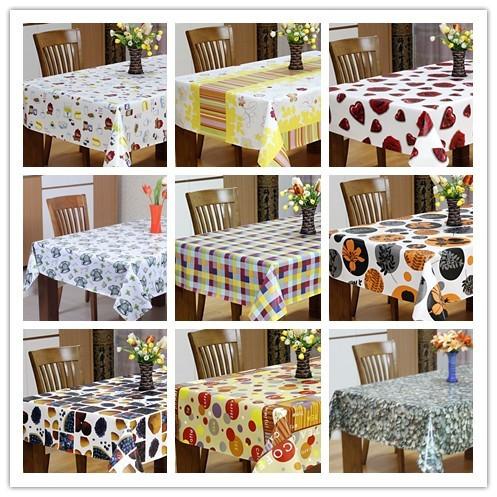} &
    \includegraphics[cframe=blue 15pt,width=0.075\textwidth,height=0.075\textwidth]{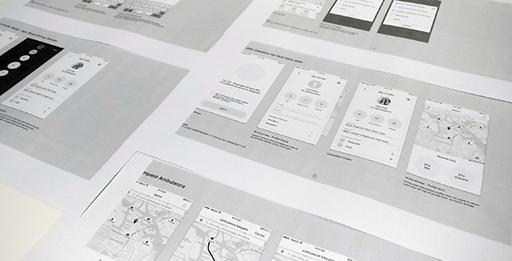} &
    \includegraphics[cframe=blue 15pt,width=0.075\textwidth,height=0.075\textwidth]{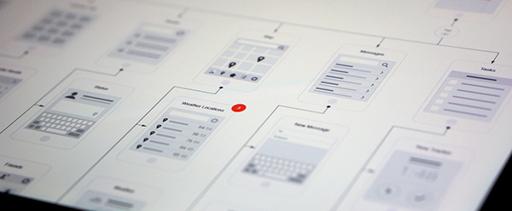} 
    \\

    \midrule

    \multirow{1}{*}[10mm]{\adjustbox{valign=c}{\rotatebox[origin=c]{90}{COCO}}} &
    \includegraphics[cframe=red 15pt,width=0.075\textwidth,height=0.075\textwidth]{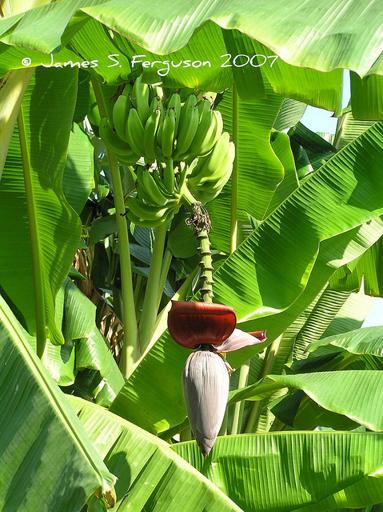} &
    \includegraphics[cframe=red 15pt,width=0.075\textwidth,height=0.075\textwidth]{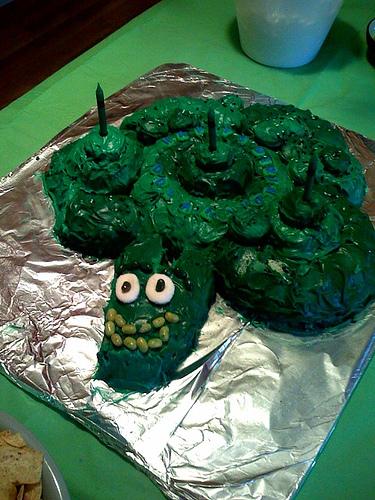} &
    \includegraphics[cframe=red 15pt,width=0.075\textwidth,height=0.075\textwidth]{images/vs_ref/coco/pali_optim/apple/0_ref_4_0.jpg} &
    \includegraphics[cframe=red 15pt,width=0.075\textwidth,height=0.075\textwidth]{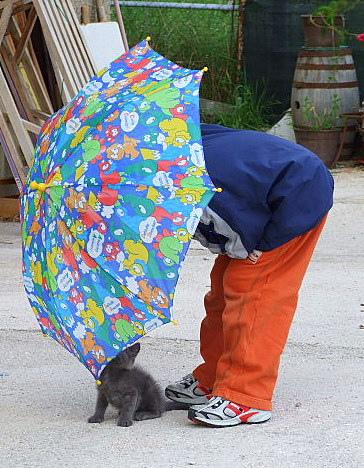} &
    \includegraphics[cframe=red 15pt,width=0.075\textwidth,height=0.075\textwidth]{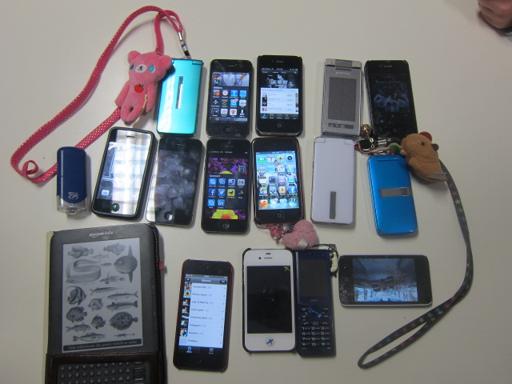} &
    \includegraphics[cframe=red 15pt,width=0.075\textwidth,height=0.075\textwidth]{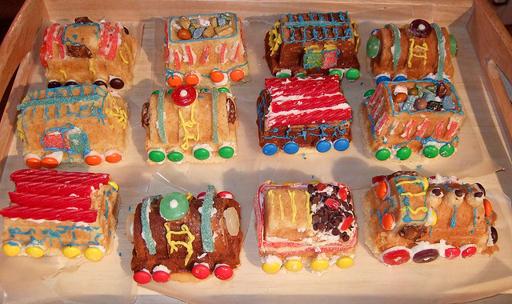} &
    \includegraphics[cframe=red 15pt,width=0.075\textwidth,height=0.075\textwidth]{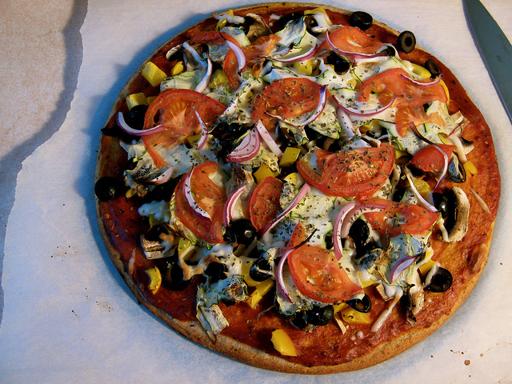} &
    \includegraphics[cframe=red 15pt,width=0.075\textwidth,height=0.075\textwidth]{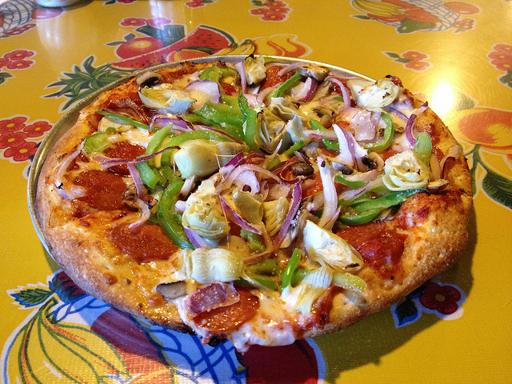} &
    \includegraphics[cframe=red 15pt,width=0.075\textwidth,height=0.075\textwidth]{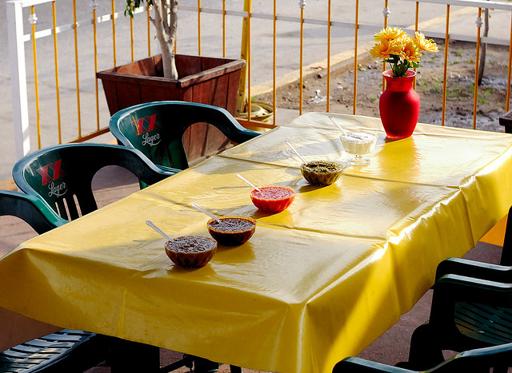} &
    \includegraphics[cframe=red 15pt,width=0.075\textwidth,height=0.075\textwidth]{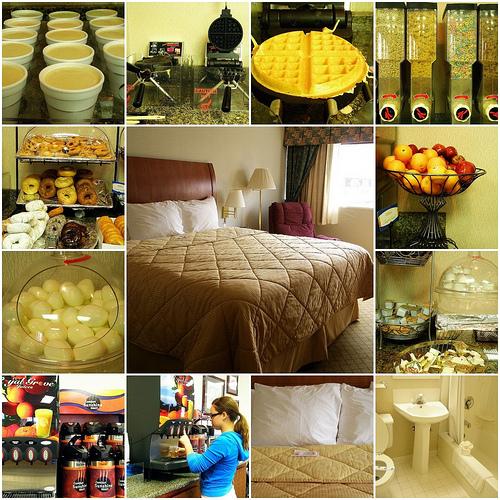} &
    \includegraphics[cframe=red 15pt,width=0.075\textwidth,height=0.075\textwidth]{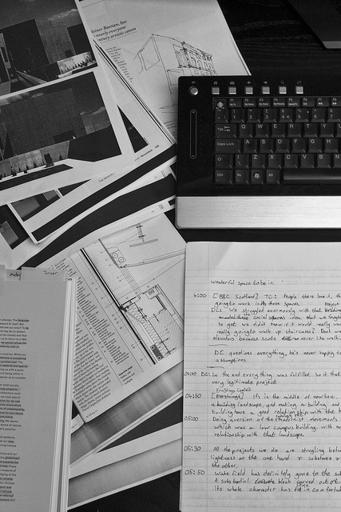} &
    \includegraphics[cframe=red 15pt,width=0.075\textwidth,height=0.075\textwidth]{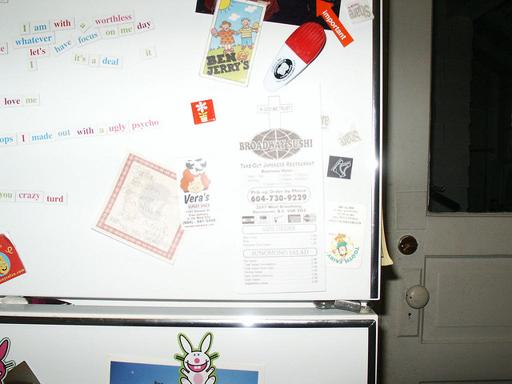} 
    \\
    \bottomrule
    \multicolumn{1}{c}{}    \\

     \multicolumn{13}{c}{\bf{Objects365}} \\
    \toprule
    &
    \multicolumn{2}{c|}{sausage} & 
    \multicolumn{2}{c|}{dessert} & 
    \multicolumn{2}{c|}{durian} &
    \multicolumn{2}{c|}{hanger} &
    \multicolumn{2}{c|}{high heels} &
    \multicolumn{2}{c}{ladder}
    \\
    \midrule
    \multirow{1}{*}[9mm]{\adjustbox{valign=c}{\rotatebox[origin=c]{90}{\ours}}} &
    \includegraphics[cframe=blue 15pt,width=0.075\textwidth,height=0.075\textwidth]{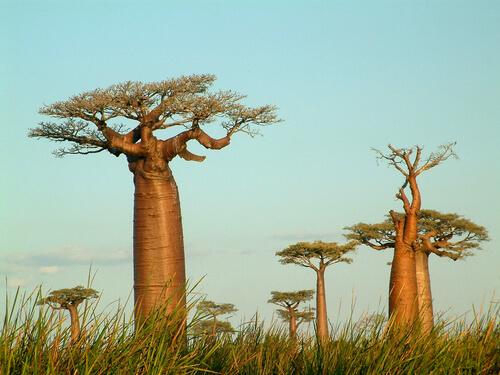} &
    \includegraphics[cframe=blue 15pt,width=0.075\textwidth,height=0.075\textwidth]{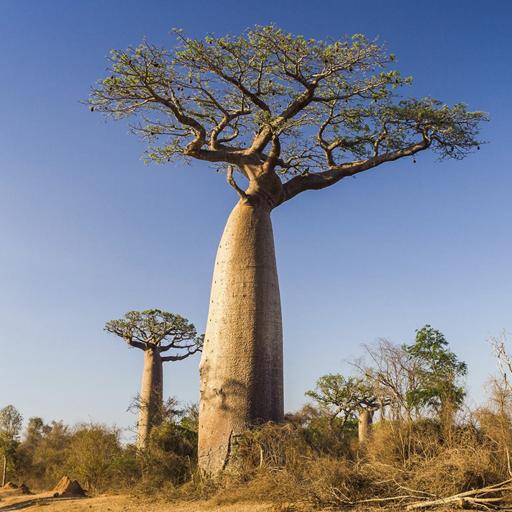} &

    \includegraphics[cframe=blue 15pt,width=0.075\textwidth,height=0.075\textwidth]{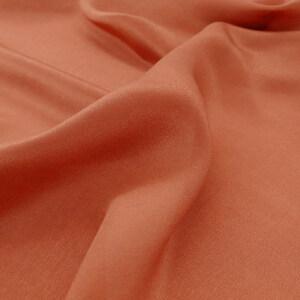} &
    \includegraphics[cframe=blue 15pt,width=0.075\textwidth,height=0.075\textwidth]{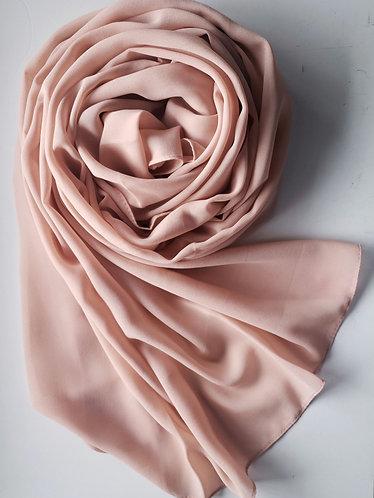} &

    \includegraphics[cframe=blue 15pt,width=0.075\textwidth,height=0.075\textwidth]{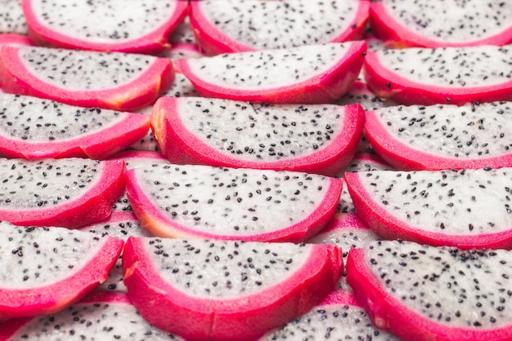} &
    \includegraphics[cframe=blue 15pt,width=0.075\textwidth,height=0.075\textwidth]{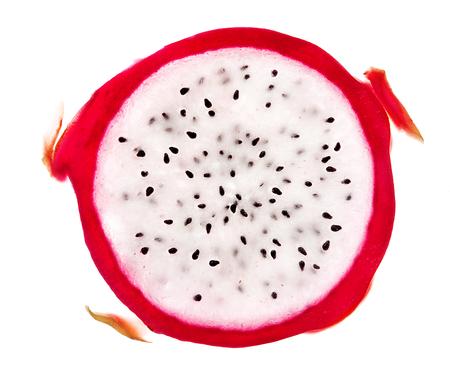} &

    \includegraphics[cframe=blue 15pt,width=0.075\textwidth,height=0.075\textwidth]{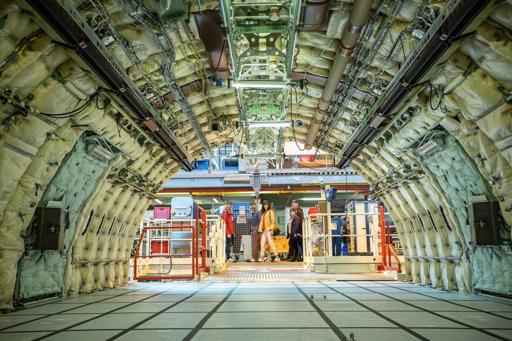} &
    \includegraphics[cframe=blue 15pt,width=0.075\textwidth,height=0.075\textwidth]{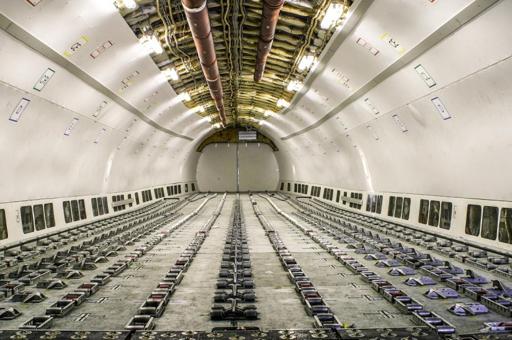} &

    \includegraphics[cframe=blue 15pt,width=0.075\textwidth,height=0.075\textwidth]{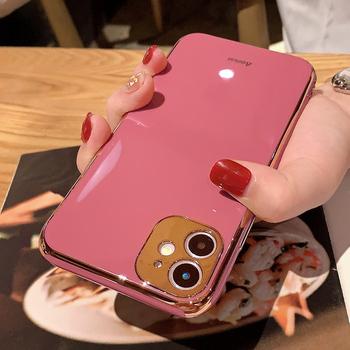} &
    \includegraphics[cframe=blue 15pt,width=0.075\textwidth,height=0.075\textwidth]{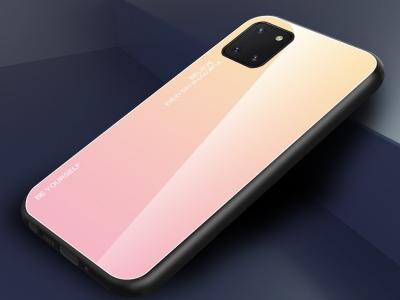} &

    \includegraphics[cframe=blue 15pt,width=0.075\textwidth,height=0.075\textwidth]{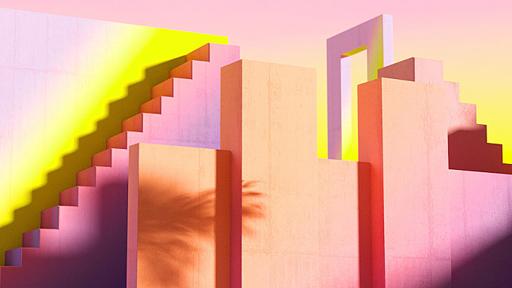} &
    \includegraphics[cframe=blue 15pt,width=0.075\textwidth,height=0.075\textwidth]{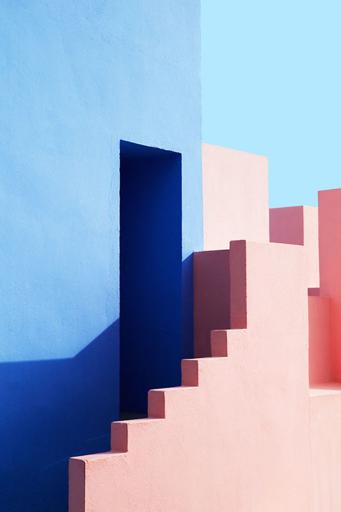} 
    \\
    \midrule
    
    \multirow{1}{*}[10mm]{\adjustbox{valign=c}{\rotatebox[origin=c]{90}{Objects365}}} &
    \includegraphics[cframe=red 15pt,width=0.075\textwidth,height=0.075\textwidth]{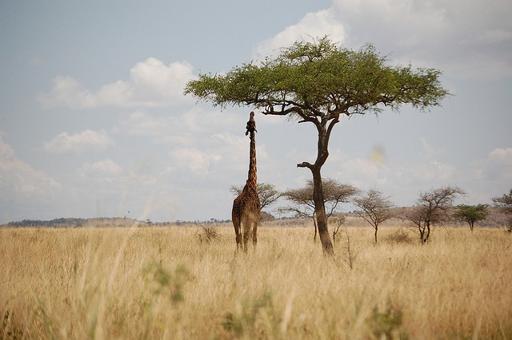} &
    \includegraphics[cframe=red 15pt,width=0.075\textwidth,height=0.075\textwidth]{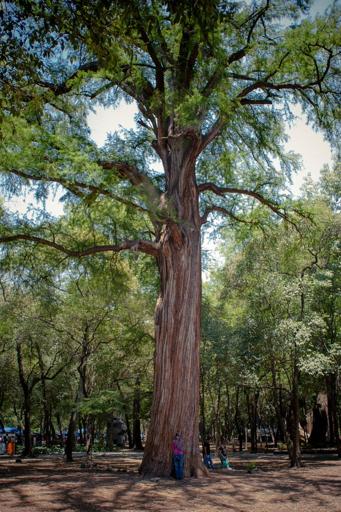} &

    \includegraphics[cframe=red 15pt,width=0.075\textwidth,height=0.075\textwidth]{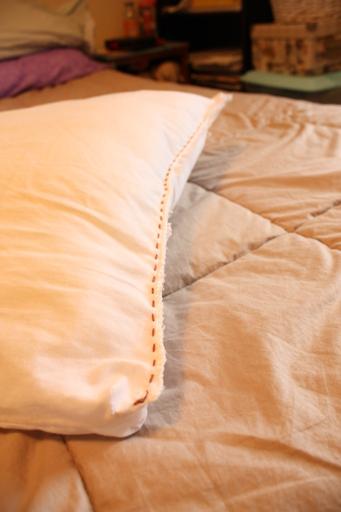} &
    \includegraphics[cframe=red 15pt,width=0.075\textwidth,height=0.075\textwidth]{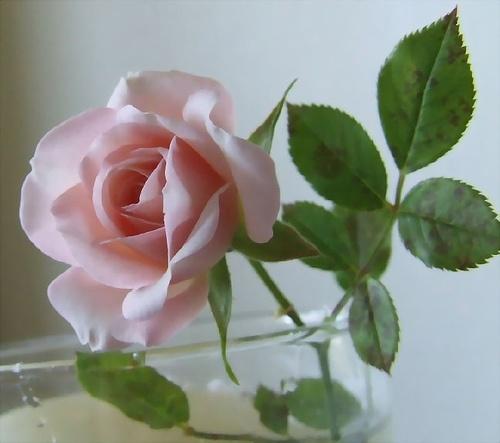} &

    \includegraphics[cframe=red 15pt,width=0.075\textwidth,height=0.075\textwidth]{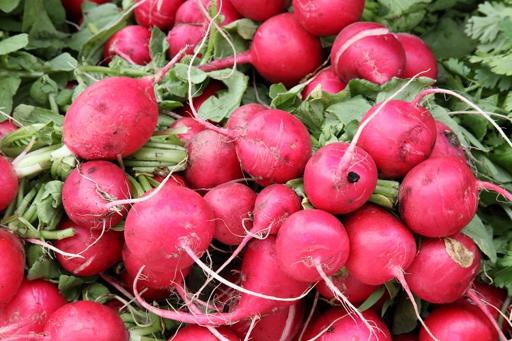} &
    \includegraphics[cframe=red 15pt,width=0.075\textwidth,height=0.075\textwidth]{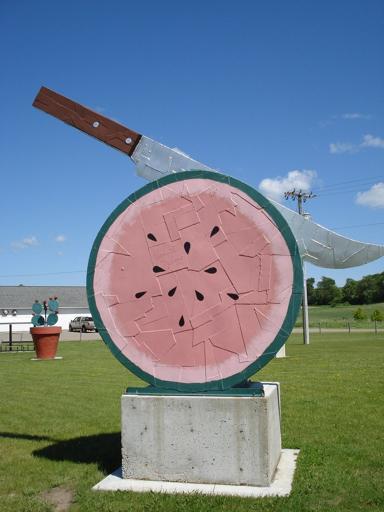} &

    \includegraphics[cframe=blue 15pt,width=0.075\textwidth,height=0.075\textwidth]{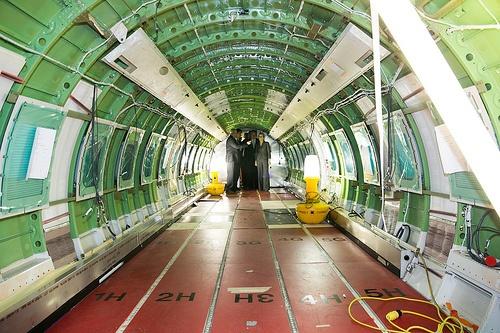} &
    \includegraphics[cframe=red 15pt,width=0.075\textwidth,height=0.075\textwidth]{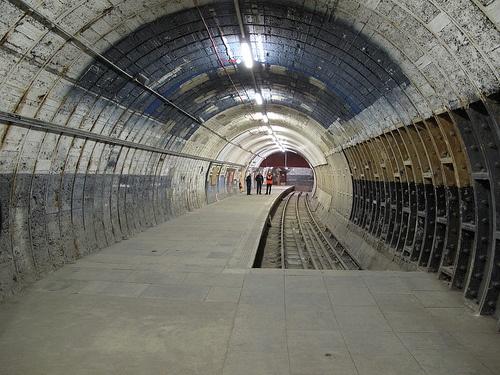} &

    \includegraphics[cframe=red 15pt,width=0.075\textwidth,height=0.075\textwidth]{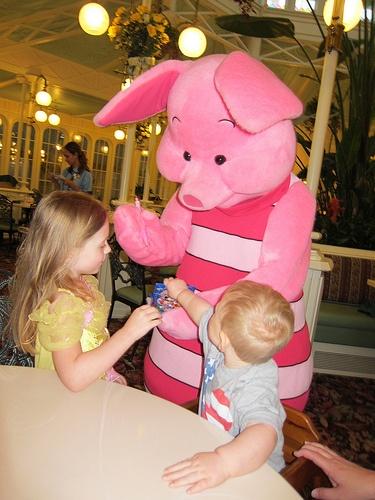} &
    \includegraphics[cframe=red 15pt,width=0.075\textwidth,height=0.075\textwidth]{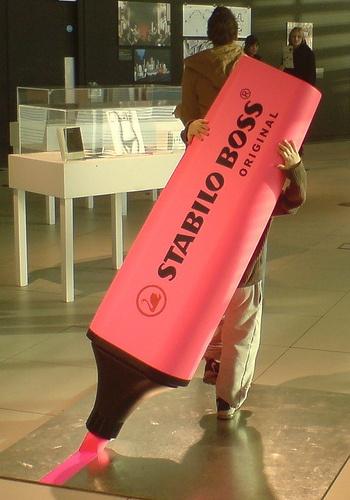} &

    \includegraphics[cframe=red 15pt,width=0.075\textwidth,height=0.075\textwidth]{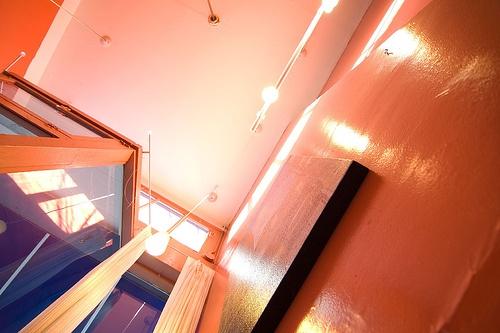} &
    \includegraphics[cframe=red 15pt,width=0.075\textwidth,height=0.075\textwidth]{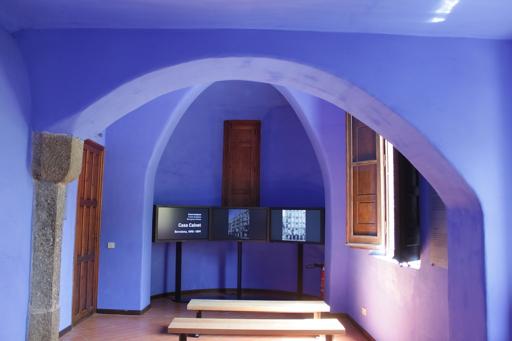} 
    \\
    \bottomrule

    \end{tabular}
    \caption{Demonstration of images that cause \pali to detect the target class, identified using \oursopt, alongside their nearest neighbors in the reference datasets COCO and Objects365. For reference images, we use a blue border to mark images that elicit a "yes" response from the VLM and a red border for a "no" response.  We show that neither the full COCO training set (80K samples) nor Objects365 (1.7M samples) contain the systematic errors uncovered by \ours, as all nearest neighbors are not detected by the VLM. This again highlights that our open-world search in ReLaion-5B is necessary to detect these hallucinations and would not possible even with such a reasonably large dataset such as Object365.  With \ours we find that, \pali incorrectly answers 'yes' for colorful 'wellington boots' as 'apple' and for 'Baobab trees' as 'sausage.' }
    \label{fig:ours_vs_ref}
\end{figure*}

\begin{figure*}[htbp]
    \centering
    
    \begin{subfigure}[t]{0.32\textwidth}
        \centering
        \includegraphics[width=\textwidth]{images/kde_paligemma.pdf}
        \caption{\pali}
    \end{subfigure}
    \hfill
    \begin{subfigure}[t]{0.32\textwidth}
        \centering
        \includegraphics[width=\textwidth]{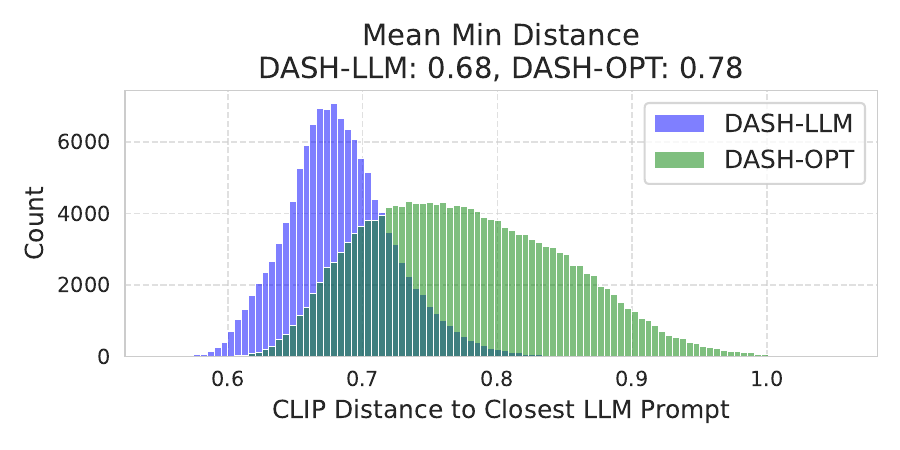}
        \caption{\llavanext \vicuna}
    \end{subfigure}
    \hfill
    \begin{subfigure}[t]{0.32\textwidth}
        \centering
        \includegraphics[width=\textwidth]{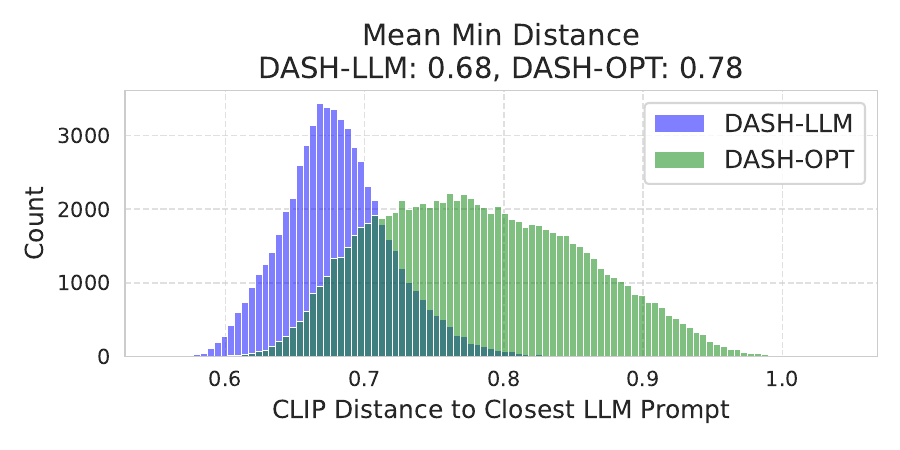}
        \caption{\llavanext \mistral}
    \end{subfigure}
    
    \caption{Extension of \cref{fig:llm_vs_opt} for \pali as well as \llavanext \vicuna and \mistral. For all VLMs \oursopt finds hallucinations which are further away from the original text queries than \oursllm. This illustrates quantitatively the higher diversity of hallucinations found by \oursopt.}
    \label{fig:app_llm_vs_opt_all}
\end{figure*}

\begin{figure*}[htbp]
        \centering
        \footnotesize
        \setlength{\tabcolsep}{1pt} %
        \renewcommand{\arraystretch}{1} %
\begin{tabular}{ccc|ccc|ccc|ccc}
\toprule
\multicolumn{12}{c}{\oursllm - Ptarmigan - Total clusters 7 - Total images 613} \\
\midrule
\multicolumn{3}{c|}{Cluster size: 190} & \multicolumn{3}{c|}{Cluster size: 167} & \multicolumn{3}{c|}{Cluster size: 129} & \multicolumn{3}{c}{Cluster size: 66} \\
\includegraphics[width=0.075\textwidth,height=0.075\textwidth]{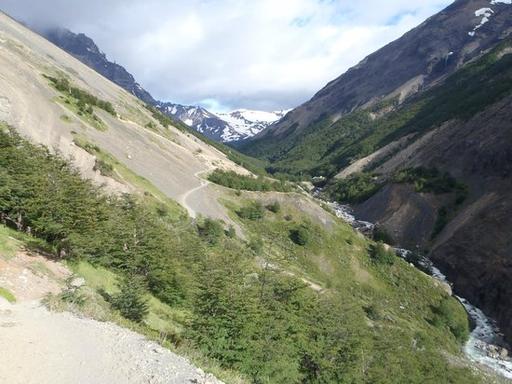} & \includegraphics[width=0.075\textwidth,height=0.075\textwidth]{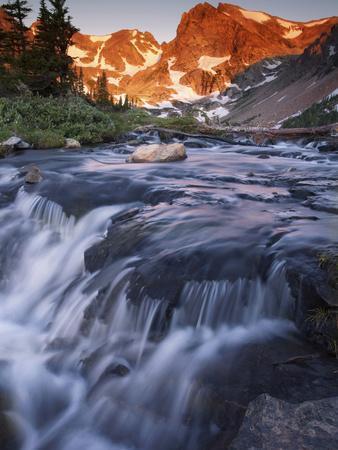} & \includegraphics[width=0.075\textwidth,height=0.075\textwidth]{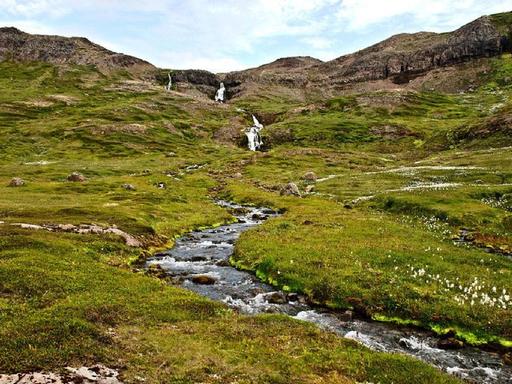} & \includegraphics[width=0.075\textwidth,height=0.075\textwidth]{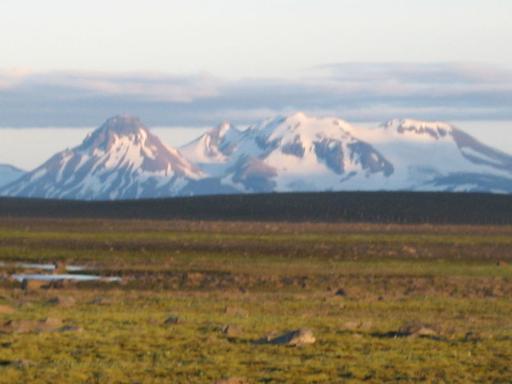} & \includegraphics[width=0.075\textwidth,height=0.075\textwidth]{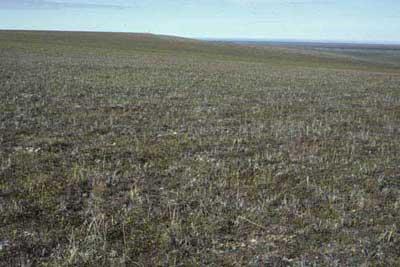} & \includegraphics[width=0.075\textwidth,height=0.075\textwidth]{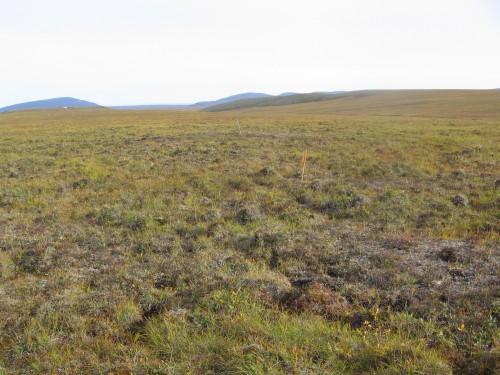} & \includegraphics[width=0.075\textwidth,height=0.075\textwidth]{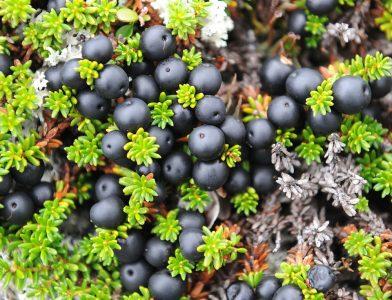} & \includegraphics[width=0.075\textwidth,height=0.075\textwidth]{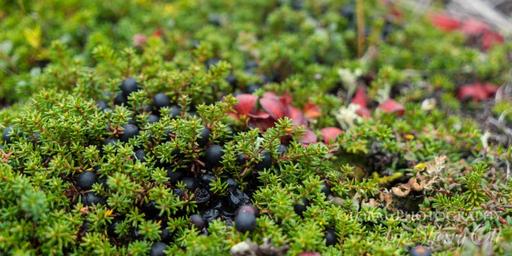} & \includegraphics[width=0.075\textwidth,height=0.075\textwidth]{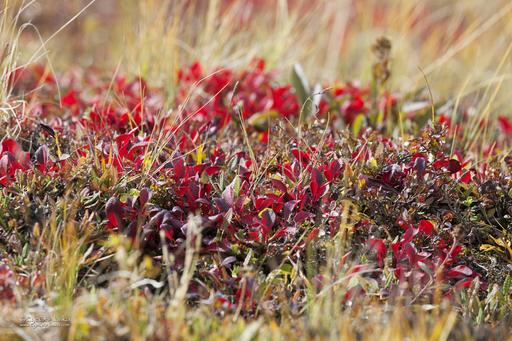} & \includegraphics[width=0.075\textwidth,height=0.075\textwidth]{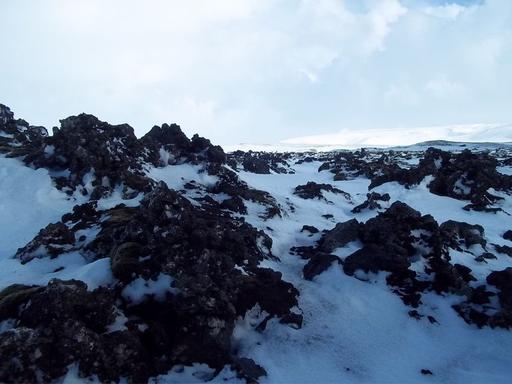} & \includegraphics[width=0.075\textwidth,height=0.075\textwidth]{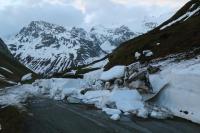} & \includegraphics[width=0.075\textwidth,height=0.075\textwidth]{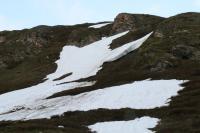} \\
\midrule
\multicolumn{3}{c|}{Cluster size: 28} & \multicolumn{3}{c|}{Cluster size: 26} & \multicolumn{3}{c|}{Cluster size: 7} & \multicolumn{3}{c}{} \\
\includegraphics[width=0.075\textwidth,height=0.075\textwidth]{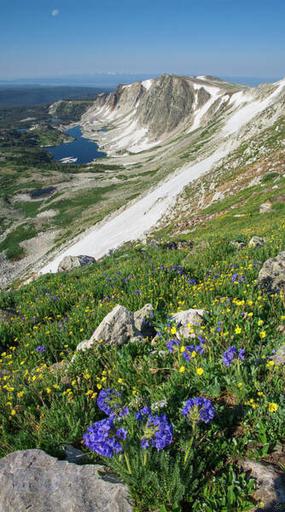} & \includegraphics[width=0.075\textwidth,height=0.075\textwidth]{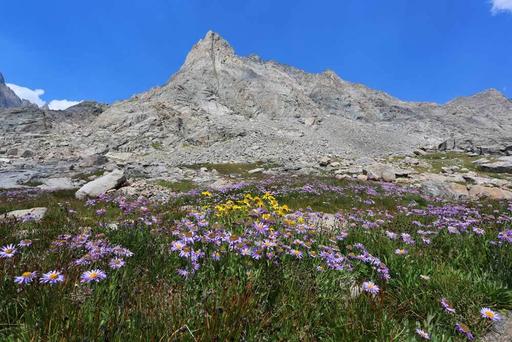} & \includegraphics[width=0.075\textwidth,height=0.075\textwidth]{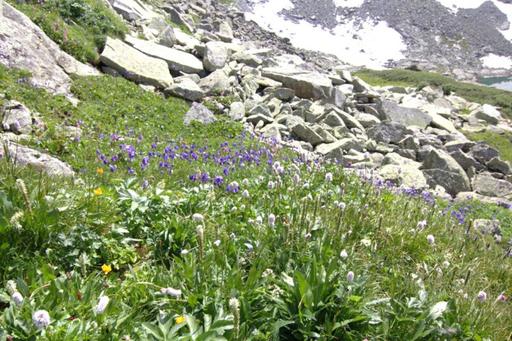} & \includegraphics[width=0.075\textwidth,height=0.075\textwidth]{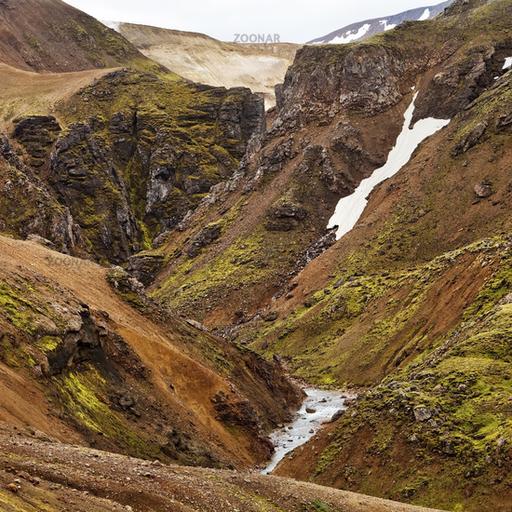} & \includegraphics[width=0.075\textwidth,height=0.075\textwidth]{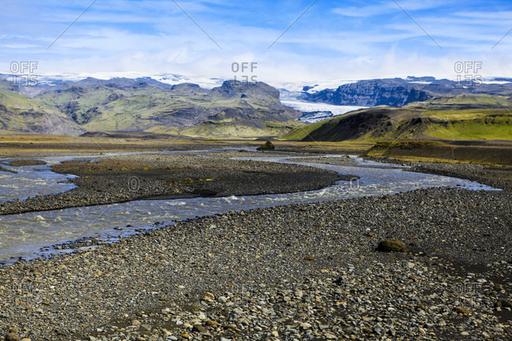} & \includegraphics[width=0.075\textwidth,height=0.075\textwidth]{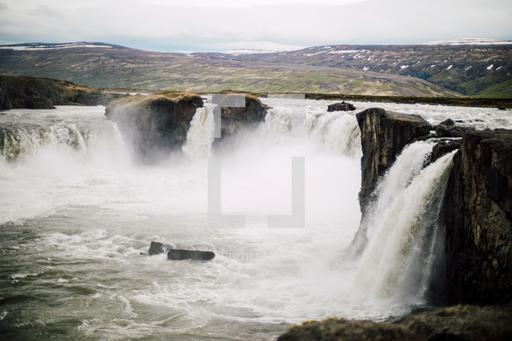} & \includegraphics[width=0.075\textwidth,height=0.075\textwidth]{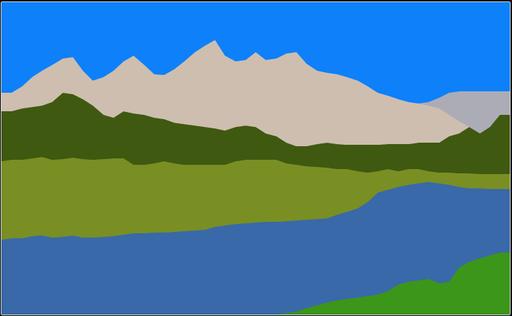} & \includegraphics[width=0.075\textwidth,height=0.075\textwidth]{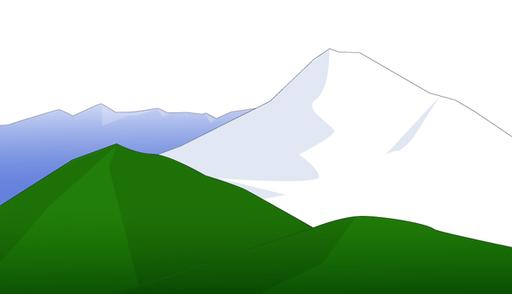} & \includegraphics[width=0.075\textwidth,height=0.075\textwidth]{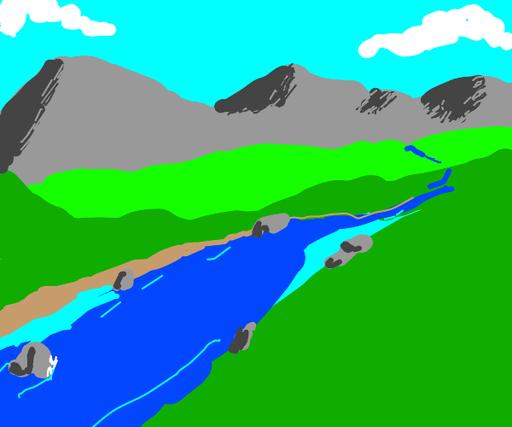} &  &  &  \\
\bottomrule
\toprule
\multicolumn{12}{c}{\oursopt - Ptarmigan - Total clusters 21 -  Total images 5079} \\
\midrule
\multicolumn{3}{c|}{Cluster size: 1556} & \multicolumn{3}{c|}{Cluster size: 904} & \multicolumn{3}{c|}{Cluster size: 721} & \multicolumn{3}{c}{Cluster size: 629} \\
\includegraphics[width=0.075\textwidth,height=0.075\textwidth]{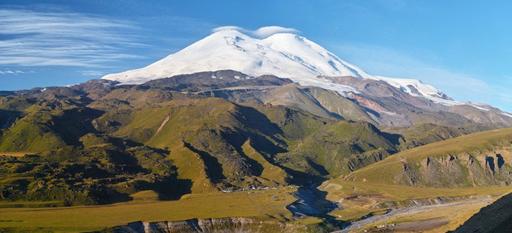} & \includegraphics[width=0.075\textwidth,height=0.075\textwidth]{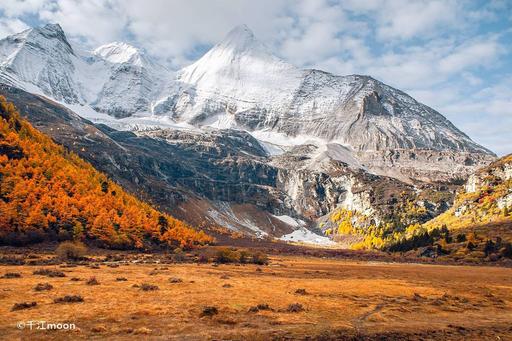} & \includegraphics[width=0.075\textwidth,height=0.075\textwidth]{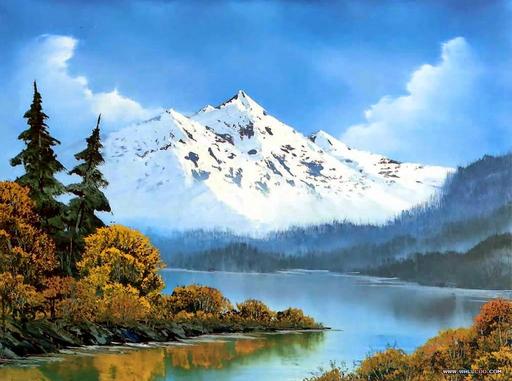} & \includegraphics[width=0.075\textwidth,height=0.075\textwidth]{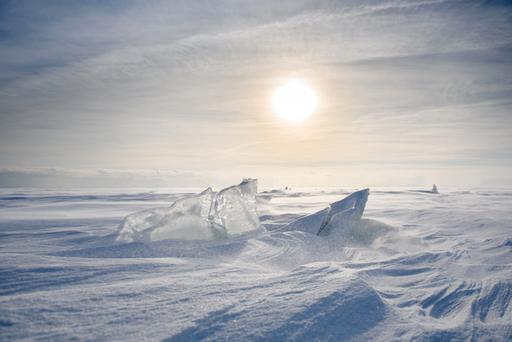} & \includegraphics[width=0.075\textwidth,height=0.075\textwidth]{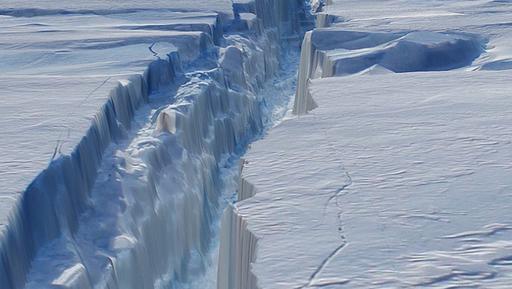} & \includegraphics[width=0.075\textwidth,height=0.075\textwidth]{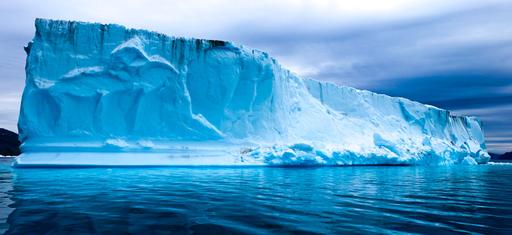} & \includegraphics[width=0.075\textwidth,height=0.075\textwidth]{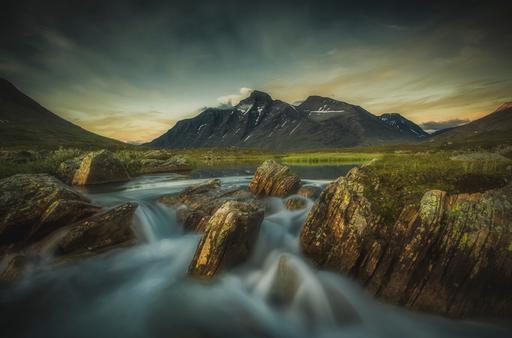} & \includegraphics[width=0.075\textwidth,height=0.075\textwidth]{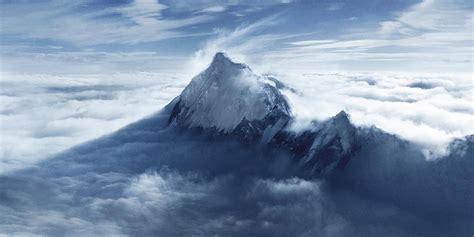} & \includegraphics[width=0.075\textwidth,height=0.075\textwidth]{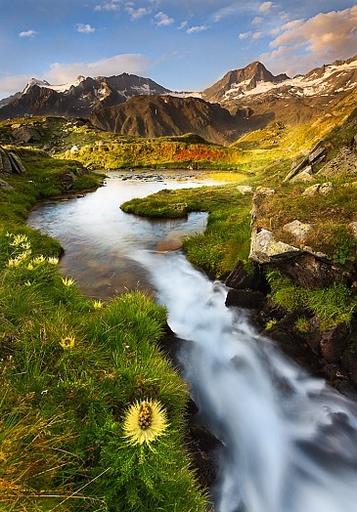} & \includegraphics[width=0.075\textwidth,height=0.075\textwidth]{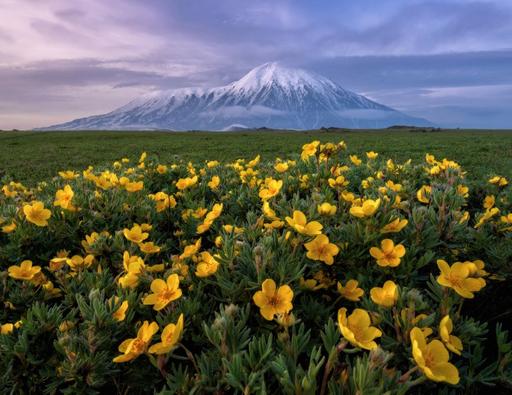} & \includegraphics[width=0.075\textwidth,height=0.075\textwidth]{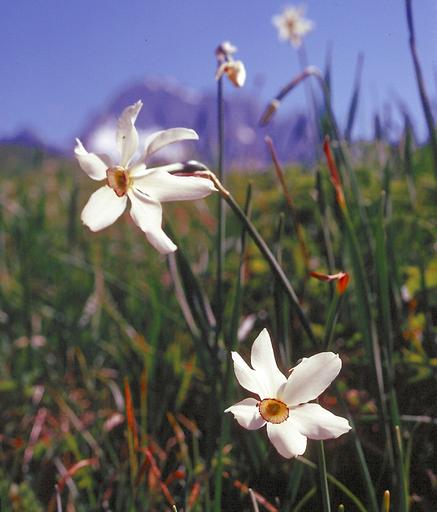} & \includegraphics[width=0.075\textwidth,height=0.075\textwidth]{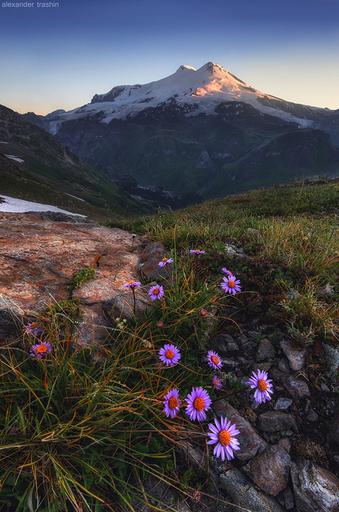} \\
\midrule
\multicolumn{3}{c|}{Cluster size: 188} & \multicolumn{3}{c|}{Cluster size: 177} & \multicolumn{3}{c|}{Cluster size: 173} & \multicolumn{3}{c}{Cluster size: 145} \\
\includegraphics[width=0.075\textwidth,height=0.075\textwidth]{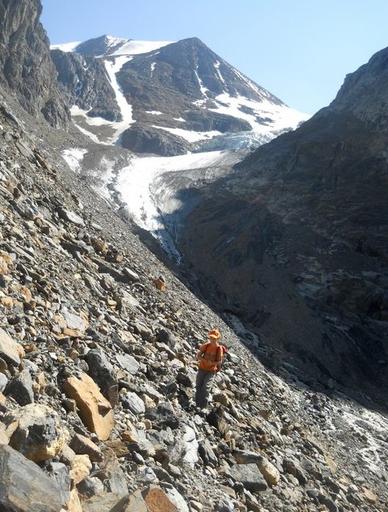} & \includegraphics[width=0.075\textwidth,height=0.075\textwidth]{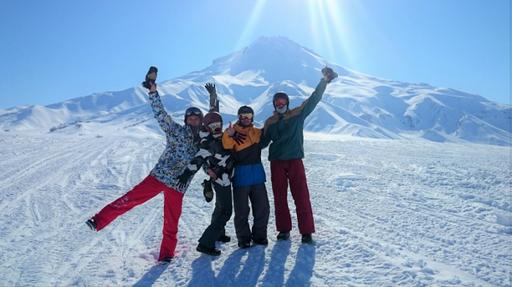} & \includegraphics[width=0.075\textwidth,height=0.075\textwidth]{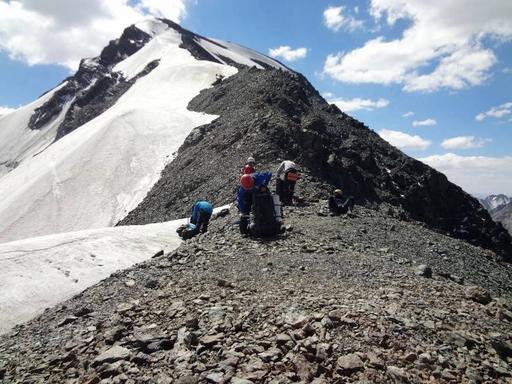} & \includegraphics[width=0.075\textwidth,height=0.075\textwidth]{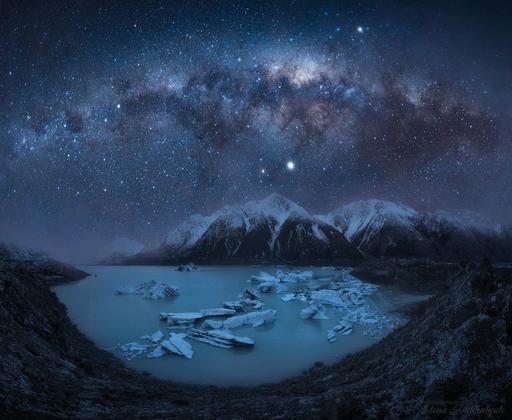} & \includegraphics[width=0.075\textwidth,height=0.075\textwidth]{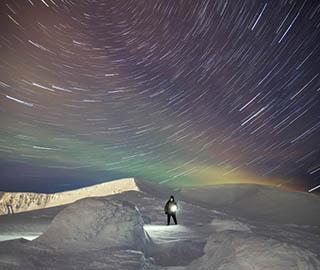} & \includegraphics[width=0.075\textwidth,height=0.075\textwidth]{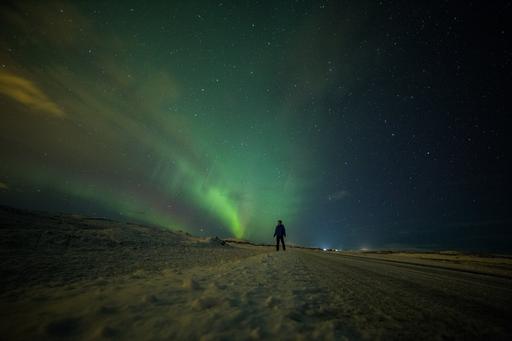} & \includegraphics[width=0.075\textwidth,height=0.075\textwidth]{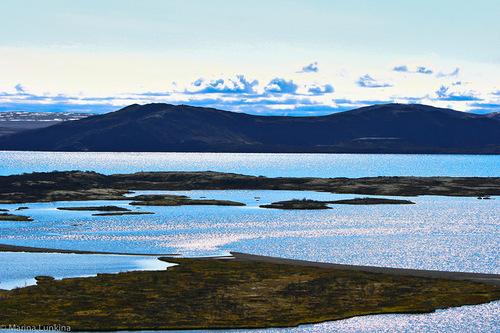} & \includegraphics[width=0.075\textwidth,height=0.075\textwidth]{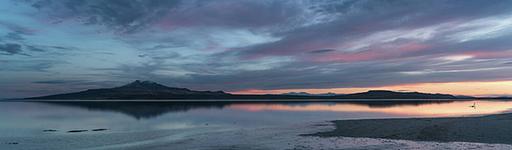} & \includegraphics[width=0.075\textwidth,height=0.075\textwidth]{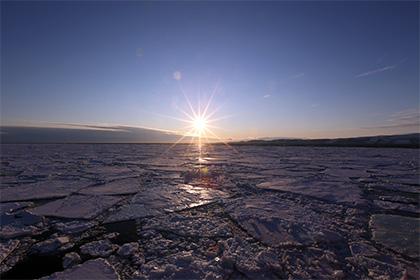} & \includegraphics[width=0.075\textwidth,height=0.075\textwidth]{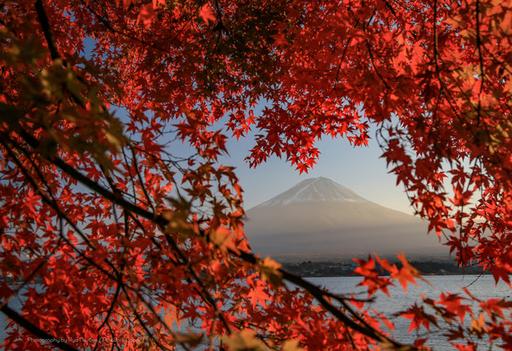} & \includegraphics[width=0.075\textwidth,height=0.075\textwidth]{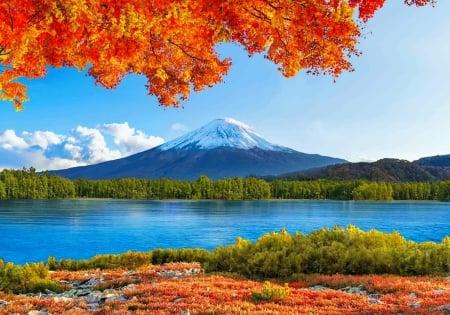} & \includegraphics[width=0.075\textwidth,height=0.075\textwidth]{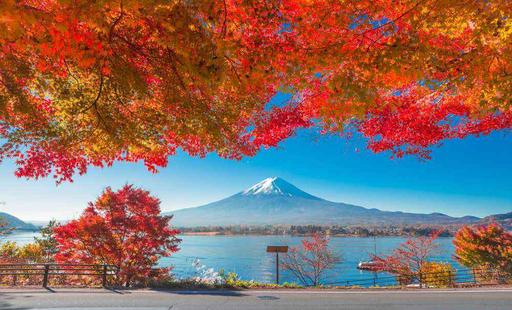} \\
\midrule
\multicolumn{3}{c|}{Cluster size: 129} & \multicolumn{3}{c|}{Cluster size: 82} & \multicolumn{3}{c|}{Cluster size: 81} & \multicolumn{3}{c}{Cluster size: 66} \\
\includegraphics[width=0.075\textwidth,height=0.075\textwidth]{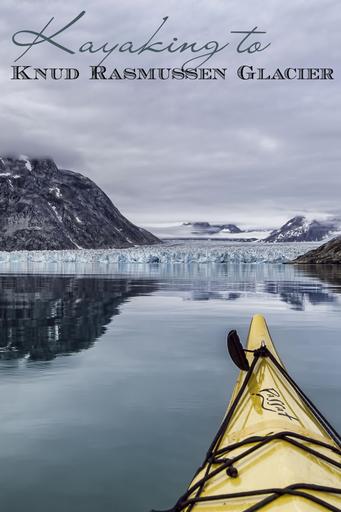} & \includegraphics[width=0.075\textwidth,height=0.075\textwidth]{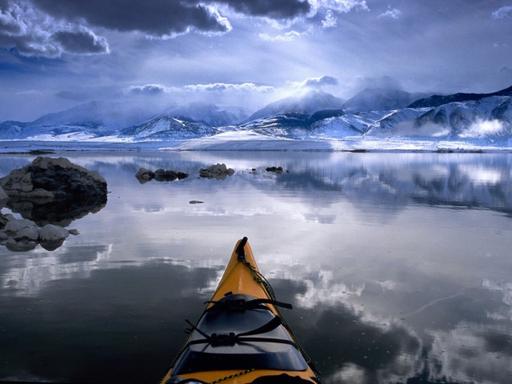} & \includegraphics[width=0.075\textwidth,height=0.075\textwidth]{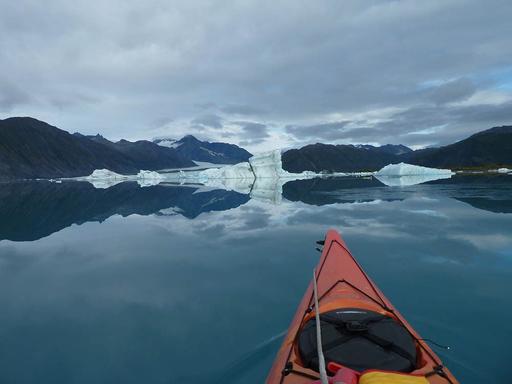} & \includegraphics[width=0.075\textwidth,height=0.075\textwidth]{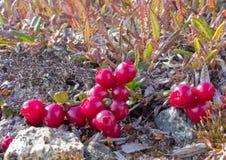} & \includegraphics[width=0.075\textwidth,height=0.075\textwidth]{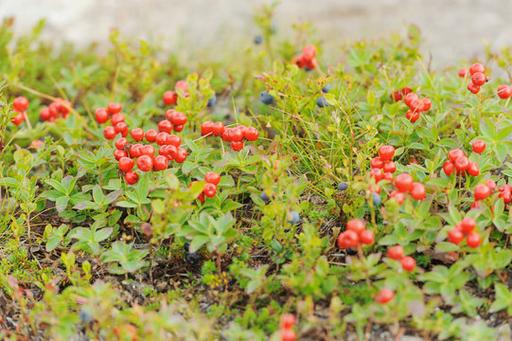} & \includegraphics[width=0.075\textwidth,height=0.075\textwidth]{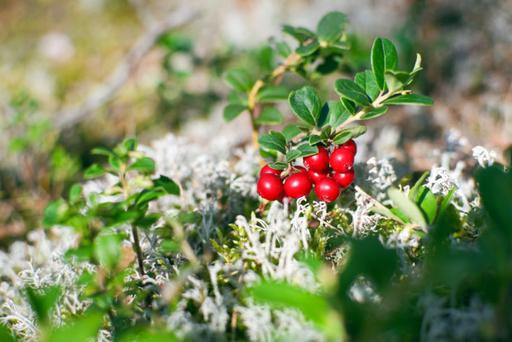} & \includegraphics[width=0.075\textwidth,height=0.075\textwidth]{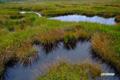} & \includegraphics[width=0.075\textwidth,height=0.075\textwidth]{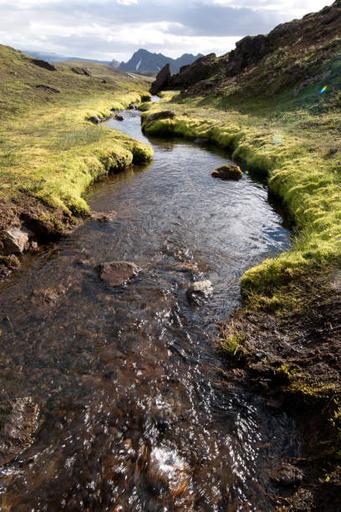} & \includegraphics[width=0.075\textwidth,height=0.075\textwidth]{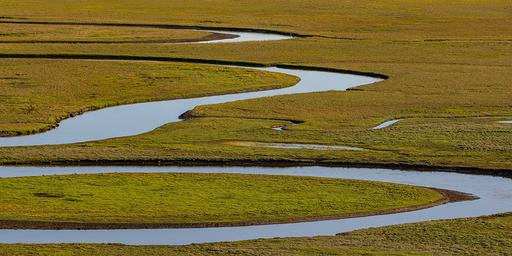} & \includegraphics[width=0.075\textwidth,height=0.075\textwidth]{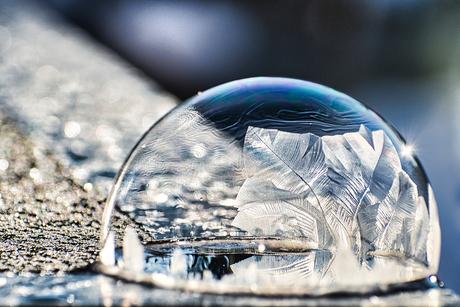} & \includegraphics[width=0.075\textwidth,height=0.075\textwidth]{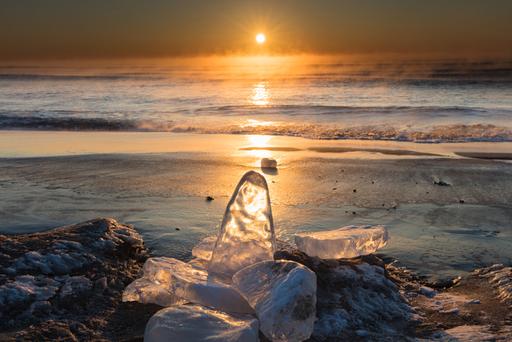} & \includegraphics[width=0.075\textwidth,height=0.075\textwidth]{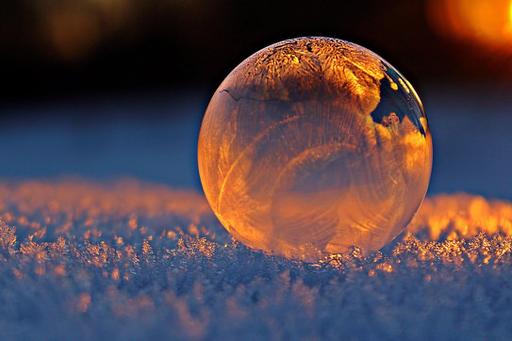} \\
\midrule
\multicolumn{3}{c|}{Cluster size: 57} & \multicolumn{3}{c|}{Cluster size: 39} & \multicolumn{3}{c|}{Cluster size: 37} & \multicolumn{3}{c}{Cluster size: 30} \\
\includegraphics[width=0.075\textwidth,height=0.075\textwidth]{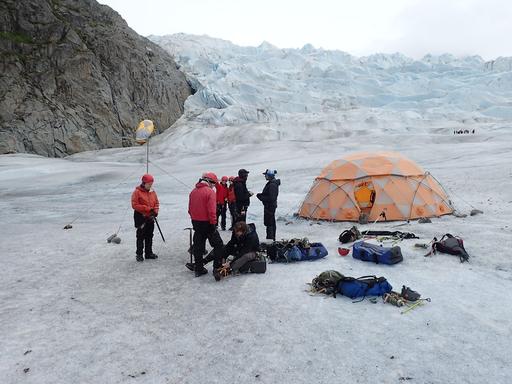} & \includegraphics[width=0.075\textwidth,height=0.075\textwidth]{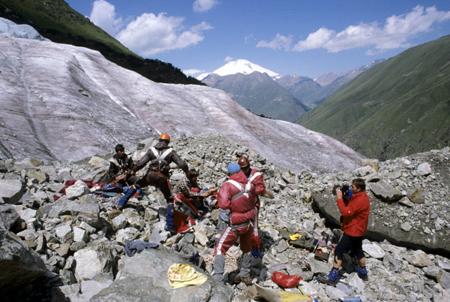} & \includegraphics[width=0.075\textwidth,height=0.075\textwidth]{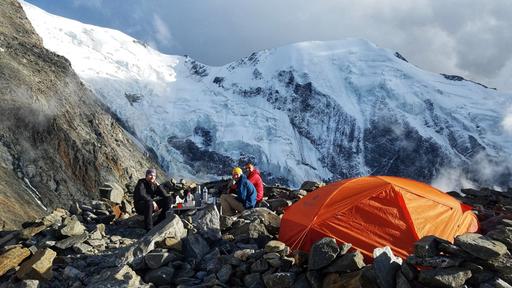} & \includegraphics[width=0.075\textwidth,height=0.075\textwidth]{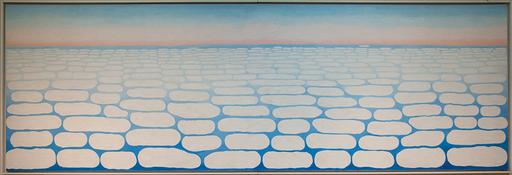} & \includegraphics[width=0.075\textwidth,height=0.075\textwidth]{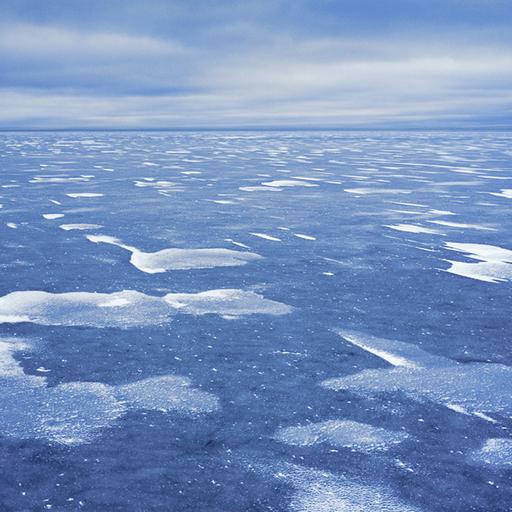} & \includegraphics[width=0.075\textwidth,height=0.075\textwidth]{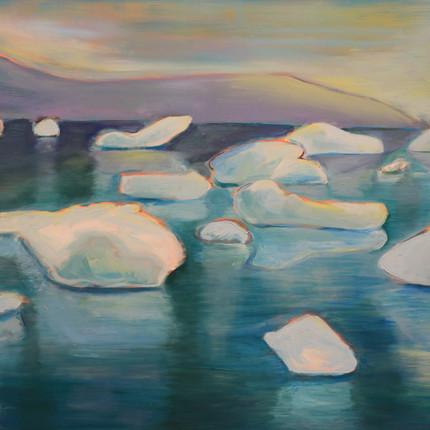} & \includegraphics[width=0.075\textwidth,height=0.075\textwidth]{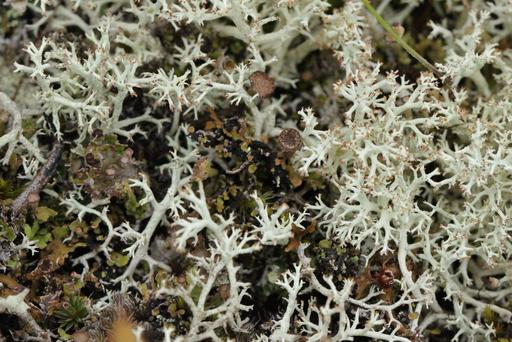} & \includegraphics[width=0.075\textwidth,height=0.075\textwidth]{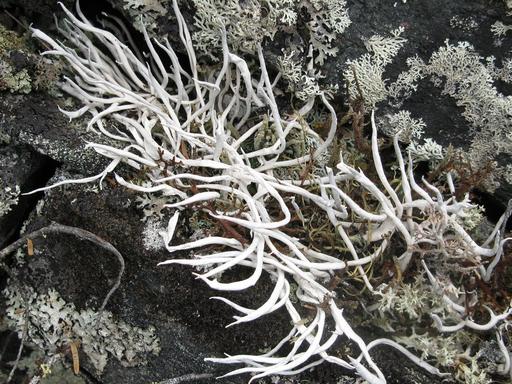} & \includegraphics[width=0.075\textwidth,height=0.075\textwidth]{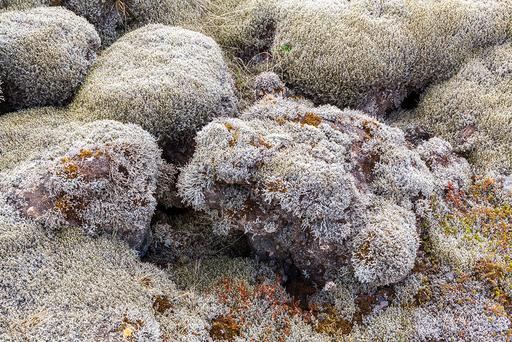} & \includegraphics[width=0.075\textwidth,height=0.075\textwidth]{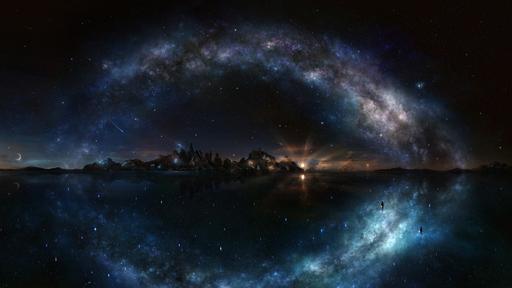} & \includegraphics[width=0.075\textwidth,height=0.075\textwidth]{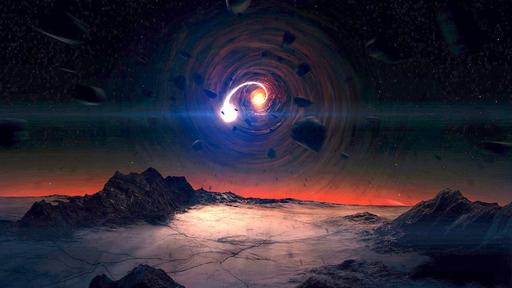} & \includegraphics[width=0.075\textwidth,height=0.075\textwidth]{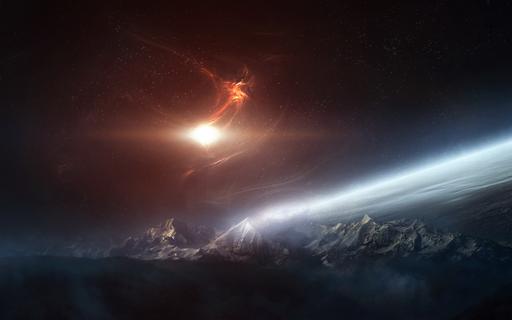} \\
\midrule
\multicolumn{3}{c|}{Cluster size: 22} & \multicolumn{3}{c|}{Cluster size: 20} & \multicolumn{3}{c|}{Cluster size: 10} & \multicolumn{3}{c}{Cluster size: 8} \\
\includegraphics[width=0.075\textwidth,height=0.075\textwidth]{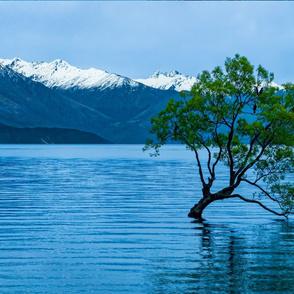} & \includegraphics[width=0.075\textwidth,height=0.075\textwidth]{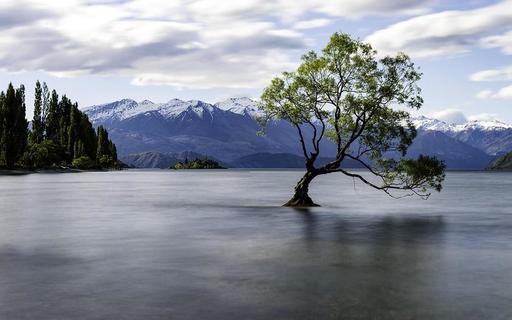} & \includegraphics[width=0.075\textwidth,height=0.075\textwidth]{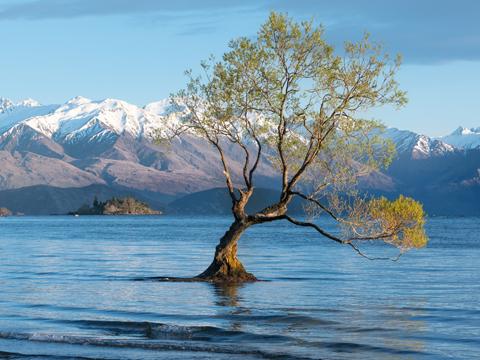} & \includegraphics[width=0.075\textwidth,height=0.075\textwidth]{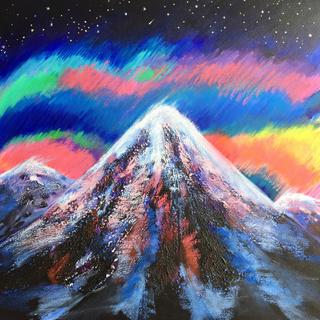} & \includegraphics[width=0.075\textwidth,height=0.075\textwidth]{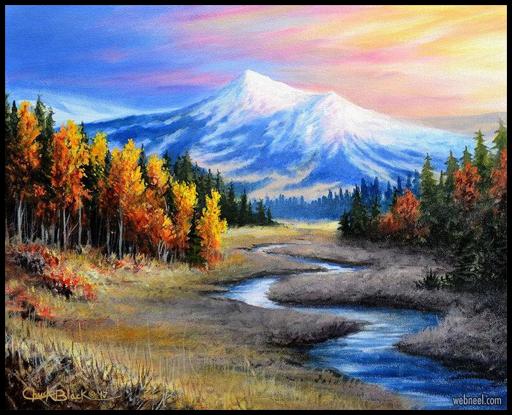} & \includegraphics[width=0.075\textwidth,height=0.075\textwidth]{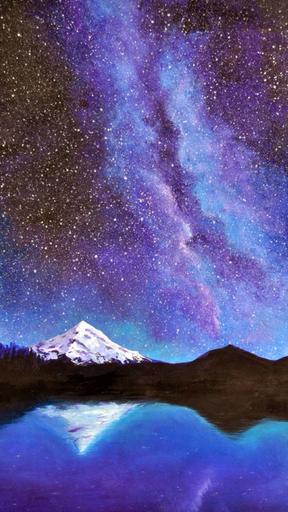} & \includegraphics[width=0.075\textwidth,height=0.075\textwidth]{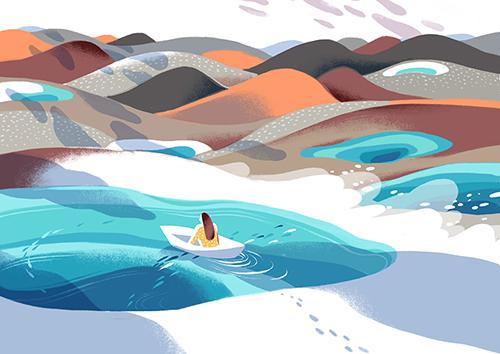} & \includegraphics[width=0.075\textwidth,height=0.075\textwidth]{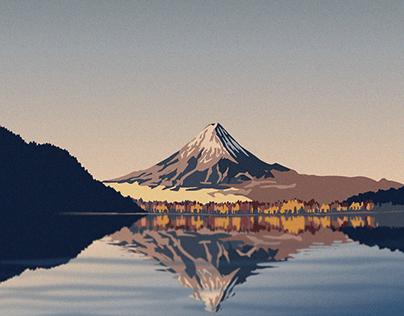} & \includegraphics[width=0.075\textwidth,height=0.075\textwidth]{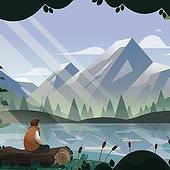} & \includegraphics[width=0.075\textwidth,height=0.075\textwidth]{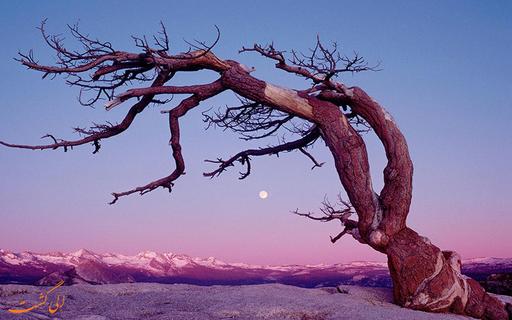} & \includegraphics[width=0.075\textwidth,height=0.075\textwidth]{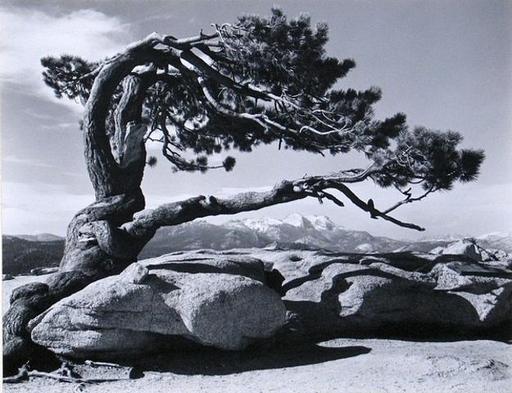} & \includegraphics[width=0.075\textwidth,height=0.075\textwidth]{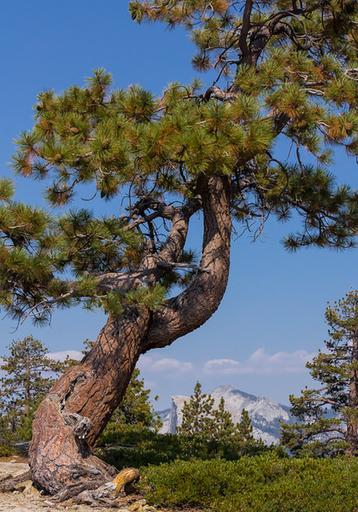} \\
\midrule
\multicolumn{3}{c|}{Cluster size: 5} & \multicolumn{3}{c}{} & \multicolumn{3}{c}{} & \multicolumn{3}{c}{} \\
\includegraphics[width=0.075\textwidth,height=0.075\textwidth]{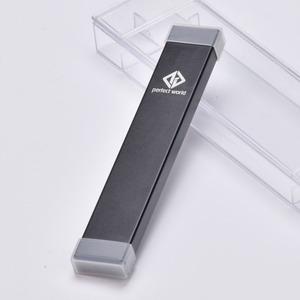} & \includegraphics[width=0.075\textwidth,height=0.075\textwidth]{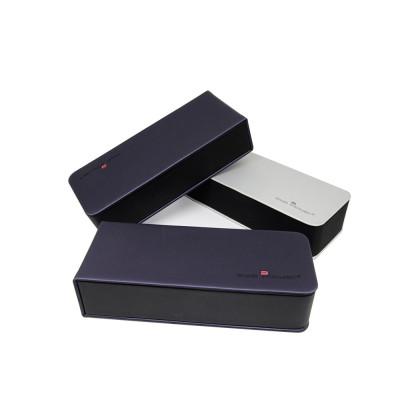} & \includegraphics[width=0.075\textwidth,height=0.075\textwidth]{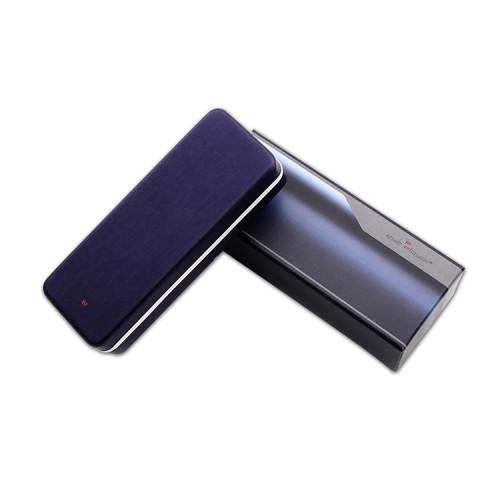}  \\
\bottomrule

\end{tabular}
\caption{All clusters found for \oursllm and \oursopt for \pali and the object ``Ptarmigan.'' While ``Ptarmigan'' refers to the bird species, the clusters include false positives such as images of mountain landscapes, alpine environments, and even abstract artistic representations or completely unrelated objects. This highlights how the VLM's understanding conflates semantic and contextual cues with visual content, leading to hallucinations. In particular, we believe that these hallucinations could be caused by places containing the name ``Ptarmigan,'' such as multiple locations called ``Ptarmigan Peak'' in Colorado, Utah, and Alaska, or ``Ptarmigan Ridge'' and ``Ptarmigan Traverse'' in Washington. While we believe that a VLM should not respond that it sees a ``Ptarmigan'' even in an image of a place with a name containing the word ``Ptarmigan,'' we also checked several of these images to verify that these places are different mountainsides with completely unrelated names. This verifies that the VLM has learned a false representation of the word ``Ptarmigan,'' which includes many different mountainsides or peaks.
Our \oursopt method, leveraging optimized queries, discovers additional ``unknown unknowns,'' such as rare or abstract scenes where a ptarmigan is highly unlikely (e.g., auroras, surreal artwork, and highly stylized objects). By creating queries for the specific target VLM, \oursopt uncovers vulnerabilities that are less intuitive or expected, revealing the VLM's susceptibility to type II hallucinations.
}
\label{fig:app_all_clusters_pali_ptarmigan}
\end{figure*}

\begin{figure*}[htbp]
        \centering
        \footnotesize
        \setlength{\tabcolsep}{0.7pt} %
        \renewcommand{\arraystretch}{1} %
\begin{tabular}{ccc|ccc|ccc|ccc|ccc}
\toprule
\multicolumn{15}{c}{\oursllm - Baumkuchen - Total clusters 17 - Total images 3028} \\
\midrule
\multicolumn{3}{c|}{Cluster size: 900} & \multicolumn{3}{c|}{Cluster size: 488} & \multicolumn{3}{c|}{Cluster size: 352} & \multicolumn{3}{c|}{Cluster size: 350} & \multicolumn{3}{c}{Cluster size: 220} \\
\bottomrule
\end{tabular}
\caption{All clusters found for \oursllm and \oursopt for \llavanext \mistral and the object "Baumkuchen". Please refer to \cref{sec:app_all_clusters} for a description.}
\label{fig:app_all_clusters_baumkuchen}
\end{figure*}

\subsection{All Clusters Visualizations}\label{sec:app_all_clusters}
In \cref{fig:app_all_clusters_pali_ptarmigan}, we present all clusters identified for \oursllm and \oursopt for the object ``Ptarmigan.'' While ``Ptarmigan'' refers to a bird species, the clusters reveal a range of false positives, including images of mountain landscapes, alpine environments, abstract artistic representations, and completely unrelated objects. This indicates that the VLM conflates semantic and contextual cues with visual content, leading to systematic hallucinations.

Interestingly, many of these errors may stem from the existence of places named ``Ptarmigan,'' such as ``Ptarmigan Peak'' in Colorado, Utah, and Alaska, or ``Ptarmigan Ridge'' and ``Ptarmigan Traverse'' in Washington. Even though these locations are unrelated to the bird, the VLM erroneously associates them with the object ``Ptarmigan.'' Our analysis confirms that these places are distinct mountainsides with unrelated names, demonstrating that the VLM has learned a flawed representation of ``Ptarmigan'' that includes a variety of unrelated mountainous scenes.

Additionally, \oursopt uncovers further ``unknown unknowns,'' such as rare or abstract scenes where a ptarmigan is highly unlikely, including auroras, surreal artwork, and stylized objects. 

In \cref{fig:app_all_clusters_baumkuchen}, we present all clusters found for \oursllm and \oursopt for the object ``Baumkuchen'' on \llavanext \mistral. For \oursllm,  we observe that the clusters include no images of Baumkuchen, a traditional German layered cake, but a variety of unrelated objects and scenes. These false positives encompass German cultural artifacts, traditional buildings, festivals, and abstract artistic representations, indicating that the VLM has conflated ``Baumkuchen'' with broader semantic or cultural cues tied to German traditions.

\oursopt uncovers additional ``unknown unknowns''. Alongside unrelated cultural goods like Christmas decorations, traditional crafts, and books which we have also found for \oursllm, we also find additional systematic vulnerabilities. For example, we find a cluster of 111 images containing fountain pens, but also a cluster of 66 images containing wooden kitchen utensils. We also note that the cluster of size 8 which contains cake does not contain any images of ``Baumkuchen''.

\subsection{\ours vs Reference Datasets}\label{sec:app_ours_vs_reference}
As stated in the main paper, we also compare our \ours images to reference datasets such as COCO or Objects365 which are commonly used to construct hallucination benchmarks. In \cref{fig:ours_vs_ref}, we demonstrate images that cause \pali to detect the target class, identified using \oursopt, alongside their nearest neighbors from the reference datasets COCO and Objects365. 
We observe that neither the full COCO training set (80K samples) nor Objects365 (1.7M samples) contain the systematic errors uncovered by \ours, as all nearest neighbors are not detected by the VLM. This highlights that our open-world search in ReLAION-5B is necessary to detect these hallucinations, which would not be possible even with reasonably large datasets like Objects365. Specifically, with \ours, we find that \pali incorrectly answers ``yes'' for colorful ``Wellington boots'' as ``apple'' and for ``Baobab trees'' as ``sausage.''

These examples illustrate the limitations of relying solely on existing datasets for identifying hallucinations in VLMs. In particular, just because a target object is contained in a dataset like COCO or Objects365, as are all examples presented in \cref{fig:ours_vs_ref}, does not guarantee that objects that are not contained in this dataset cannot cause a VLM to hallucinate the target object.  Our method uncovers novel failure cases that are absent in standard benchmarks, emphasizing the importance of an open-world search strategy for comprehensive evaluation.

\subsection{Larger exploration range of \oursopt over \oursllm}
In \cref{fig:app_llm_vs_opt_all}, we present extended version of \cref{fig:llm_vs_opt} for \pali, \llavanext \vicuna, and \llavanext \mistral which demonstrates that \oursopt achieves a greater diversity of images than \oursllm.

\section{Object Detector: False Negative Rate vs False Positive Rate}\label{app:objectdetector}
\begin{figure*}[htbp]
    \centering
    \includegraphics[width=1.0\textwidth]{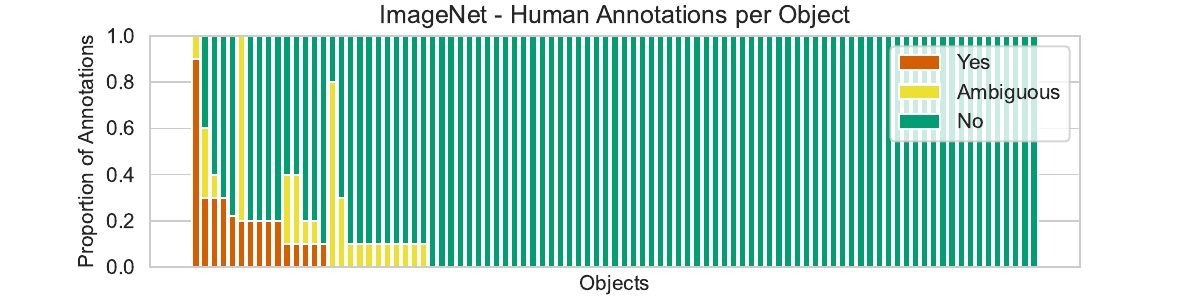}
    \includegraphics[width=1.0\textwidth]{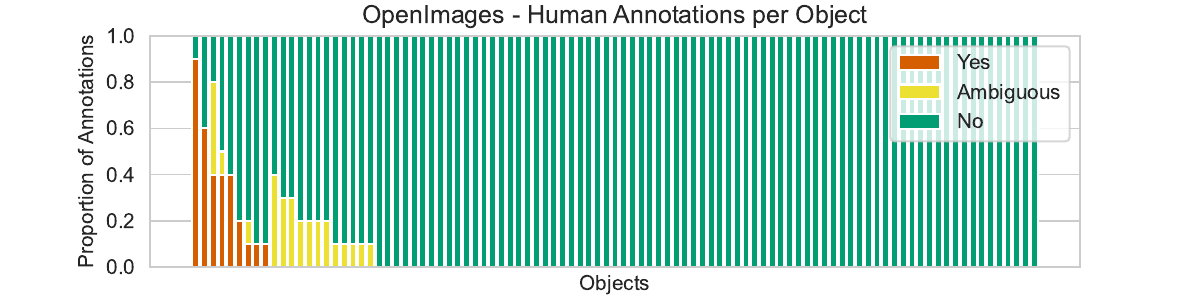}
    \includegraphics[width=1.0\textwidth]{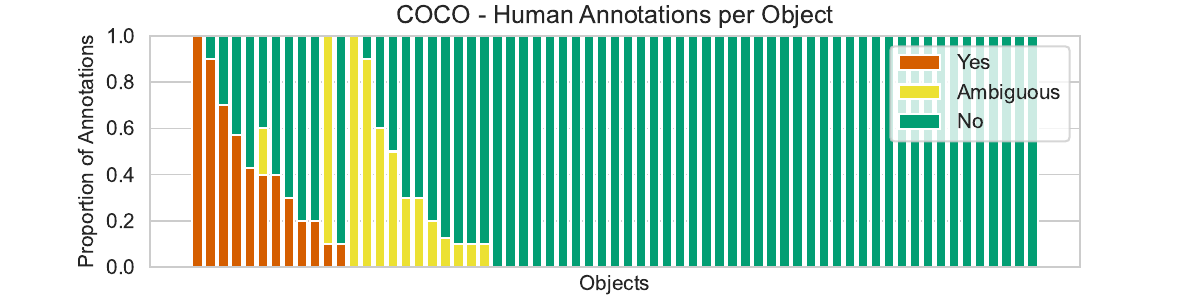}
    \includegraphics[width=1.0\textwidth]{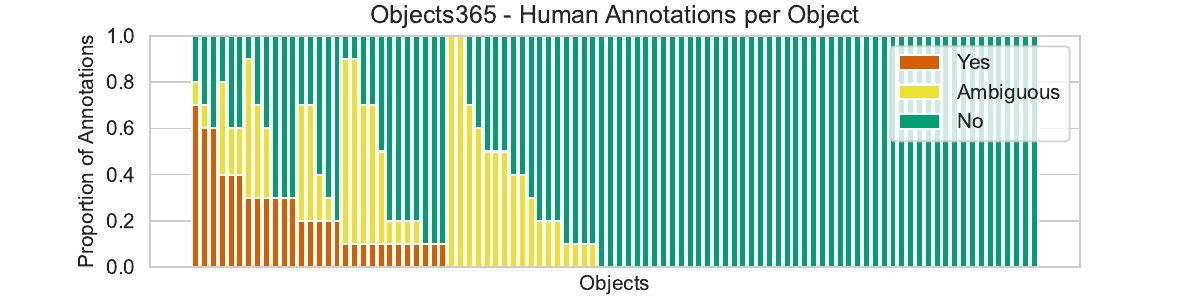}
    \caption{"yes", "no" an "ambiguous" rates in our human evaluation for the 4 datasets used for object labels in our evaluation. Each bar represents one object, for which we manually labeled 10 images for \oursopt on \pali. We note that most objects do not contain any instances of the object and instead, most errors come from few object categories where the object detector itself has a systematic issue. We show qualitative examples in \cref{fig:app_detection_ambiguous} and \cref{fig:app_detection_yes}.}
    \label{fig:app_detection_bars}
\end{figure*}

\begin{figure*}[htbp]
    \centering
    \setlength{\tabcolsep}{1pt} %
    \renewcommand{\arraystretch}{1} %
    \begin{tabular}{cc|cc|cc|cc}
        \toprule
        \multicolumn{4}{c|}{ImageNet} &
        \multicolumn{4}{c|}{OpenImages} \\
        \midrule
        \multicolumn{2}{c|}{"Barn"} & 
        \multicolumn{2}{c|}{"Bookshop"} & 
        \multicolumn{2}{c|}{"Kai yang"} &
        \multicolumn{2}{c}{"Cowry"} 
        \\
         \includegraphics[width=0.12\textwidth,height=0.12\textwidth]{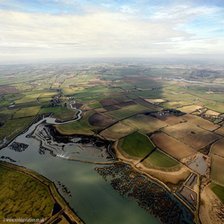} & 
         \includegraphics[width=0.12\textwidth,height=0.12\textwidth]{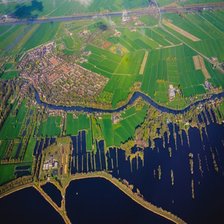} & 
         \includegraphics[width=0.12\textwidth,height=0.12\textwidth]{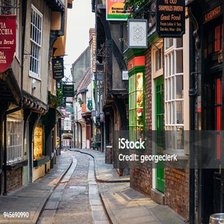} & 
         \includegraphics[width=0.12\textwidth,height=0.12\textwidth]{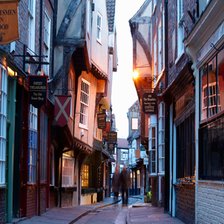} & 
         \includegraphics[width=0.12\textwidth,height=0.12\textwidth]{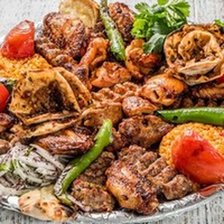} & 
         \includegraphics[width=0.12\textwidth,height=0.12\textwidth]{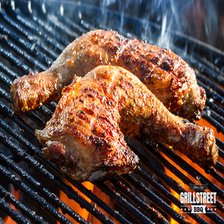} & 
         \includegraphics[width=0.12\textwidth,height=0.12\textwidth]{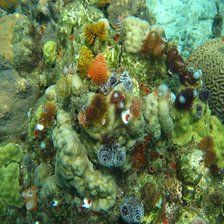} & 
         \includegraphics[width=0.12\textwidth,height=0.12\textwidth]{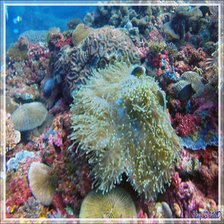}
         \\
        \bottomrule
         
         \\
         \toprule
         \multicolumn{4}{c|}{COCO} &
        \multicolumn{4}{c|}{Objects365} \\
        \midrule
        \multicolumn{2}{c}{"Airplane"} & 
        \multicolumn{2}{c|}{"Train"} & 
        \multicolumn{2}{c}{"Glasses"} &
        \multicolumn{2}{c}{"Soccer"} \\
         \includegraphics[width=0.12\textwidth,height=0.12\textwidth]{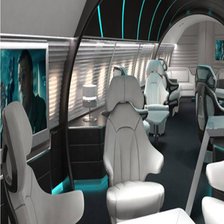} & 
         \includegraphics[width=0.12\textwidth,height=0.12\textwidth]{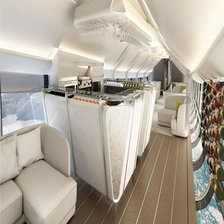} & 
         \includegraphics[width=0.12\textwidth,height=0.12\textwidth]{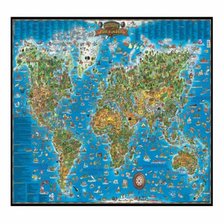} & 
         \includegraphics[width=0.12\textwidth,height=0.12\textwidth]{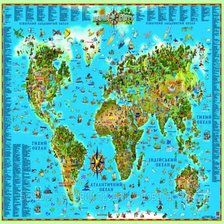} & 
         \includegraphics[width=0.12\textwidth,height=0.12\textwidth]{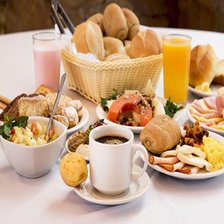} & 
         \includegraphics[width=0.12\textwidth,height=0.12\textwidth]{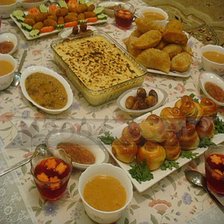} & 
         \includegraphics[width=0.12\textwidth,height=0.12\textwidth]{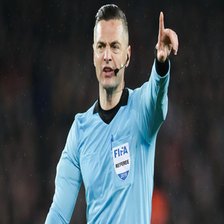} & 
         \includegraphics[width=0.12\textwidth,height=0.12\textwidth]{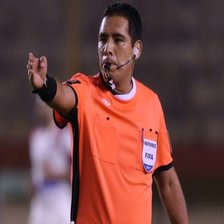} 
         \\
        \bottomrule
    \end{tabular}
    \caption{Examples of images labeled as "ambiguous" in our human evaluation are shown. For "Barn" and "Bookshop," the limited image resolution makes it difficult to identify the objects in the image; for instance, distinguishing a house from a barn in an aerial view is nearly impossible. For "Kai yang," a Thai dish with chicken, while the depicted dishes might contain chicken, it is challenging to determine whether they are specifically "Kai yang." Notably, a reverse image search labels the first image as "kebab." 
For "cowry," small sea snails, even if human labelers could not identify any, it is difficult to guarantee their absence in the image. For "airplane," the interiors shown could represent futuristic airplane or train designs. For "train," the image resolution is too low to infer the presence of specific objects. 
For the two objects from Objects365, the ambiguity mainly arises from the object labels themselves. For example, "glasses" in the dataset refers to eyewear, but the images often contain multiple glass objects. Similarly, "soccer" refers exclusively to a soccer ball in Objects365, whereas the sport itself is not a well-defined object. This creates ambiguity about whether we should label referees as "yes" or "no."
}
    \label{fig:app_detection_ambiguous}
\end{figure*}

\begin{figure*}[htbp]
    \centering
    \setlength{\tabcolsep}{1pt} %
    \renewcommand{\arraystretch}{1} %
    \begin{tabular}{cc|cc|cc|cc}
        \toprule
        \multicolumn{4}{c|}{ImageNet} &
        \multicolumn{4}{c|}{OpenImages} \\
        \midrule
        \multicolumn{2}{c|}{"Mountain bike" } & 
        \multicolumn{2}{c|}{"Pot"} & 
        \multicolumn{2}{c|}{"Car"} &
        \multicolumn{2}{c}{"Car seat"} 
        \\
         \includegraphics[width=0.12\textwidth,height=0.12\textwidth]{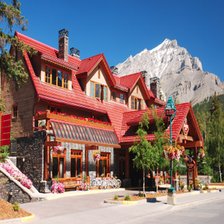} & 
         \includegraphics[width=0.12\textwidth,height=0.12\textwidth]{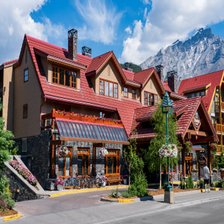} & 
         \includegraphics[width=0.12\textwidth,height=0.12\textwidth]{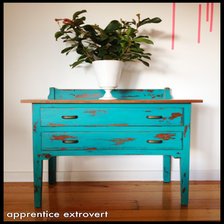} & 
         \includegraphics[width=0.12\textwidth,height=0.12\textwidth]{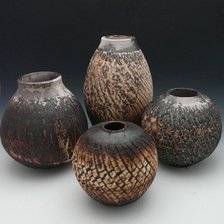} & 
         \includegraphics[width=0.12\textwidth,height=0.12\textwidth]{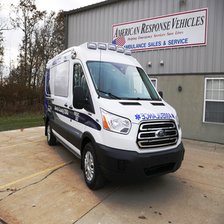} & 
         \includegraphics[width=0.12\textwidth,height=0.12\textwidth]{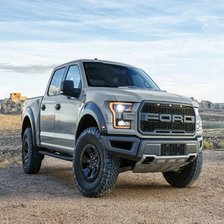} & 
         \includegraphics[width=0.12\textwidth,height=0.12\textwidth]{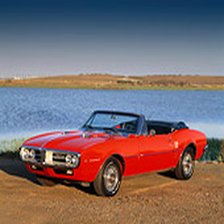} & 
         \includegraphics[width=0.12\textwidth,height=0.12\textwidth]{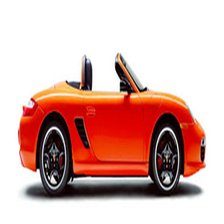} 
         \\
        \bottomrule
         \\
         \toprule
        \multicolumn{4}{c|}{COCO} &
        \multicolumn{4}{c|}{Objects365} \\
        \midrule
        \multicolumn{2}{c|}{"Mouse"} & 
        \multicolumn{2}{c|}{"Potted plant"} & 
        \multicolumn{2}{c|}{"Faucet"} &
        \multicolumn{2}{c}{"Egg"} 
        \\
         \includegraphics[width=0.12\textwidth,height=0.12\textwidth]{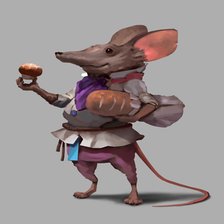} & 
         \includegraphics[width=0.12\textwidth,height=0.12\textwidth]{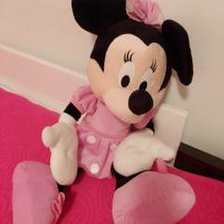} & 
         \includegraphics[width=0.12\textwidth,height=0.12\textwidth]{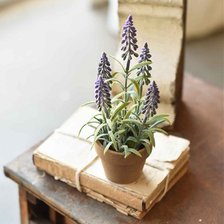} & 
         \includegraphics[width=0.12\textwidth,height=0.12\textwidth]{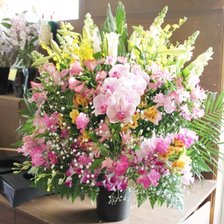} & 
         \includegraphics[width=0.12\textwidth,height=0.12\textwidth]{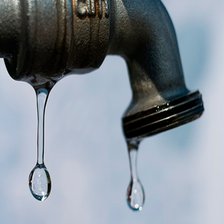} & 
         \includegraphics[width=0.12\textwidth,height=0.12\textwidth]{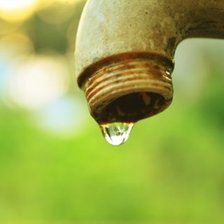} & 
         \includegraphics[width=0.12\textwidth,height=0.12\textwidth]{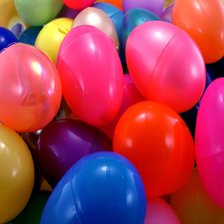} & 
         \includegraphics[width=0.12\textwidth,height=0.12\textwidth]{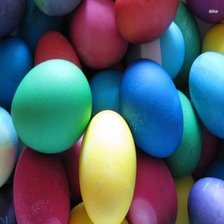}          
         \\

         & 
    \end{tabular}
    \caption{We highlight several failure cases of the object detector identified during our human evaluation. All images presented here have a confidence score below the threshold of $0.1$ and are therefore not rejected by our automated pipeline. 
For "mountain bike," the primary issue arises from the objects being very small and difficult to spot. In the case of other objects, such as "Pot" or "Faucet," the objects are clearly visible, but the object detector fails to recognize these instances. For "Car," the detector does not seem to classify trucks or vans as cars. For "Mouse" and "Egg," the detector struggles with distribution shifts, failing to recognize comic or plush mice and colored eggs, respectively.
}
    \label{fig:app_detection_yes}
\end{figure*}

For the object detector \owl~\cite{minderer2024scaling} in our pipeline, we pass the object name \obj and the image to the model. The model then returns a predefined number of bounding boxes, each with a confidence score in the range [0,1]. We take the maximum confidence over all bounding boxes and use this as the probability of the image containing the object, \ie, $p_\text{det}\left( \text{\obj} \mid \text{img} \right)$. We then reject all images where $p_\text{det}$ is greater than our threshold of $0.1$.

To verify our automatic pipeline, and especially the conservative threshold for the object detector, we manually labeled 10 random images for each object for \oursopt on \pali. As stated in \cref{sec:exp-retrievalresults}, we use the labels ``\yes'' if the object is visible, ``\no'' if it is absent, and ``ambiguous'' for corner cases. Across all images, we find that 5.2\% contain the object and 7.8\% are ambiguous.

We additionally provide a per-dataset breakdown over object classes in \cref{fig:app_detection_bars}, where we plot the ``yes,'' ``no,'' and ``ambiguous'' ratios. Notably, most objects do not contain any instances of the specified object. Instead, the majority of errors stem from a few object categories where the object detector exhibits systematic issues. Qualitative examples are shown in \cref{fig:app_detection_ambiguous} and \cref{fig:app_detection_yes}.

We observed that some images were labeled as ``ambiguous'' in our human evaluation due to various factors. For instance, in the cases of ``Barn'' and ``Bookshop,'' the limited image resolution made it difficult to identify specific objects; distinguishing a house from a barn in an aerial view is nearly impossible. For ``Kai yang,'' a Thai chicken dish, while the depicted dishes might contain chicken, it is challenging to determine whether they are specifically ``Kai yang.'' Interestingly, a reverse image search labels the first image as ``kebab.'' 

Similarly, for ``Cowry,'' which refers to small sea snails, even if our human labelers could not identify any, it is difficult to guarantee their absence in the image. In the case of ``Airplane,'' the interiors shown could represent futuristic airplane or train designs, making it ambiguous. For ``Train,'' the low image resolution hinders the inference of specific objects' presence.

Furthermore, ambiguity arises from the object labels themselves in some datasets. For example, in Objects365, ``Glasses'' refers to eyewear, but the images often contain multiple glass objects, causing confusion. Likewise, ``Soccer'' refers exclusively to a soccer ball in Objects365, whereas the sport itself is not a well-defined object, leading to uncertainty about whether to label images of referees as ``\yes'' or ``no.''

In addition to ambiguous cases, we identified several failure cases of the object detector during our human evaluation. All images in these cases had a confidence score below the threshold of $0.1$ and were therefore not rejected by our automated pipeline. For ``Mountain bike,'' the primary issue was that the objects were very small and difficult to spot. In other instances, such as ``Pot'' or ``Faucet,'' the objects are clearly visible, but the object detector failed to recognize them. For ``Car,'' the detector did not classify trucks or vans as cars. Similarly, for ``Mouse'' and ``Egg,'' the detector struggled with distribution shifts, failing to recognize comic or plush mice and colored eggs, respectively.

These observations suggest that while our object detector generally performs well, there are specific categories and scenarios where it struggles, either due to ambiguity in object definitions or limitations in detecting certain object variations.

\begin{figure*}[t]
    \centering
    \includegraphics[width=\linewidth]{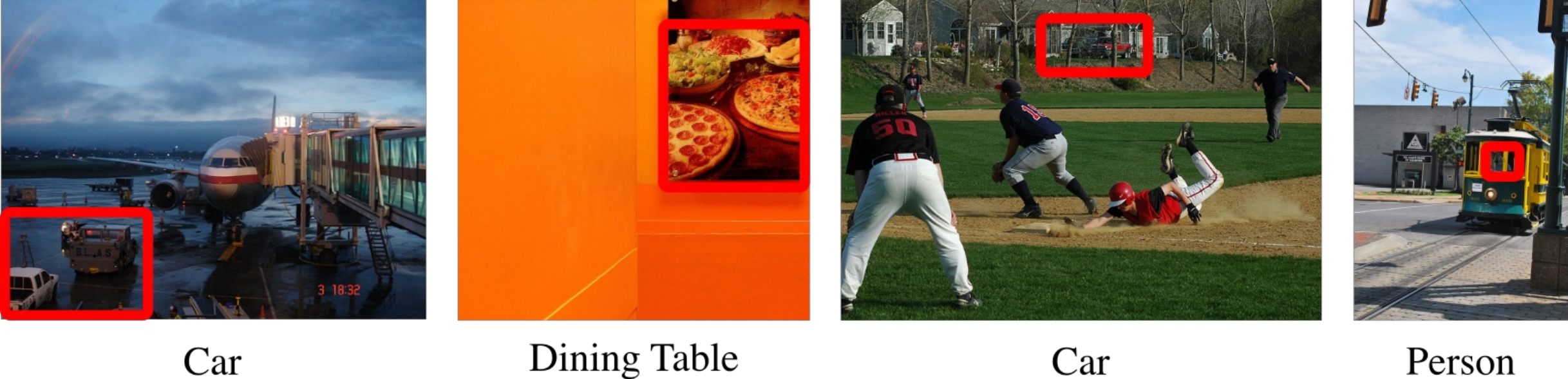} 
    \caption{\textbf{COCO annotations errors in POPE (ground truth ``no''):} We show four examples where the POPE ground truth label for the question ``Is there a \emph{object} in the image?'' is ``no'' although the object is present in the image. We mark the location of the object with a red bounding box.}
    \label{fig:pope-errors}
\end{figure*}

\yn{
\section{Effect of COCO annotation errors on POPE}
\label{app:pope-errors}
Current VLMs only produce a small number of false positives on the POPE benchmark, e.g. \pali predicts ``\yes'' on 137 out of the 4500 samples which do not contain the corresponding object according to the COCO annotations. We re-annotate these images and assign the labels
\begin{itemize}
    \item ``yes'' if the object is visible in the image,
    \item ``no'' if the object is \textbf{not} visible in the image,
    \item ``ambiguous'' for corner cases where it is not clear whether the object is present or not.
\end{itemize}
The result of our labeling is that 35 ($25.5\%$) of the alleged false positives actually do contain the object which means that the model reply ``\yes'' is actually correct (see Fig.~\ref{fig:pope-errors} for examples). In addition, 31 ($22.6\%$) of the images receive the label "ambiguous". This large amount of label noise among the remaining false positives indicates that the POPE benchmark is saturated.
}

\section{Transfer}
\label{app:transfer}

\begin{figure*}[htbp]
    \centering
    \small
    \setlength{\tabcolsep}{1pt} %
    \renewcommand{\arraystretch}{1} %
    \begin{tabular}{ccc|ccc|ccc}
        \toprule
        \multicolumn{3}{c|}{"Anglerfish"} &
        \multicolumn{3}{c|}{"Electric Ray"} &
        \multicolumn{3}{c}{"Artifical Nails"} 
        \\
        \multicolumn{3}{c|}{\makecell{\qwen-VL-72B: "yes" \\ \llama-3.2-Vision: "no"}} &
        \multicolumn{3}{c|}{\makecell{\qwen-VL-72B: "yes" \\ \llama-3.2-Vision: "no"}} &
        \multicolumn{3}{c}{\makecell{\qwen-VL-72B: "yes" \\ \llama-3.2-Vision: "no"}} \\
        
         \includegraphics[width=0.10\textwidth,height=0.10\textwidth]{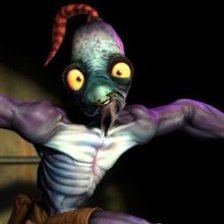} & 
         \includegraphics[width=0.10\textwidth,height=0.10\textwidth]{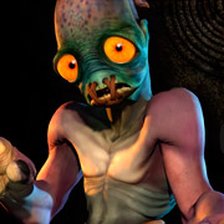} & 
         \includegraphics[width=0.10\textwidth,height=0.10\textwidth]{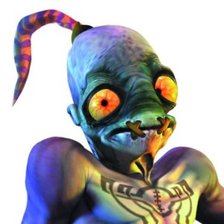} & 
         \includegraphics[width=0.10\textwidth,height=0.10\textwidth]{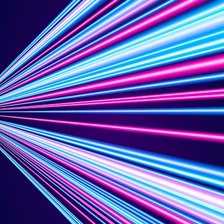} & 
         \includegraphics[width=0.10\textwidth,height=0.10\textwidth]{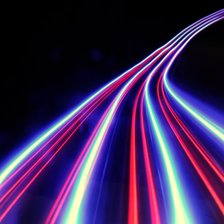} & 
         \includegraphics[width=0.10\textwidth,height=0.10\textwidth]{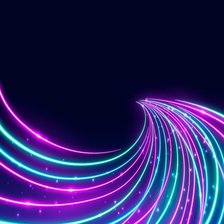} & 
         \includegraphics[width=0.10\textwidth,height=0.10\textwidth]{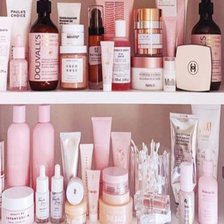} & 
         \includegraphics[width=0.10\textwidth,height=0.10\textwidth]{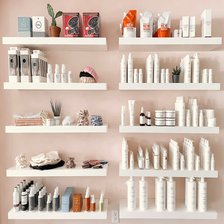} & 
         \includegraphics[width=0.10\textwidth,height=0.10\textwidth]{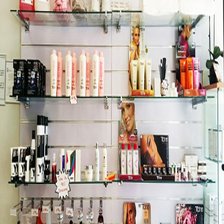}  
         \\
        \bottomrule
         
         \\
        \toprule
        \multicolumn{3}{c|}{"Concertina"} &
        \multicolumn{3}{c|}{"Ford Tourneo"} &
        \multicolumn{3}{c}{"Pipette"} 
        \\
        \multicolumn{3}{c|}{\makecell{\qwen-VL-72B: "no" \\ \llama-3.2-Vision: "yes"}} &
        \multicolumn{3}{c|}{\makecell{\qwen-VL-72B: "no" \\ \llama-3.2-Vision: "yes"}} &
        \multicolumn{3}{c}{\makecell{\qwen-VL-72B: "no" \\ \llama-3.2-Vision: "yes"}} \\
        
         \includegraphics[width=0.10\textwidth,height=0.10\textwidth]{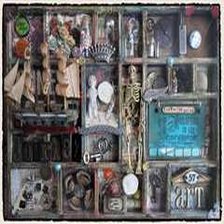} & 
         \includegraphics[width=0.10\textwidth,height=0.10\textwidth]{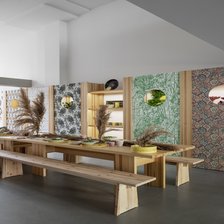} & 
         \includegraphics[width=0.10\textwidth,height=0.10\textwidth]{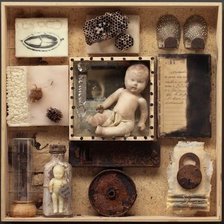} & 
         \includegraphics[width=0.10\textwidth,height=0.10\textwidth]{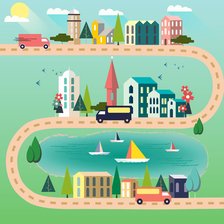} & 
         \includegraphics[width=0.10\textwidth,height=0.10\textwidth]{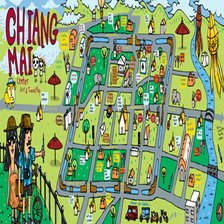} & 
         \includegraphics[width=0.10\textwidth,height=0.10\textwidth]{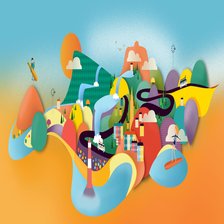} & 
         \includegraphics[width=0.10\textwidth,height=0.10\textwidth]{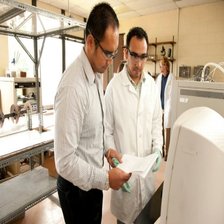} & 
         \includegraphics[width=0.10\textwidth,height=0.10\textwidth]{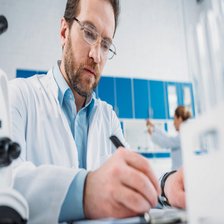} & 
         \includegraphics[width=0.10\textwidth,height=0.10\textwidth]{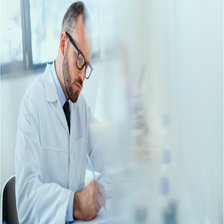} 
         \\
        \bottomrule
    \end{tabular}
    \caption{We demonstrate several images where \qwen-VL-72B and \llama-3.2-Vision disagree. Note that all images do not contain the object and are thus hallucinations by the model responding with "yes". This further demonstrates that even the best available open-weight models are not robust to hallucinations.}
    \label{fig:app_qwen_vs_llama}
\end{figure*}

\subsection{VLM Models}\label{sec:app_vlm_models}
For all models except Prismatic, we use the Transformers library~\cite{wolf2019huggingface} with the official checkpoints. For Prismatic~\cite{karamcheti2024prismatic}, we use the official implementation. Model details, including links to the specific models files used can be found in \cref{tab:app_vlms}. Note that \qwen-VL is not based on \siglip as stated in \cref{tab:transfer} but instead uses a custom ViT with 675m parameters. We will correct this in the final version.

\begin{table*}[htbp]
\centering
\footnotesize
\begin{tabular}{l|l|l|l}
\textbf{VLM Model} & \textbf{LLM} & \textbf{Vision Encoder} & \textbf{Checkpoint}\\
\toprule
    \pali-3B~\cite{beyer2024paligemma} & Gemma-2B~\cite{team2024gemma} & \siglip-So400m 224px~\cite{zhai2023siglip} & \href{https://huggingface.co/google/paligemma-3b-mix-224}{paligemma-3b-mix-224}\\ 
    \midrule
    \llavanext-\mistral-7B~\cite{liu2024llavanext,liu2023improvedllava}  & \vicuna-7B~\cite{peng2023vicuna} & CLIP ViT-L 224px~\cite{radford2021clip} & \href{https://huggingface.co/llava-hf/llava-v1.6-vicuna-7b-hf}{llava-v1.6-vicuna-7b-hf} \\ 
    \llavanext-\vicuna-7B~\cite{liu2024llavanext,liu2023improvedllava} & \mistral-7B~\cite{jiang2023mistral7b} & CLIP ViT-L 224px~\cite{radford2021clip} & \href{https://huggingface.co/llava-hf/llava-v1.6-mistral-7b-hf}{llava-v1.6-mistral-7b-hf}\\ 
    \llavanext-\llama-8B~\cite{li2024llavanext-strong,liu2023improvedllava} & \llama-3.0-8B~\cite{dubey2024llama3herdmodels} & CLIP ViT-L 224px~\cite{radford2021clip} & \href{https://huggingface.co/llava-hf/llama3-llava-next-8b-hf}{llama3-llava-next-8b-hf}\\ 
    \midrule
    \prismatic \clip~\cite{karamcheti2024prismatic} &  \vicuna-7B~\cite{peng2023vicuna} & CLIP ViT-L 224px~\cite{radford2021clip} & \href{https://huggingface.co/TRI-ML/prismatic-vlms/tree/main/clip-224px%2B7b}{prismatic-vlms/clip-224px+7b} \\
    \prismatic \siglip~\cite{karamcheti2024prismatic} &  \vicuna-7B~\cite{peng2023vicuna} & \siglip-So400m 224px~\cite{zhai2023siglip} & \href{https://huggingface.co/TRI-ML/prismatic-vlms/tree/main/siglip-224px%2B7b}{prismatic-vlms/siglip-224px+7b} \\
    \prismatic \dino~\cite{karamcheti2024prismatic} &  \vicuna-7B~\cite{peng2023vicuna} & DINOv2 ViT-L 224px~\cite{oquab2023dinov2} & \href{https://huggingface.co/TRI-ML/prismatic-vlms/tree/main/dinov2-224px%2B7b}{prismatic-vlms/dinov2-224px+7b} \\
    \midrule
    \qwen-VL-7B-Instruct & \qwen-7B~\cite{yang2024qwen2} & Custom ViT 675m & \href{https://huggingface.co/Qwen/Qwen2-VL-7B-Instruct}{Qwen2-VL-7B-Instruct} \\
    \qwen-VL-72B-Instruct & \qwen-72B~\cite{yang2024qwen2} & Custom ViT 675m & \href{https://huggingface.co/Qwen/Qwen2-VL-72B-Instruct}{Qwen2-VL-72B-Instruct} \\
    \midrule
    \llama-3.2-11B-Vision-Instruct & \llama-3.1-8B~\cite{dubey2024llama3herdmodels} & Custom ViT & \href{https://huggingface.co/meta-llama/Llama-3.2-11B-Vision-Instruct}{Llama-3.2-11B-Vision-Instruct} \\
    \bottomrule
\end{tabular}
\caption{VLMs used for transfer experiments}
\label{tab:app_vlms}

\end{table*}

\subsection{True positive rate}
\label{app:tpr}
For each object from ImageNet, COCO, and Objects365, we collect 100 images of the corresponding class from the official validation set. On these images, the average TPR is computed by counting the frequency of the correct response ``yes''.

\subsection{\qwen-VL vs \llama 3.2-VL}
In the main paper, we have already shown some examples from \oursb where, by design, both \qwen-VL and \llama 3.2-VL hallucinate. While both these models are quite robust to hallucinations, we also want to understand where they differ. To do this, we show several examples in \cref{fig:app_qwen_vs_llama} where only one model hallucinates. This demonstrates that even the best available open-weight models are still vulnerable to hallucinations but also differ substantially in terms of vulnerabilities, likely due to larger differences in architecture, vision encoder, LLM, and training data. 

\yn{
\section{\ours-B}
\label{app:benchmark}

As described in \ref{app:pope-errors}, most models only produce a small number of false positives on POPE which also contain a large amount of label errors. Therefore, we propose a new benchmark \ours-B based on our retrieval results to enable a more reliable and rigorous evaluation of object hallucinations. Tab.~\ref{app:benchmark} contains results for \ours-B and POPE for a range of VLMs.

\subsection{Image Selection}
We select the images for the benchmark using the following steps:
\begin{itemize}
    \item We merge the images found by \oursllm and \oursopt over all three source models.
    \item These images are filtered by requiring a successful transfer to both Qwen2-72B and Llama 3.2-VL-11B, the best performing models in \ref{tab:transfer}, in order to exclude errors which are specific to biases of the three source models.
    \item We select 70 objects and two human labelers verify that the selected images do not contain the corresponding object.
    \item The number of images is limited to at least 3 and at most 50. 
    \item For each object, the same amount of positive samples, i.e. images that contain the object, are added. These images are retrieved using the Flickr API \cite{flickrAPI} and annotated by a human labeler to ensure that the object is clearly contained.
\end{itemize}

\subsection{Metrics}
The performance measure on \ours-B is the accuracy over all negative and positive samples. In Tab.~\ref{app:benchmark}, we also report the true negative rate (TNR) and true positive rate (TPR) individually. A downside of measuring accuracy is that a trivial model that always replies ``\yes'' (or always ``\no'') achieves an accuracy of $50\%$. This behaviour can be avoided by considering the harmonic mean of TNR and TPR instead which results in a value of 0 for the trivial case. We also report this metric (HM) in Tab.~\ref{app:benchmark} but observe no significant effect on the results (apart from \llavanext-Vicuna). Note that the results for the three source models are biased as they were used in the creation of the benchmark. Similarly, Qwen2-72B and Llama-3.2-11B are not reported as they produce a TNR of 1.0 by design.
}
\begin{table*}
\centering
\footnotesize

\begin{tabular}{l |c|c|c|c|c}
\toprule
Benchmark & POPE & \multicolumn{4}{c}{\ours-B}\\
Metric & Acc. & Acc. & TNR & TPR & HM \\
\midrule

\pali-3B~\cite{beyer2024paligemma}& $87.2\%$  & $62.0\%$  & $26.4\%$  & $97.7\%$  & $41.6\%$ \\
\lnvicuna~\cite{liu2024llavanext, peng2023vicuna}& $87.6\%$ & $53.7\%$  & $10.4\%$  & $96.9\%$  & $18.7\%$ \\
\lnmistral~\cite{liu2024llavanext,jiang2023mistral7b}& $88.0\%$  & $61.7\%$  & $30.1\%$  & $93.4\%$  & $45.5\%$ \\
\midrule
\lnllama~\cite{liu2024llavanext, dubey2024llama3herdmodels}& $88.0\%$ & $65.2\%$  & $37.0\%$  & $93.4\%$  & $53.0\%$ \\
Llava-OneVision~\cite{li2024llavaOneVision} & $88.7\%$  & $75.1\%$  & $60.2\%$  & $90.1\%$  & $72.2\%$ \\
\pali-2-3B~\cite{steiner2024paligemma} & $88.8\%$  & $68.9\%$  & $40.9\%$  & $96.8\%$  & $57.5\%$ \\
\pali-2-10B~\cite{steiner2024paligemma} & $87.7\%$ & $69.8\%$  & $48.0\%$  & $91.6\%$  & $63.0\%$ \\
Ovis2-1B~\cite{lu2024ovis} & $88.9\%$ & $64.6\%$  & $35.1\%$  & $94.0\%$  & $51.1\%$ \\
Ovis2-2B~\cite{lu2024ovis} & $89.4\%$ & $61.7\%$  & $27.3\%$  & $96.1\%$  & $42.5\%$ \\
Ovis2-4B~\cite{lu2024ovis} & $90.3\%$ & $64.8\%$  & $31.0\%$  & $98.6\%$  & $47.2\%$ \\
Ovis2-8B~\cite{lu2024ovis} & $94.9\%$ & $71.4\%$  & $44.8\%$  & $98.0\%$  & $61.5\%$ \\
InternVL2.5-8B~\cite{chen2024expanding} $\star$& $90.6\%$ & $71.7\%$  & $47.2\%$  & $96.2\%$  & $63.3\%$ \\
InternVL2.5-26B~\cite{chen2024expanding} $\star$ & $90.6\%$  & $77.5\%$  & $57.3\%$  & $97.8\%$  & $72.2\%$ \\
InternVL2.5-38B~\cite{chen2024expanding} $\star$ & $90.7\%$ & $76.2\%$  & $54.8\%$  & $97.6\%$  & $70.2\%$ \\
InternVL2.5-78B~\cite{chen2024expanding} $\star$ & $90.8\%$ & $74.1\%$  & $50.3\%$  & $97.8\%$  & $66.5\%$ \\ 
InternVL2.5-8B-MPO~\cite{chen2024expanding} $\dag$ & $89.1\%$ & $69.4\%$  & $42.3\%$  & $96.4\%$  & $58.8\%$ \\
InternVL2.5-26B-MPO~\cite{chen2024expanding} $\dag$& $90.7\%$  & $76.1\%$  & $54.8\%$  & $97.4\%$  & $70.1\%$ \\
GPT-4o-mini$\star$ & $84.2\%$ & $86.3\%$ & $77.0\%$  & $95.7\%$  & $85.3\%$ \\

\bottomrule
\multicolumn{6}{r}{$\star$: POPE result from %
\cite{chen2024expanding}, \dag : POPE result from \cite{duan2024vlmevalkit}}\\
\end{tabular}

\caption{\textbf{\ours-B:} We report accuracy (for POPE and \oursb) as well as the true negative rate (TNR), true positive positive rate (TPR), and the harmonic mean of TNR and TPR (HM). While the accuracy reflects the detection-hallucination trade-off, the individual values of TNR and TPR can give further insides into the vulnerability to \hallu s. Note that \pali-3B, \lnvicuna, and \lnmistral were used in the creation of the benchmark.}
\label{tab:app-benchmark}

\end{table*}

\section{Fine-tuning on \ours}\label{app:mitigation}
Can we utilize the images retrieved by \ours to mitigate the vulnerability to systematic hallucinations? To test this hypothesis, we perform a small scale experiment by fine-tuning \pali-3B with LoRA\cite{hu2021lora} on our retrieval results. Used hyperparameters are provided in \cref{tab:app-ft-hyper}.

\subsection{Data}
\label{app:mit-data}
\ours retrieves images, where the object is not present in the image. Therefore, the ground truth answer to the question ``Can you see an \emph{object} in this image?'' is always ``\no''. We additionally retrieve images containing the object and add them to the training data to preserve the model's ability to recognize the object. For each object, we add 200 negative samples, i.e. images where ``\no'' is the correct reply, and 400 positive samples, i.e. ``\yes'' is the correct reply, at random to the training set.

\textbf{Negative samples}:
For the negative samples, i.e. images where the ground truth answer is ``\no'', we use all images resulting from \oursllm and \oursopt (both for \pali). We split these images into two disjoint subsets:

\begin{itemize}
    \item \emph{\evalonly}: For each object, one of the found clusters is selected and all corresponding images are placed in the validation set.
    \item \emph{Train}: All remaining images are used to sample images for the fine-tuning dataset. We further filter these images to ensure that they do not contain the object by requiring that \llama 3.2-VL and \qwen-VL answer with ``no''.
\end{itemize}

\textbf{Positive samples}: We generate a diverse set of prompts including the objects using \llama 3.2 and use them to retrieve images from ReLAION. The resulting images are filtered by the object detector (threshold $>$ 0.1) and \llama 3.2 (response ``\yes'').

\begin{table}[htbp]
    \centering
    \begin{tabular}{c|c}
        \hline
        Optimizer & ADAM \\
        $\beta_1$ & $0.9$ \\
        $\beta_2$ & $0.999$ \\
        Learning rate & 1e-6 \\
        Number of epochs & 5 \\
        Batchsize & 32 \\
        \hline
        LoRA rank & 8 \\
        \hline
    \end{tabular}
    \caption{Fine-tuning hyperparameters}
    \label{tab:app-ft-hyper}
\end{table}

\subsection{Results}

We report several metrics for \pali-3B and our fine-tuned version (+ft) in \cref{tab:app-ft}, comparing their performance on different tasks:
\begin{itemize}
    \item \textbf{Systematic hallucinations}: The accuracy, i.e. ratio of correctly replying with ``no'', on the \evalonly set.%
    \item \textbf{Hallucination benchmarks with similar tasks:} We report the Amber score and the accuracies on Amber Existence and R-Bench.
    \item \textbf{Effect on other tasks:} We evaluate two VQA benchmarks (TextVQA~\cite{singh2019towards}, VQAv2~\cite{balanced_vqa_v2}) and two captioning benchmarks (COCO\cite{lin2014coco}, Flickr30k~\cite{Young2014FromID}) and report accuracies and CiDER scores, respectively.
    \item \textbf{Performance on positive samples:} The TPR-ICO for the objects from ImageNet, COCO, and Objects365 are evaluated as described in \cref{app:tpr}.
\end{itemize}
The fine-tuned version (+$\text{ft}_{pre}$) significantly improves over \pali-3B on unseen clusters (\evalonly, $+77.7\%$). It also shows slightly better results on related hallucination benchmark with increases in the Amber score ($+0.2$), as well as higher accuracies on Amber Existence ($+3\%$) and R-Bench ($+1.1\%$). The performance decreases slightly for more general VQA tasks ($-1.4\%$ and $-0.9\%$) and captioning tasks ($-1.3$ and $+0.1\%$). The reduction of the TPR-ICO ($-5.1\%$) is due to a significant drop of the TPR on Objects365 ($-12.1\%$). A possible reason for this is a mismatch between the image distributions of the retrieved positive samples, where the object is prominently visible in the image, and Objects365, where objects often occur only in small bounding boxes inside the image. Evidence for this is shown in the last column (+$\text{ft}$) of \cref{tab:app-ft}: We repeated the fine-tuning on a different dataset where we replaced all retrieved positive samples for objects from Objects365 with images from the original Objects365 training set. In this setting with more positive than negative samples, the fine-tuned model even improves TPR on all three datasets but also improves less on the hallucination tasks. This experiment indicates that the images retrieved by \ours can also be used to mitigate the problem of systematic hallucination by including them into a fine-tuning routine.

\begin{table}[htbp]
    \centering
    \footnotesize
    \begin{tabular}{c|c||c|c|c}
        Dataset & Metric & PaliG & +$\text{ft}_{pre}$ & +$\text{ft}$ \\
        \midrule
        \midrule
        
        \evalonly & Acc. & $0.0\%$ & $\mathbf{77.7\%}$ & $57.6\%$ \\ %
        Amber & Score & $93.5$ & $93.7$ & $\mathbf{94.0}$ \\ %
        Amber Ex. & Acc. & $93.2\%$ & $\mathbf{96.2\%}$ & $95.4\%$ \\ %
        R-Bench & Acc. & $79.9\%$ & $\mathbf{81.0\%}$ & $80.2\%$ \\ %
        \midrule
        TextVQA & Acc. & $\mathbf{57.6}\%$ & $56.2\%$ & $56.5\%$ \\ %
        VQAv2 & Acc. & $\mathbf{83.1\%}$ & $82.2\%$ & $82.4\%$ \\ %
        \midrule
        COCO & CiDER & $\mathbf{124.5}$ & $123.2$ & $121.3$ \\ %
        Flickr30k & CiDER & $77.4$ & $\mathbf{77.5}$ & $77.1$ \\ %
        \midrule 
        ImageNet & TPR & $90.0\%$ & $90.0\%$ & $\mathbf{93.4}\%$ \\
        COCO & TPR & $84.0\%$ & $80.4\%$ & $\mathbf{88.8}\%$ \\
        Objects365 & TPR & $69.0\%$ & $56.6\%$ & $\mathbf{73.0}\%$ \\
        TPR-ICO & TPR & $81.1\%$ & $76.0\%$ & $\mathbf{85.1}\%$ \\ %
        \midrule
        \oursb & Acc. & $56.4\%$ & - & $\mathbf{68.0}\%$ \\
        \oursb & TNR & $26.4\%$ & - & $\mathbf{45.9}\%$ \\
        \oursb & TPR & $86.4\%$ & - & $\mathbf{90.0}\%$ \\
        \midrule
        
    \end{tabular}
    \caption{Accuracies on our \evalonly set, Amber Existence, and R-Bench and TPR on positive samples from the validation sets of ImageNet, COCO and OpenImages. Fine-tuning on \ours results (+$\text{ft}_{pre}$) can improve robustness against hallucinations significantly, even on clusters not seen during training. It also improves on related hallucination benchmarks while the performance on more general VQA and captioning tasks becomes slightly worse. The reduction in TPR-ICO is caused by the retrieved positive samples for Objects365. After replacing these with images (+$\text{ft}$) from the original training set of Objects365, the fine-tuning even improves average TPR-ICO.}
    \label{tab:app-ft}
\end{table}

\section{Reverse Task}
\label{app:reverse}
\begin{figure*}[htbp]
    \centering
    \setlength{\tabcolsep}{1pt} %
    \renewcommand{\arraystretch}{1} %
    \begin{tabular}{cccccccc}
    \multicolumn{8}{c}{Durian}\\
    \includegraphics[width=0.11\textwidth,height=0.11\textwidth]{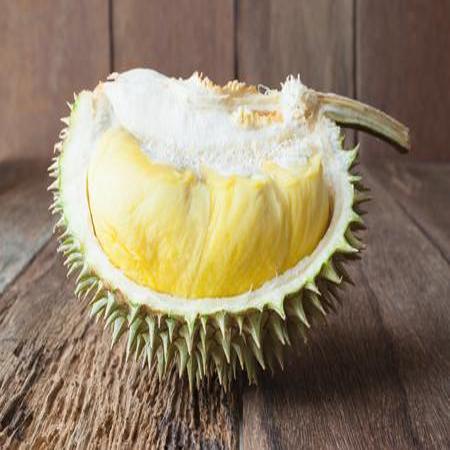} &
     \includegraphics[width=0.11\textwidth,height=0.11\textwidth]{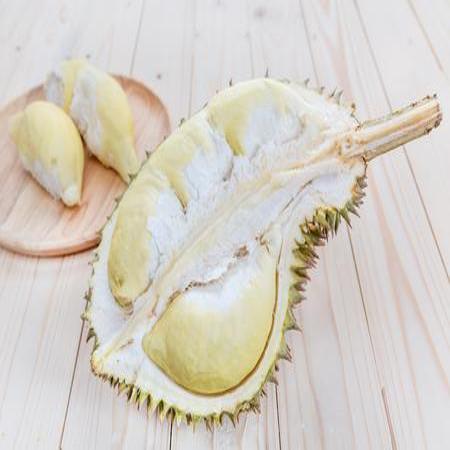} &
     \includegraphics[width=0.11\textwidth,height=0.11\textwidth]{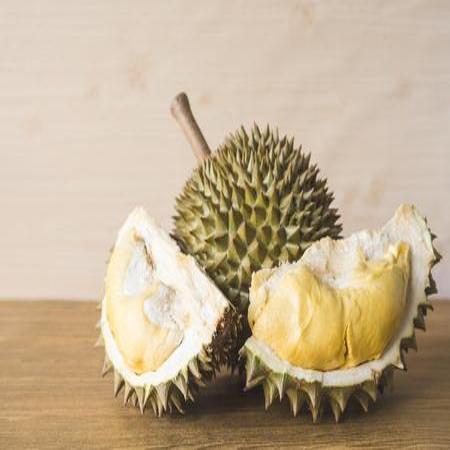} &
     \includegraphics[width=0.11\textwidth,height=0.11\textwidth]{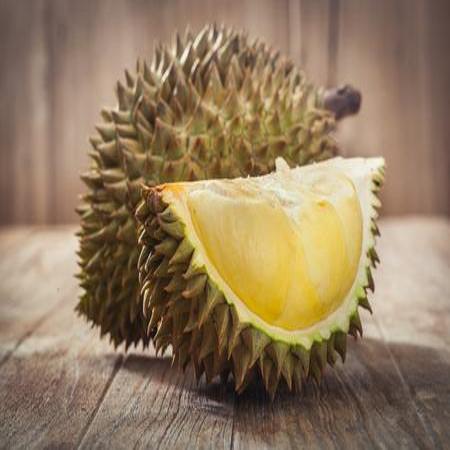} &
     \includegraphics[width=0.11\textwidth,height=0.11\textwidth]{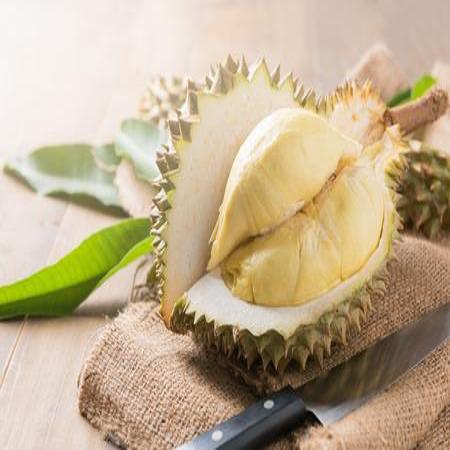} &
     \includegraphics[width=0.11\textwidth,height=0.11\textwidth]{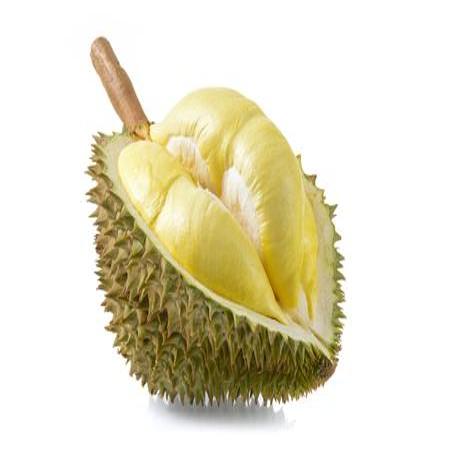} &
     \includegraphics[width=0.11\textwidth,height=0.11\textwidth]{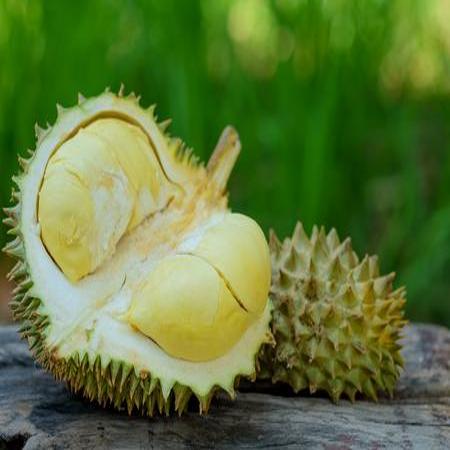} &
     \includegraphics[width=0.11\textwidth,height=0.11\textwidth]{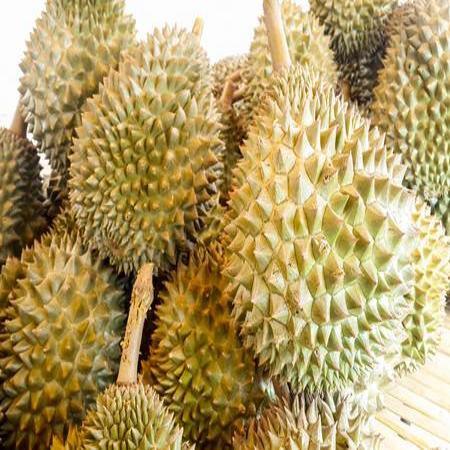} 
     \\
     \multicolumn{8}{c}{Lamp}\\
     \includegraphics[width=0.11\textwidth,height=0.11\textwidth]{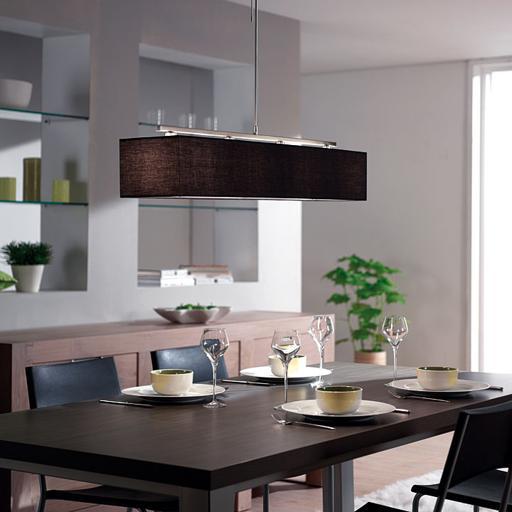} &
     \includegraphics[width=0.11\textwidth,height=0.11\textwidth]{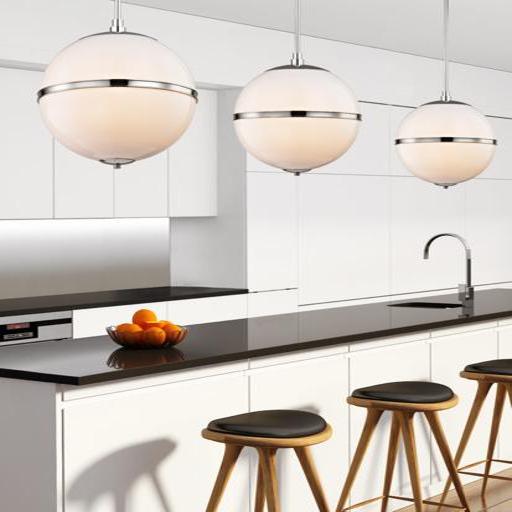} &
     \includegraphics[width=0.11\textwidth,height=0.11\textwidth]{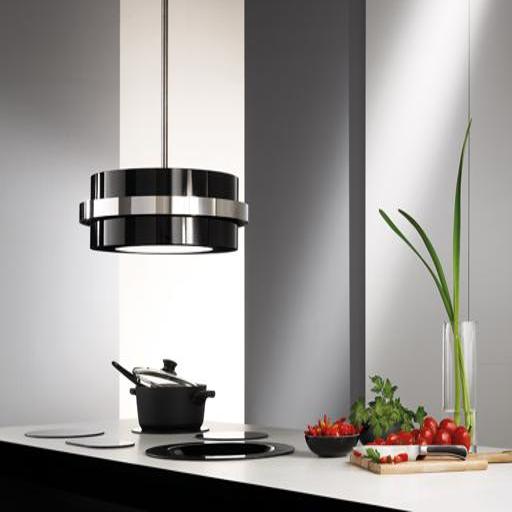} &
     \includegraphics[width=0.11\textwidth,height=0.11\textwidth]{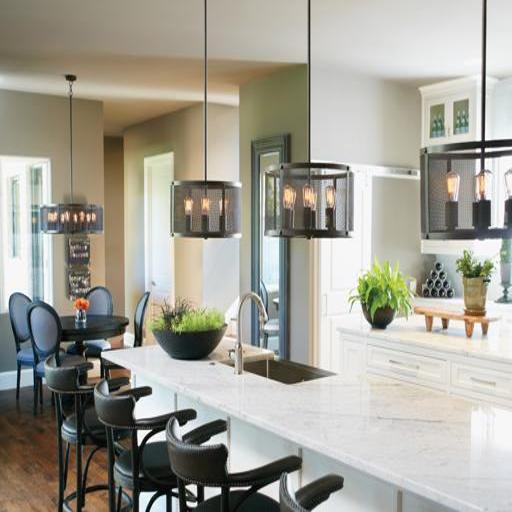} &
     \includegraphics[width=0.11\textwidth,height=0.11\textwidth]{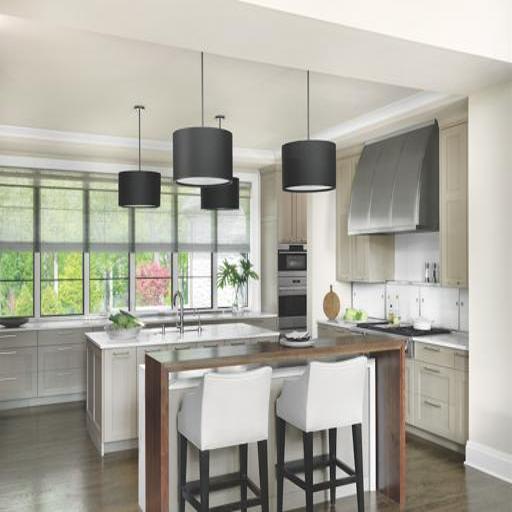} &
     \includegraphics[width=0.11\textwidth,height=0.11\textwidth]{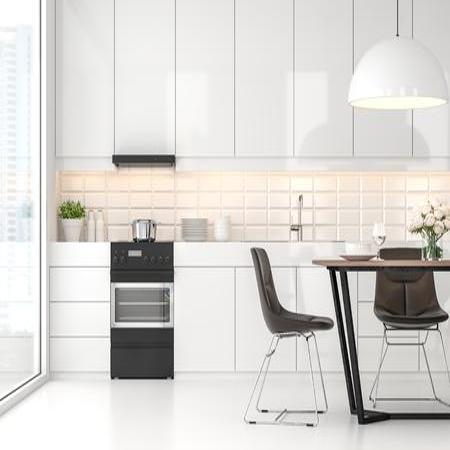} &
     \includegraphics[width=0.11\textwidth,height=0.11\textwidth]{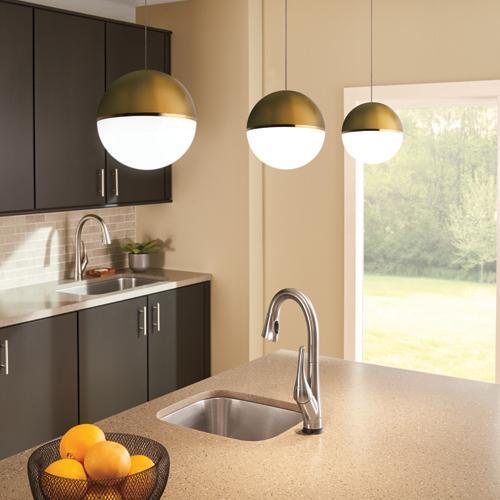} &
     \includegraphics[width=0.11\textwidth,height=0.11\textwidth]{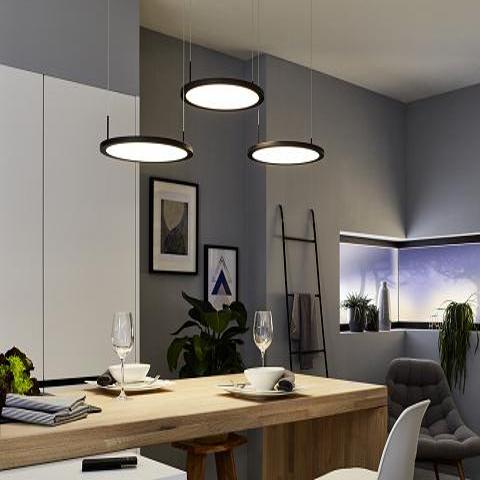} 
     \\
     \multicolumn{8}{c}{Cello}\\
     \includegraphics[width=0.11\textwidth,height=0.11\textwidth]{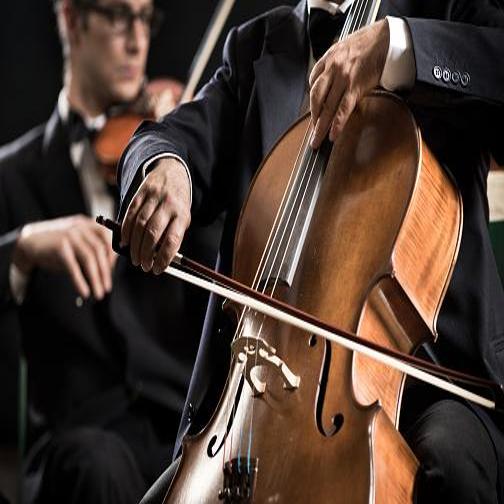} &
     \includegraphics[width=0.11\textwidth,height=0.11\textwidth]{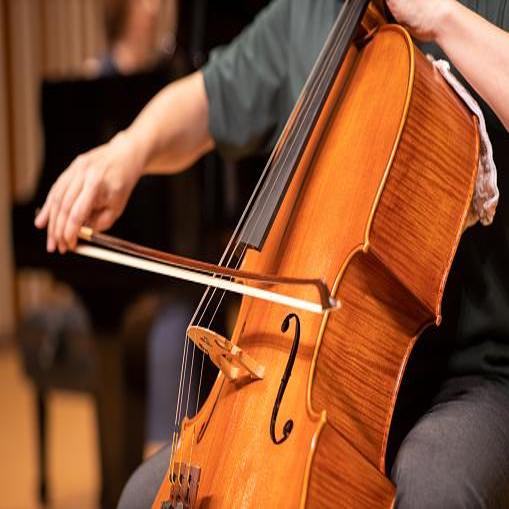} &
     \includegraphics[width=0.11\textwidth,height=0.11\textwidth]{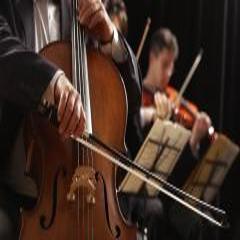} &
     \includegraphics[width=0.11\textwidth,height=0.11\textwidth]{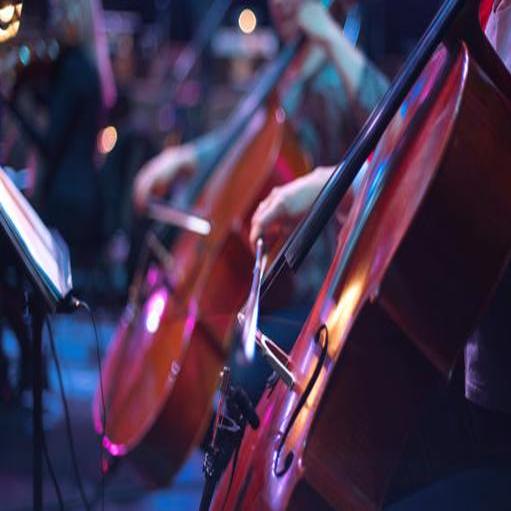} &
     \includegraphics[width=0.11\textwidth,height=0.11\textwidth]{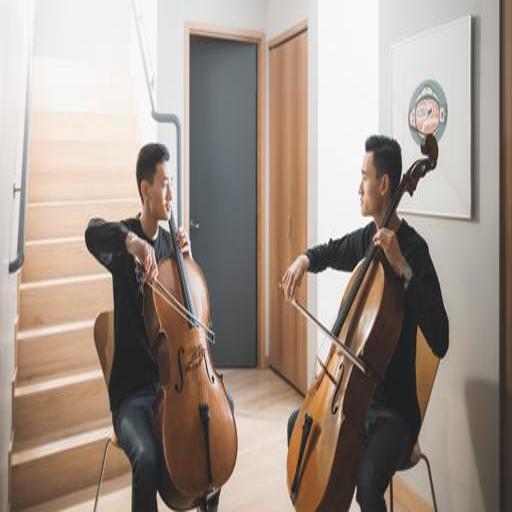} &
     \includegraphics[width=0.11\textwidth,height=0.11\textwidth]{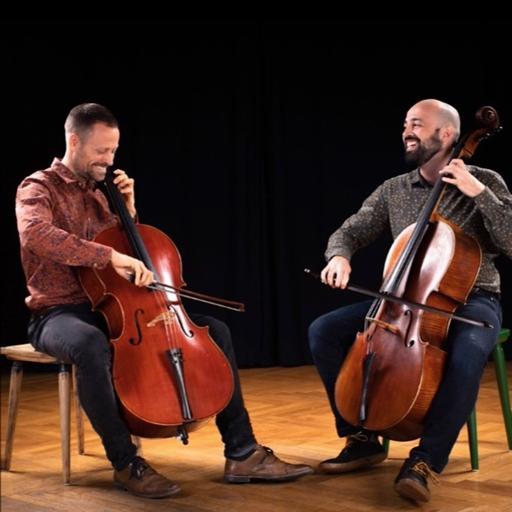} &
     \includegraphics[width=0.11\textwidth,height=0.11\textwidth]{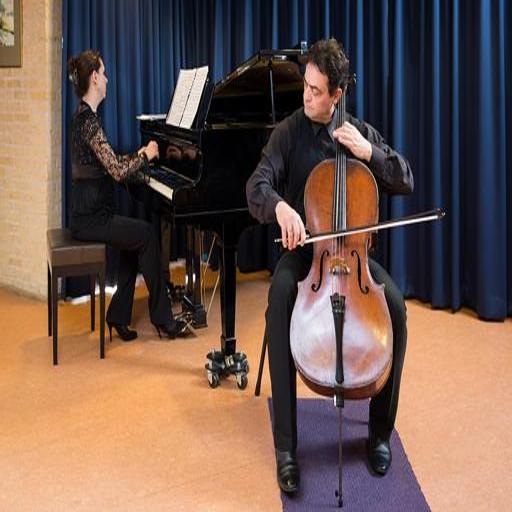} &
     \includegraphics[width=0.11\textwidth,height=0.11\textwidth]{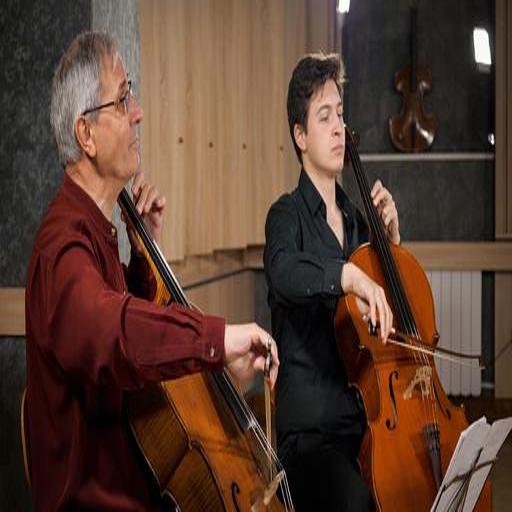}
     \\
     \multicolumn{8}{c}{Chicken}\\
     \includegraphics[width=0.11\textwidth,height=0.11\textwidth]{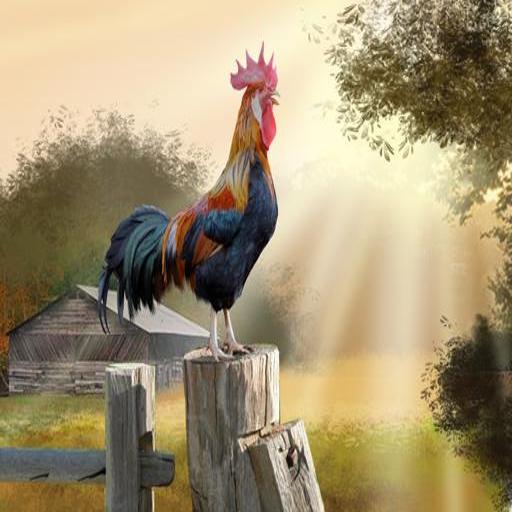} &
     \includegraphics[width=0.11\textwidth,height=0.11\textwidth]{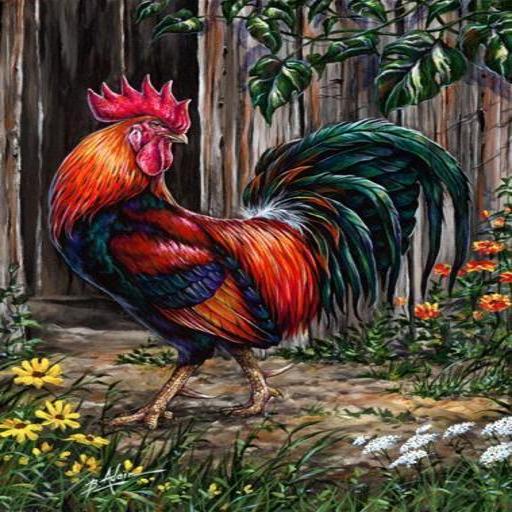} &
     \includegraphics[width=0.11\textwidth,height=0.11\textwidth]{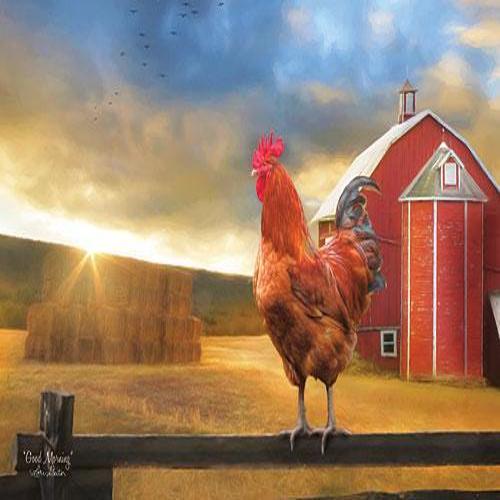} &
     \includegraphics[width=0.11\textwidth,height=0.11\textwidth]{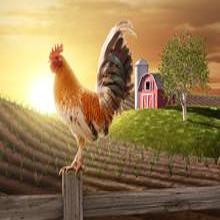} &
     \includegraphics[width=0.11\textwidth,height=0.11\textwidth]{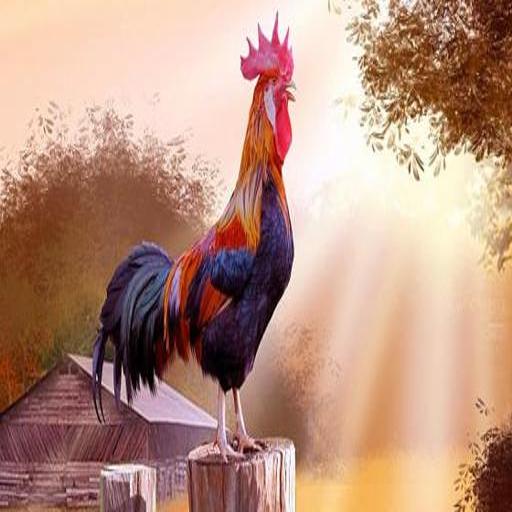} &
     \includegraphics[width=0.11\textwidth,height=0.11\textwidth]{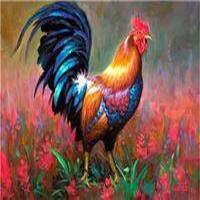} &
     \includegraphics[width=0.11\textwidth,height=0.11\textwidth]{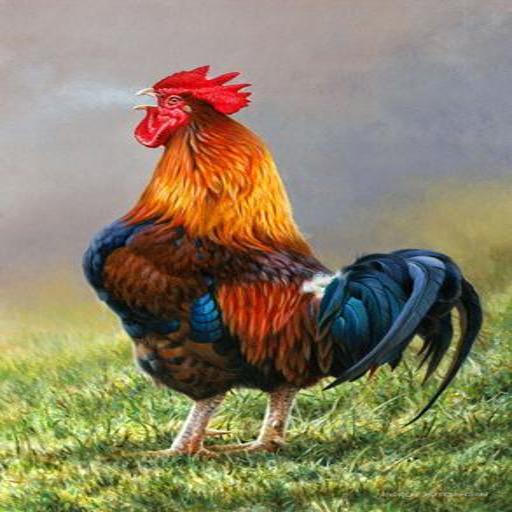} &
     \includegraphics[width=0.11\textwidth,height=0.11\textwidth]{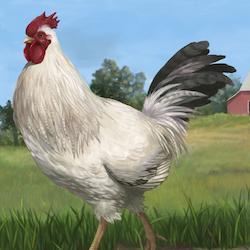}
\end{tabular}
    \caption{\textbf{Reverse Task} We show three clusters found by \ours for the reverse task using \llavanext \vicuna: The VLM response to "Can you see a OBJ in this image?" is "no" although the object is clearly visible in the image.}
    
    \label{fig:reverse_long}
\end{figure*}

\yn{We apply the \oursllm pipeline to the reverse task, where the VLM outputs ``\no'' despite the object being visible in the image. We adjust the LLM prompt accordingly (see \ref{fig:llm-prompt_reverse}), reverse the object detector threshold, and use a larger value. Figure \ref{fig:reverse_long} presents example clusters. While this experiment serves as a proof of concept, we observe that the object detector performs worse in this direction and should be replaced for larger-scale experiments.
Overall, the benefits of \ours are more pronounced in the setting discussed in the main paper, as the number of images containing a given object is much smaller than the number of images that do \emph{not} contain the object.}
\begin{figure*}[p] %
  \centering
  \begin{minipage}{\textwidth}
    \begin{lstlisting}
You are a creative prompt generator. Your task is to:

1. **Accept an object name** (provided by the user).
2. **Generate 20 different image prompts** in realistic everyday settings, filled with various common objects, where the specified object is present but not necessarily the focus of the scene-so it might be overlooked by an object detection system.

---

## Context & Objectives

1. **Purpose**:  
   - We want to depict the given object in real-life scenarios that include multiple other items typically found in the setting.  
   - The object should be there, but the scene should be busy or populated enough that the object isn't the sole focus.  
   - The style should be **highly realistic**, as if taken by a camera.

2. **Guiding Techniques**:  
   - **Crowded Scenes**: Combine the specified object with many other objects commonly found in the same environment (e.g., living rooms, offices, kitchens, garages).  
   - **Non-Focal Positioning**: Place the object off to the side or partially in the background, so it doesn't immediately draw attention.  
   - **Realistic Keywords**: To enhance the lifelike quality, you can use any of these keywords in your prompts:
     - photo-real
     - hyper-detailed
     - 8k resolution
     - cinematic lighting
     - DSLR
     - natural lighting
     - raw photo
     - high dynamic range
     - real-world texture
     - unposed

---

## Detailed Instructions

1. **Input**:  
   You will receive a single word or short phrase specifying the object (e.g., "chair," "cup," "clock," "bag," etc.).

2. **Output**:  
   - Produce **20 unique prompts**, each describing a realistic photograph in which the object is present among various other items typically found in that scenario.  
   - Use some of the realism keywords to convey a high-quality, real-world style.  
   - Ensure the object is not the main focus but simply part of a busier environment.

3. **Format**:  
   - Number each prompt **from 1 to 20, using a colon** (e.g., `1: Prompt text`, `2: Prompt text`, ..., `20: Prompt text`).  
   - Each prompt should be concise but mention multiple items and the general setting.

---
    \end{lstlisting}
  \end{minipage}
  \caption{\oursllm prompt for generating the text queries for the reverse task (1/2)}
\end{figure*}

\begin{figure*}[p] %
  \centering
  \begin{minipage}{\textwidth}
    \begin{lstlisting}

## Examples of Prompts

*(Using `<OBJECT_NAME>` as a placeholder - these are short samples, not fully detailed.)*

- **Living Room Scenario**  
  *"A photo-real image of a cozy living room with a sofa, coffee table, TV, potted plants, and a small `<OBJECT_NAME>` tucked beside a stack of magazines."*

- **Office Setting**  
  *"A hyper-detailed view of an open-plan office featuring desks, laptops, file cabinets, a water cooler, and a `<OBJECT_NAME>` placed casually near a window sill."*

- **Kitchen Scene**  
  *"A raw photo of a busy kitchen counter with plates, utensils, fruits, and a `<OBJECT_NAME>` resting behind a jar of spices."*

Please use many different such scenarios instead of restricting yourself to the ones from these examples.
Possible scenarios would be an office, a train station, a garden, a living room, a kitchen, a hallway, outdoors, in the city, landscape.
Try to think of a scenario that matches the object and that allows you to add in different objects that could occur with it. 

Please try out different scenarios for each object in the different prompts.
Make sure to not repeat too similar prompts and rather create a sufficient variety of prompts. 

These examples show:
- The `<OBJECT_NAME>` is included but not emphasized.
- The setting has multiple other common objects.

---

## Final Output Format

When the user provides the object name, respond with exactly **20 prompts**, numbered with colons, in the form:

1: [Prompt text]
2: [Prompt text] 
... 
20: [Prompt text]

Each prompt should describe a realistic scene filled with everyday objects, incorporating the given object without making it the sole focus.

    \end{lstlisting}
  \end{minipage}
  \caption{\oursllm follow-up prompt for generating the text queries for the reverse task (2/2)}
  \label{fig:llm-prompt_reverse}
\end{figure*}

\section{Transfer across prompts}
\label{app:different-prompts}
During our experiments we use the prompt ``Can a \emph{object} be seen in the image?''. In \cref{tab:different-prompts}, we evaluate a range of 10 different prompts for the three source models on their corresponding \oursllm and \oursopt subsets.

\begin{table*}
    \centering
    \small
    \begin{tabular}{l|ccc|ccc}
        \toprule
          & \multicolumn{3}{c|}{\oursllm} & \multicolumn{3}{c}{\oursopt}\\
          \midrule
         Prompt Transfer Rate & \pali & \lnvicuna & \lnmistral  & \pali & \lnvicuna & \lnmistral\\ 
         \midrule
Can a OBJ be seen in the image?      & 0.861 & 0.923 & 0.824 & 0.878 & 0.872 & 0.832 \\
Does this image have a OBJ?          & 0.890 & 0.891 & 0.806 & 0.859 & 0.863 & 0.781 \\
Does the image show a OBJ?           & 0.860 & 0.737 & 0.733 & 0.832 & 0.712 & 0.733 \\
Does this image contain a OBJ?       & 0.857 & 0.790 & 0.858 & 0.801 & 0.739 & 0.843 \\
Does this picture include a OBJ?     & 0.824 & 0.782 & 0.886 & 0.795 & 0.735 & 0.865 \\
Is a OBJ depicted in this image?     & 0.851 & 0.913 & 0.748 & 0.805 & 0.901 & 0.755 \\
Is there a OBJ present in the image? & 0.787 & 0.807 & 0.860 & 0.758 & 0.735 & 0.870 \\
Is there a OBJ in this image?        & 0.717 & 0.818 & 0.827 & 0.643 & 0.771 & 0.828 \\
Is a OBJ shown in the image?         & 0.670 & 0.880 & 0.792 & 0.593 & 0.835 & 0.804 \\
Is OBJ visible in the image?         & 0.467 & 0.833 & 0.655 & 0.385 & 0.782 & 0.655 \\
Is OBJ in the image?                 & 0.348 & 0.906 & 0.702 & 0.278 & 0.873 & 0.690 \\
          \midrule
Average                              & 0.739 & 0.844 & 0.790 & 0.693 & 0.802 & 0.787 \\
Standard Deviation                   & 0.179 & 0.062 & 0.072 & 0.200 & 0.069 & 0.071 \\
\bottomrule
    \end{tabular}
    \caption{\textbf{Transfer across prompts:} While transfer rates for \lnvicuna and \lnmistral are stable, \pali was pretrained on this task using the prompt "Is OBJ in the image?" and shows lower transfer rates on similar prompts. However, this improved robustness against systematic hallucinations does not generalize to less similar prompts.}
    \label{tab:different-prompts}
\end{table*}

\end{document}